%% file: main.tex
\documentclass[10pt,twocolumn,letterpaper]{article}

\usepackage{cvpr}              
\usepackage[accsupp]{axessibility}  

\usepackage[utf8]{inputenc} 
\usepackage[T1]{fontenc}    
\usepackage{url}            
\usepackage{booktabs}       
\usepackage{amsfonts}       
\usepackage{nicefrac}       
\usepackage{microtype}      
\usepackage{xcolor}         
\usepackage{amsmath}
\usepackage{adjustbox}

\usepackage{pifont}

\newcommand{\yt}{y_t}
\newcommand{\train}{\mathcal{D}^{tr}}
\newcommand{\trainb}{\mathcal{D}^{b}}
\newcommand{\trainc}{\mathcal{D}^{c}}

\definecolor{lightgray}{gray}{0.5}
\input{utils/includes}
\input{utils/general_utils}

\definecolor{cvprblue}{rgb}{0.21,0.49,0.74}
\usepackage[pagebackref,breaklinks,colorlinks,allcolors=cvprblue]{hyperref}

\def\paperID{15842} 
\def\confName{CVPR}
\def\confYear{2025}

\title{PSBD: Prediction Shift Uncertainty Unlocks Backdoor Detection}

\author{Wei Li$^1$, Pin-Yu Chen$^2$, Sijia Liu$^3$, Ren Wang$^{1}$\thanks{Corresponding author: Ren Wang (rwang74@iit.edu)}\\
$^1$Illinois Institute of Technology\\
$^2$IBM Research\\ $^3$Michigan State University}

\begin{document}
\maketitle
\input{sec/0_abstract}
\input{sec/1_introduction}
\input{sec/2_relative_work}
\input{sec/3_preliminary}
\input{sec/4_method}
\input{sec/5_experiments}
\input{sec/6_conclusion}

\section*{Acknowledgement}

This work is supported by the NSF under Grants 2246157 and 2319243. We are thankful for the computational resources made available through NSF ACCESS and Argonne Leadership Computing Facility.

{
    \small
    \bibliographystyle{ieeenat_fullname}
    \bibliography{ref,ref_backdoor}
}

\input{sec/7_appendix}

\end{document}

%% file: utils/includes.tex
\usepackage{multirow,mathtools }

\usepackage{pifont}
\usepackage{color, colortbl}

\usepackage{blindtext}
\usepackage{lipsum}

\usepackage{multirow}
\usepackage{graphicx}
\usepackage{listings}

\usepackage[most]{tcolorbox}

\usepackage{subcaption}

%% file: utils/general_utils.tex
\usepackage{color, colortbl}
\definecolor{LightCyan}{rgb}{0.78,0.94,1}
\definecolor{Gray}{gray}{0.85}
\definecolor{verylightgray}{gray}{0.95}

%% file: sec/0_abstract.tex
\begin{abstract}
\label{sec:abs}
Deep neural networks are susceptible to backdoor attacks, where adversaries manipulate model predictions by inserting malicious samples into the training data. Currently, there is still a significant challenge in identifying suspicious training data to unveil potential backdoor samples. 
In this paper, we propose a novel method, Prediction Shift Backdoor Detection (PSBD), leveraging an uncertainty-based approach requiring minimal unlabeled clean validation data. PSBD is motivated by an intriguing Prediction Shift (PS) phenomenon, where poisoned models' predictions on clean data often shift away from true labels towards certain other labels with dropout applied during inference, while backdoor samples exhibit less PS. We hypothesize PS results from the neuron bias effect, making neurons favor features of certain classes. PSBD identifies backdoor training samples by computing the Prediction Shift Uncertainty (PSU), the variance in probability values when dropout layers are toggled on and off during model inference. Extensive experiments have been conducted to verify the effectiveness and efficiency of PSBD, which achieves state-of-the-art results among mainstream detection methods. The code is available at \url{https://github.com/WL-619/PSBD}.
\end{abstract}

%% file: sec/1_introduction.tex
\section{Introduction}
\label{sec:intro}
The proliferation of Deep Neural Networks (DNNs) has heralded a new era in artificial intelligence, driving progress across diverse sectors, including computer vision, autonomous driving and healthcare personalization \cite{li2024controlnet++,qin2023freeseg,muhammad2020deep,subramanian2022digital}. Yet, as their application scope broadens, DNNs have become increasingly vulnerable from a security standpoint. One of the most notable threats in this arena is the rise of backdoor attacks \cite{gu2017badnets,gan2022triggerless,nguyen2021wanet,chou2023backdoor}. These attacks involve the surreptitious insertion of altered samples into training data, enabling attackers to subtly manipulate a DNN's output, leading to incorrect predictions under certain triggers. The implications of such vulnerabilities are especially severe in contexts demanding high security, as they can lead to catastrophic outcomes \cite{bai2023physics,chen2022adversarial}. 

\begin{figure}[t]
\centering
\includegraphics[width=0.48\textwidth]{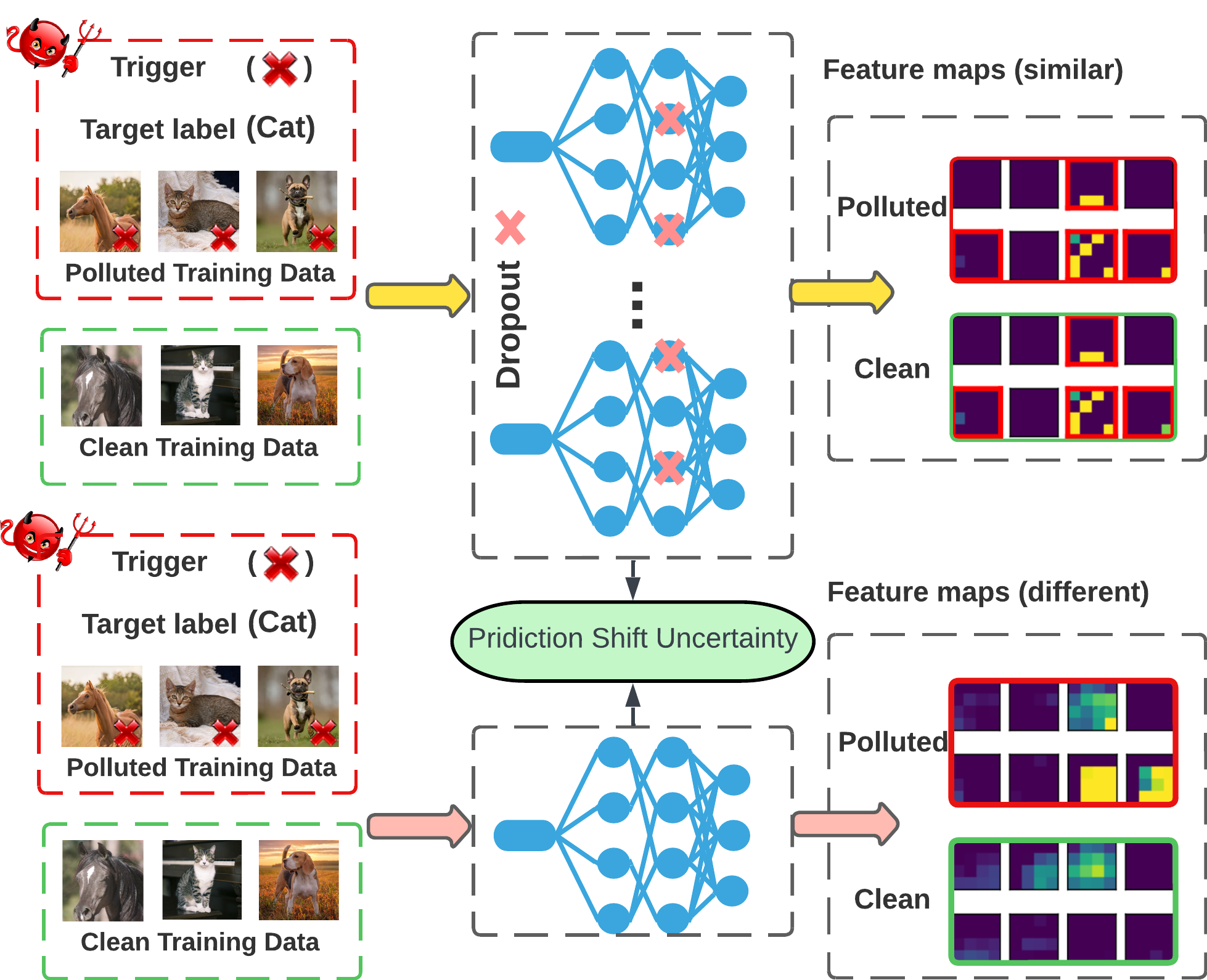}
\caption{A simple conceptual diagram of the Prediction Shift Backdoor Detection (PSBD) framework. The introduction of dropout during the inference stage induces a neuron bias effect in the model, causing the final feature maps of clean data and backdoor data to become highly similar, ultimately leading to the occurrence of the Prediction Shift phenomenon, which serves as a basis for detecting backdoor training data.}
\label{fig:PSBD}
\end{figure}

Notwithstanding an increased recognition of these risks, there are many different type of defense strategies currently, such as backdoor model reconstruction \cite{zhao2020bridging,borgnia2021strong, huang2022backdoor, pal2023towards,lu2024purification}, backdoor model detection \cite{ wang2020practical, shen2021backdoor, xu2021detecting, cai2022randomized}, and poison suppression \cite{li2021anti}. However, the majority of these methods primarily focus on determining whether trained models contain backdoors or on mitigating the impact of such vulnerabilities. In contrast, there is a noticeable shortage of advanced and efficient approaches that can proactively identify backdoor training samples at the initial stages. Current research in identifying backdoor training data often suffers from either a low true positive rate - indicating a low detection rate of backdoor data, or a high false positive rate - indicating a high error rate in identifying clean data \cite{guo2023scale, gao2019strip,huang2023distilling,chen2018detecting,pal2024backdoor}, as detailed in Table~\ref{tab: poison rate 0.1}. This issue is largely attributed to the focus of most existing research on data-level operations without utilizing the inherent properties of the models themselves.
To address this identified gap, we offer a new perspective - the model predictive uncertainty and propose a novel backdoor data detection approach named Prediction Shift Backdoor Detection (PSBD), which is inspired by an intriguing Prediction Shift (PS) phenomenon.

The PS phenomenon is observed when the predictions made by a poisoned model on clean data tend to deviate from the correct labels, moving towards other certain labels, especially when dropout is used during inference. Conversely, the predictions on backdoor data generated by both classical and advanced attacks remain relatively stable. This observation of PS led us to hypothesize the existence of a weights-neuron bias in DNN models, which we called the ``neuron bias'' effect, where some certain paths in the network become predisposed towards the specific class after training, and the backdoor samples have different paths compared with clean data. 
Figure~\ref{fig:PSBD} provides a brief overview of the PSBD framework and the neuron bias effect. Under normal conditions, the feature maps (the final convolutional layer) of clean and backdoor data exhibit significant differences. However, after applying the dropout, the neuron bias effect is induced within the model, causing the feature maps of clean and backdoor data to become strikingly similar, ultimately resulting in the occurrence of the PS phenomenon.
We also provided more detailed experimental explanations for verifying the neuron bias effect in section~\ref{subsection Neuron Bias}.

Driven by these insights, Prediction Shift Uncertainty (PSU) is designed to measure the strength of PS that computes the variability in prediction confidences when a model evaluates a sample with both enabled and disabled dropout. A lower PSU value indicates a higher likelihood of the sample being malicious. By calculating PSU, the PSBD approach can effectively segregate backdoor data from clean data using a small set of label-free clean validation data. 

Our approach represents a significant stride in backdoor data detection. Unlike existing methods, it focuses on the inherent uncertainty within the model, analyzing how dropout influences prediction probabilities of clean and backdoor samples. 
In summary, our main contributions are four-fold:
\begin{itemize}
    \item We reveal the PS phenomenon, showing that poisoned model predictions on clean data tend to deviate from ground true labels towards specific other labels when dropout is applied during inference, while backdoor data exhibits less PS. 
    \item We present a novel insight into the vulnerability of DNNs to backdoor attacks, linking it to the model's inherent predictive uncertainty. Our analysis delves into the impact of dropout on PS and introduces the concept of neuron bias within DNN models. 
    \item We propose the PSBD method, a simple yet powerful uncertainty-based approach for detecting backdoor training data, marking a significant advancement in the field. 
    \item We conduct extensive experiments across multiple benchmark datasets, rigorously evaluating our method under diverse attack scenarios and comparing it with a variety of defenses, demonstrating its effectiveness and robustness. 
\end{itemize}

%% file: sec/2_relative_work.tex
\section{Related Work}
\label{sec:related_work}

In this section, we explore the existing literature on backdoor attacks in neural networks and the defense strategies developed to counter them.

\paragraph{Backdoor Attacks.} Backdoor attacks are particularly dangerous, injecting triggers into a target model that cause it to misclassify inputs containing these triggers while operating normally on unaltered samples \cite{gu2017badnets, liu2018trojaning, chen2017targeted}. Initial approaches to backdoor attacks, such as BadNets \cite{gu2017badnets} and Blend attacks \cite{chen2017targeted}, involved embedding obvious trigger patterns like square patches into the input data. These methods evolved into more covert techniques, like clean-label attacks \cite{turner2019label}, which subtly poison samples of the target class using adversarial methods without obvious label changes, enhancing their stealthiness. Recent advancements have led to even more refined attacks, like WaNet \cite{nguyen2021wanet}, which introduces triggers that are specific to individual samples.

\paragraph{Backdoor Defenses.} Researchers have developed various defenses against backdoor attacks. These include efforts for backdoor trigger recovery \cite{wang2019neural, guo2019tabor, liu2022complex, xiang2022post, hu2021trigger}, which focus on identifying and reverse-engineering the attacker's trigger, and strategies for backdoor model reconstruction \cite{borgnia2021strong, huang2022backdoor, pal2023towards}, aimed at purging the backdoor model of its malicious elements. Methods for model detection \cite{kolouri2020universal, wang2020practical, shen2021backdoor, xu2021detecting} are employed to ascertain whether a model has been tainted with backdoor samples.  
{Backdoor sample detection evaluates whether a given sample triggers backdoor behavior in a model. Spectral Signatures (SS) \cite{tran2018spectral} employs deep feature statistics to differentiate between clean and backdoor samples, but its robustness weakens with varying poisoning rates \cite{hayase2021spectre}. STRIP \cite{gao2019strip} blends potentially backdoored samples with a small subset of clean samples and then using the entropy of the predictions for detection. Scale-up (SCP) \cite{guo2023scale} identifies and filters malicious testing samples by analyzing their prediction consistency during pixel-wise amplification. These methods primarily concentrate on altering input data, uncovering input masks, or distinguishing the feature representations of backdoor and benign samples.}

Nevertheless, these methods consider varying inputs and often experience low detection rates of backdoor training data or high error rates in identifying clean training data, as both clean and backdoor features can either remain intact or disappear when inputs are scaled. Our paper highlights the shortcomings of relying on input data variability and introduces a novel detection method that leverages model-level uncertainty, thereby surpassing the performance of methods based on input uncertainty. 

%% file: sec/3_preliminary.tex
\section{Preliminaries}
\label{preliminaries}

\subsection{Backdoor Attacks and Our Objective} 
Backdoor attacks in machine learning involve embedding a covert behavior into a neural network during training. This is typically done by poisoning the training dataset $\train$ with a set of malicious examples $\trainb$, such that the poisoned training dataset becomes $\trainc \cup \trainb$, where $\trainc$ represents the clean part of the training dataset. The objective function for training a model with a poisoned training dataset can be represented as $\min_{\boldsymbol{\theta}} \mathcal{L}(\trainc \cup \trainb;\boldsymbol{\theta})$, where $\boldsymbol{\theta}$ denotes the model parameters and $\mathcal{L}$ is the loss function. The model behaves normally on standard inputs but produces specific, attacker-chosen target label $\yt$ when a particular trigger is present. Such vulnerabilities pose a serious risk, especially in applications where model integrity is critical. Our objective is to maximize the detection of backdoor instances in $\trainb$ while minimizing the instances in $\trainc$ falsely identified as backdoor data.

\subsection{Threat Model} 

In our framework, we consider distinct capabilities and objectives for the attacker and defender within a black-box context. These roles are outlined as follows:

\paragraph{Attacker's Capabilities and Objectives.} The attacker has the ability to poison the training dataset but lacks insight into the training process itself. The primary objective is to manipulate the training data so that the model being trained exhibits erroneous behavior during testing when a specific trigger is present in the input, while maintaining standard performance on benign inputs.

\paragraph{Defender's Capabilities and Objectives.}
\label{defender's goal}
The primary goal of the defender is to ascertain which training data samples have been compromised by backdoor poisoning. In this scenario, the defender has full control over the training process. \textbf{Given a suspicious poisoned dataset, the defender is allowed to freely use it to train the model, adopting any model architecture and training strategies}. The defender lacks prior information regarding several key aspects: the existence of backdoor samples within the dataset, the proportion of these poisoned samples, the nature of the attack (including the trigger pattern and target label), and the specific class from which the backdoor samples originate. 
Additionally, we also assume that the defender possesses a limited set of extra label-free clean validation data, and it is also prevalent in many prior works that study backdoor defenses \cite{liu2018trojaning,li2021neural, guo2023scale}.

\subsection{Dropout Layers in Neural Networks} 

Dropout is a regularization technique that mitigates overfitting in neural networks. It randomly deactivates a subset of neurons during training, which can be mathematically described as $\mathbf{h}' = \mathbf{h} \odot \mathbf{m}$, where $\mathbf{h}$ is the output vector of a layer, $\mathbf{m}$ is a binary mask vector where each element is independently drawn from a Bernoulli distribution with probability $p$ (dubbed dropout rate), and $\odot$ denotes element-wise multiplication. During training, the expected output of a neuron is scaled by $p$, as only a fraction of the neurons are active. 
In many practical implementations, all neurons are active during inference, but their outputs are scaled by $p$ to account for the larger active network, ensuring consistency between the training and inference phases. In our study, inference-phase dropout is implemented.

%% file: sec/4_method.tex
\section{Method}

In this section, we offer a new perspective on the inherent predictive uncertainty within the model for the vulnerability of DNNs to backdoor attacks. We begin with two pilot studies to explore the predictive uncertainty of the model on the clean data and backdoor data. Then, we present our Prediction Shift Backdoor Detection (PSBD) method.

\subsection{A Spark of Inspiration: MC-Dropout Predictive Uncertainty}
\label{pilot study 1}
The model uncertainty is a metric that measures the extent to which the model's predictions can be trusted, and can be understood as what a model does not know. One is mainly interested in the model uncertainty that is propagated onto a prediction, the so-called predictive uncertainty \cite{gawlikowski2021survey}. 
Previous work has indicated that backdoor data 
 contains robust features and is potentially easier to learn compared to clean data  
 \cite{li2021anti,wang2020practical}.
 Therefore, we expect that the model should exhibit lower uncertainty towards backdoor data compared to clean data. 

\begin{figure}[!h]
    \centering
     \begin{subfigure}[h]{0.23\textwidth}
         \centering
         \includegraphics[width=\textwidth]{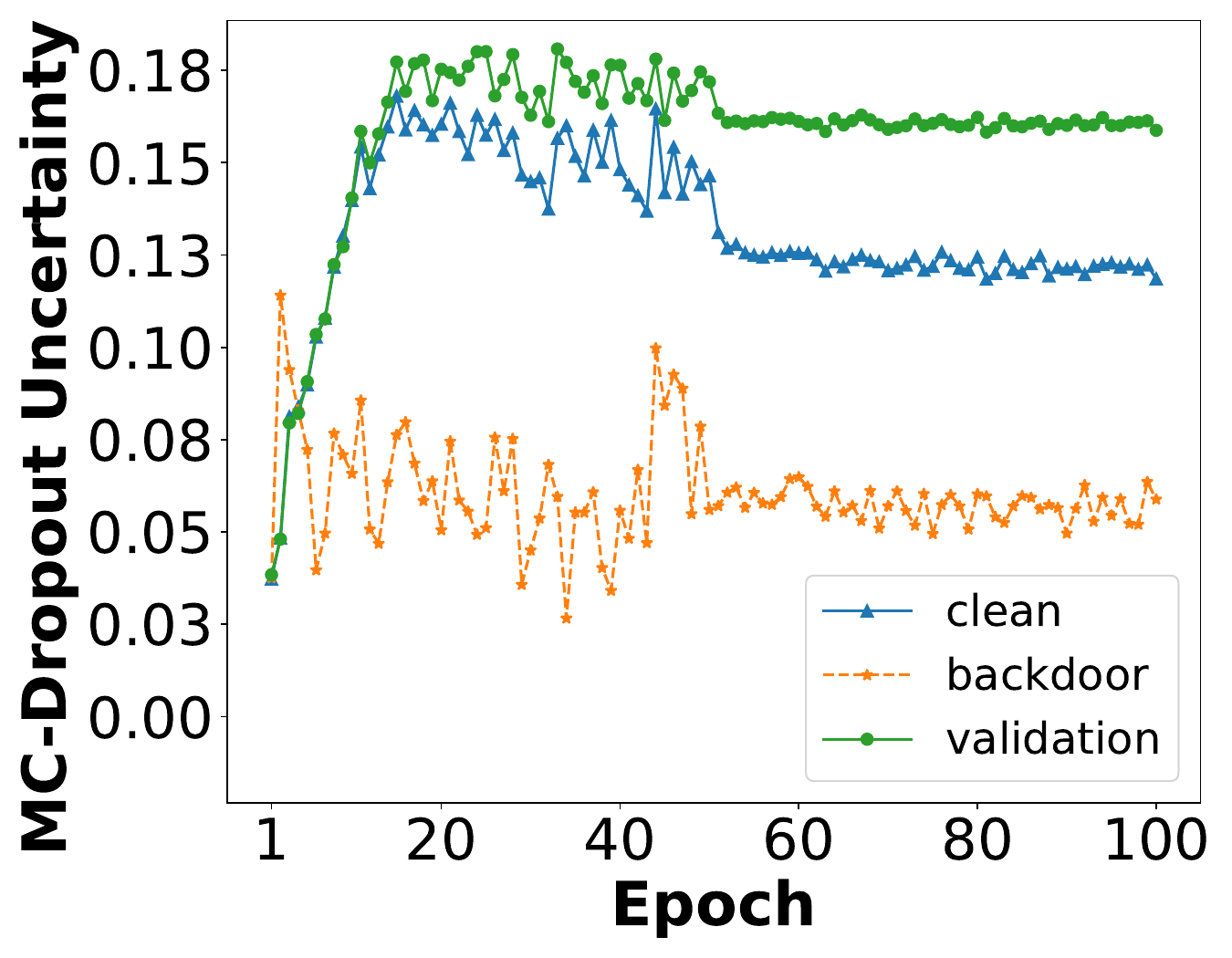}
         \caption{BadNets}
         \label{fig:pilot study 1 badnets}
     \end{subfigure}
     \begin{subfigure}[h]{0.23\textwidth}
         \centering
         \includegraphics[width=\textwidth]{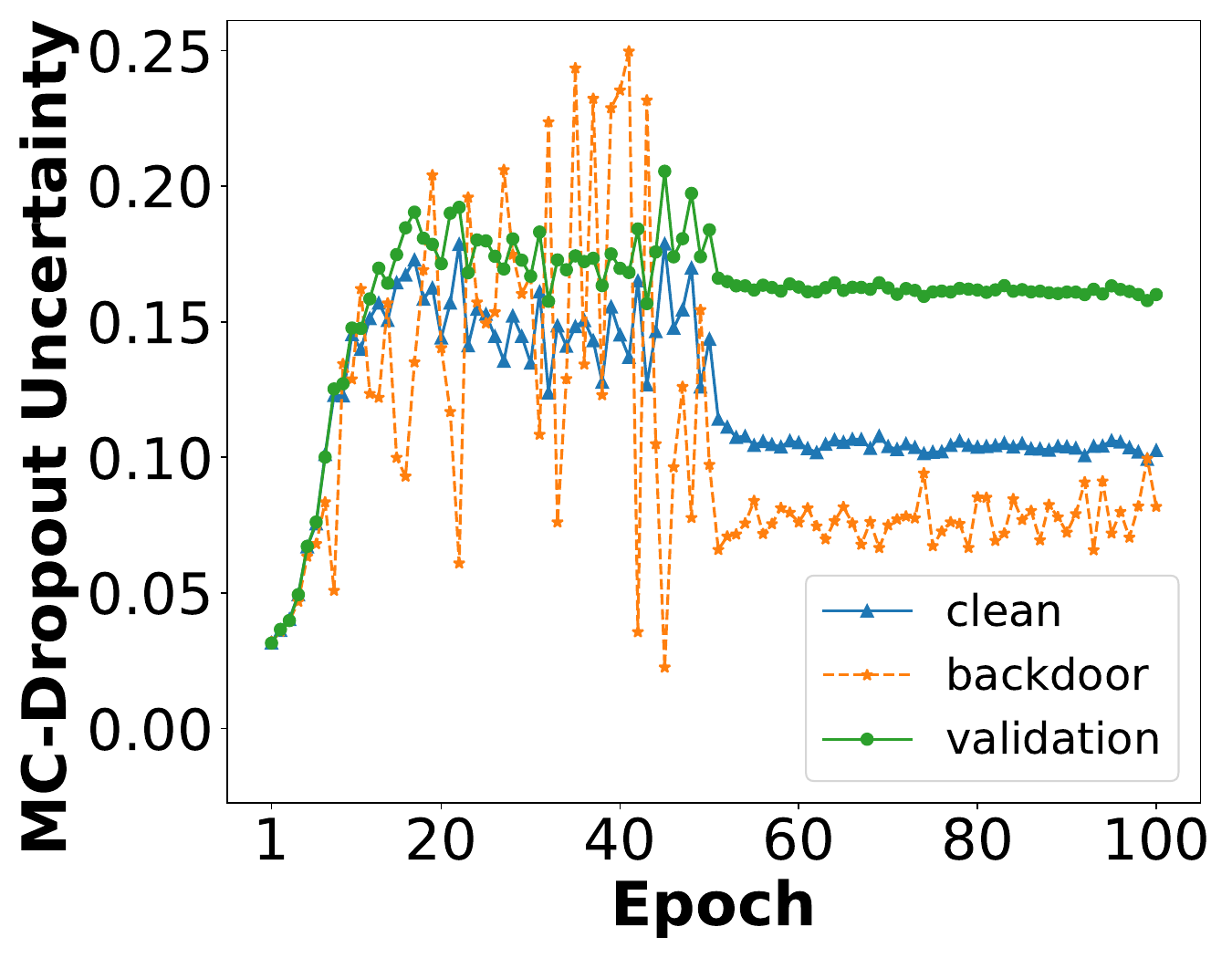}
         \caption{WaNet}
         \label{fig:pilot study 1 wanet}
     \end{subfigure}
     \vspace{-2mm}
    \caption{The average MC-Dropout uncertainty of clean training data, backdoor training data, and clean validation data under poisoned models.}

\end{figure}
We use a widespread model predictive uncertainty approximation method - Monte Carlo Dropout (MC-Dropout) \cite{gal2016dropout} to explore the model predictive uncertainty of the three types of data - the clean training data, the backdoor training data, and the clean validation data. 
MC-Dropout activates dropout during inference, allowing multiple forward passes. The model's final predicted confidence is the average of these passes, while the standard deviation of the highest confidence class indicates predictive uncertainty.

\begin{figure*}[ht]
    \centering
     \begin{subfigure}[h]{0.3\textwidth}
         \centering
         \includegraphics[width=\textwidth]{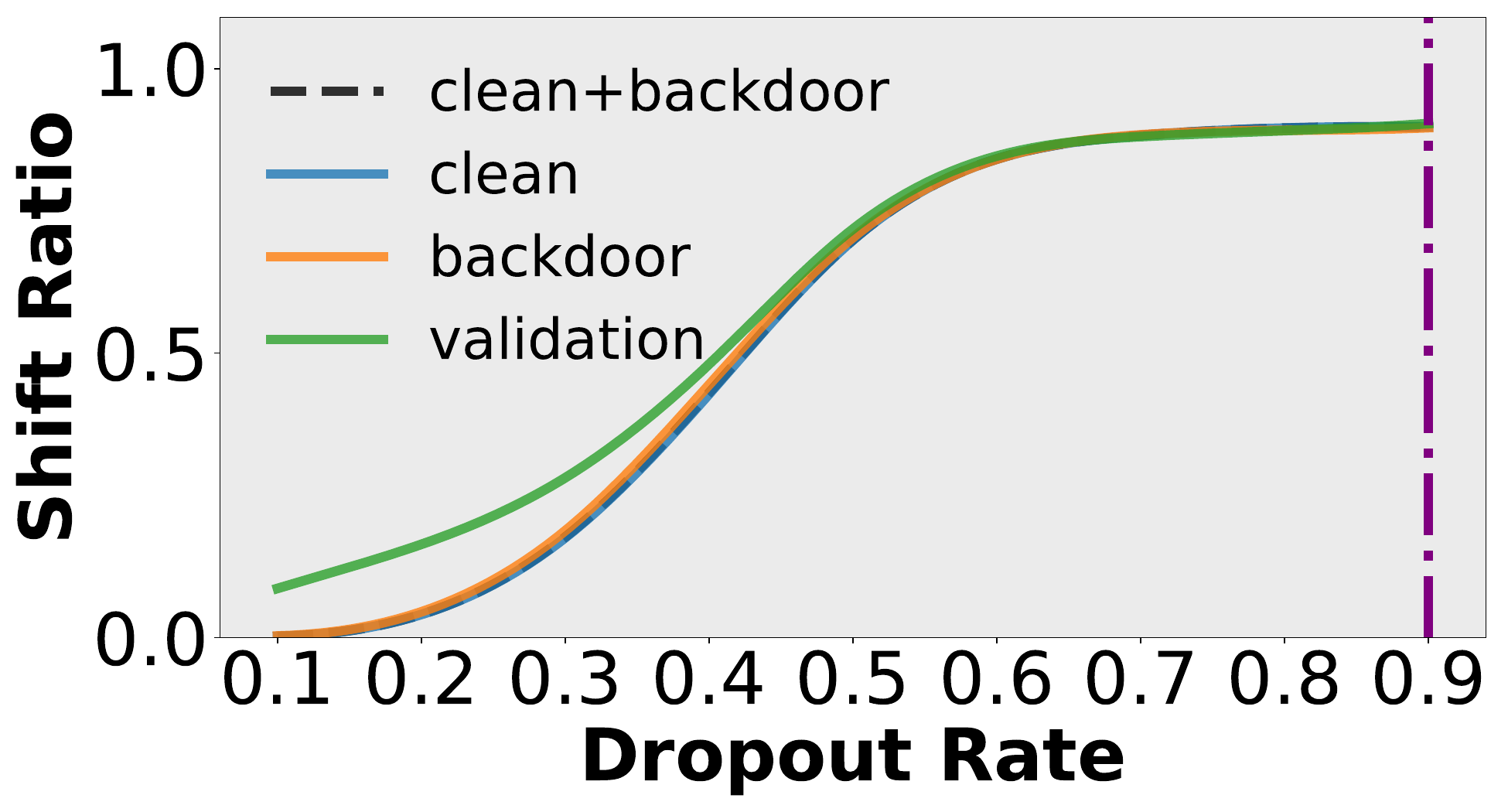}
     \end{subfigure}
     \hfill
     \begin{subfigure}[h]{0.3\textwidth}
         \centering
         \includegraphics[width=\textwidth]{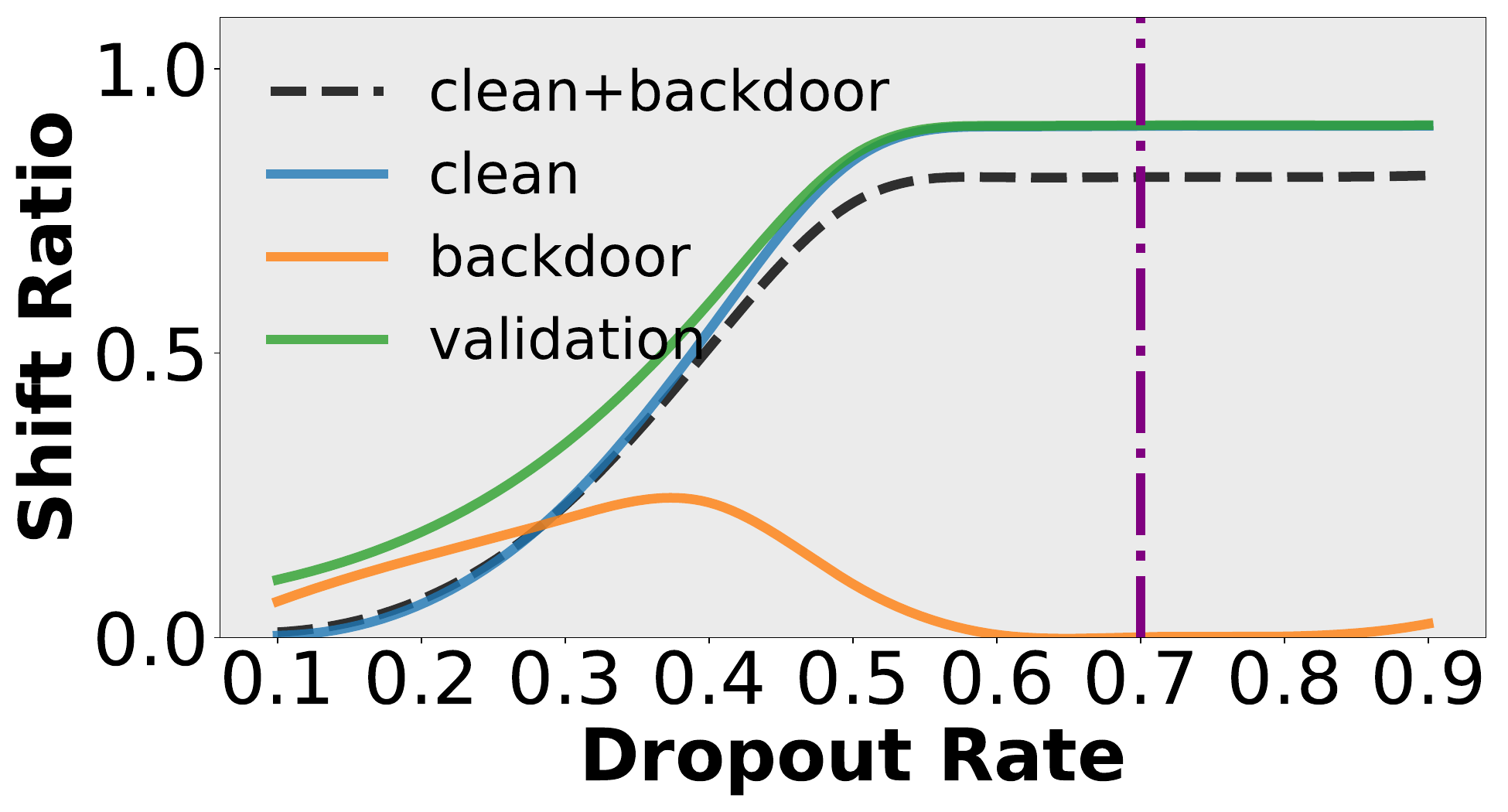}
     \end{subfigure}
     \hfill
     \begin{subfigure}[h]{0.3\textwidth}
         \centering
         \includegraphics[width=\textwidth]{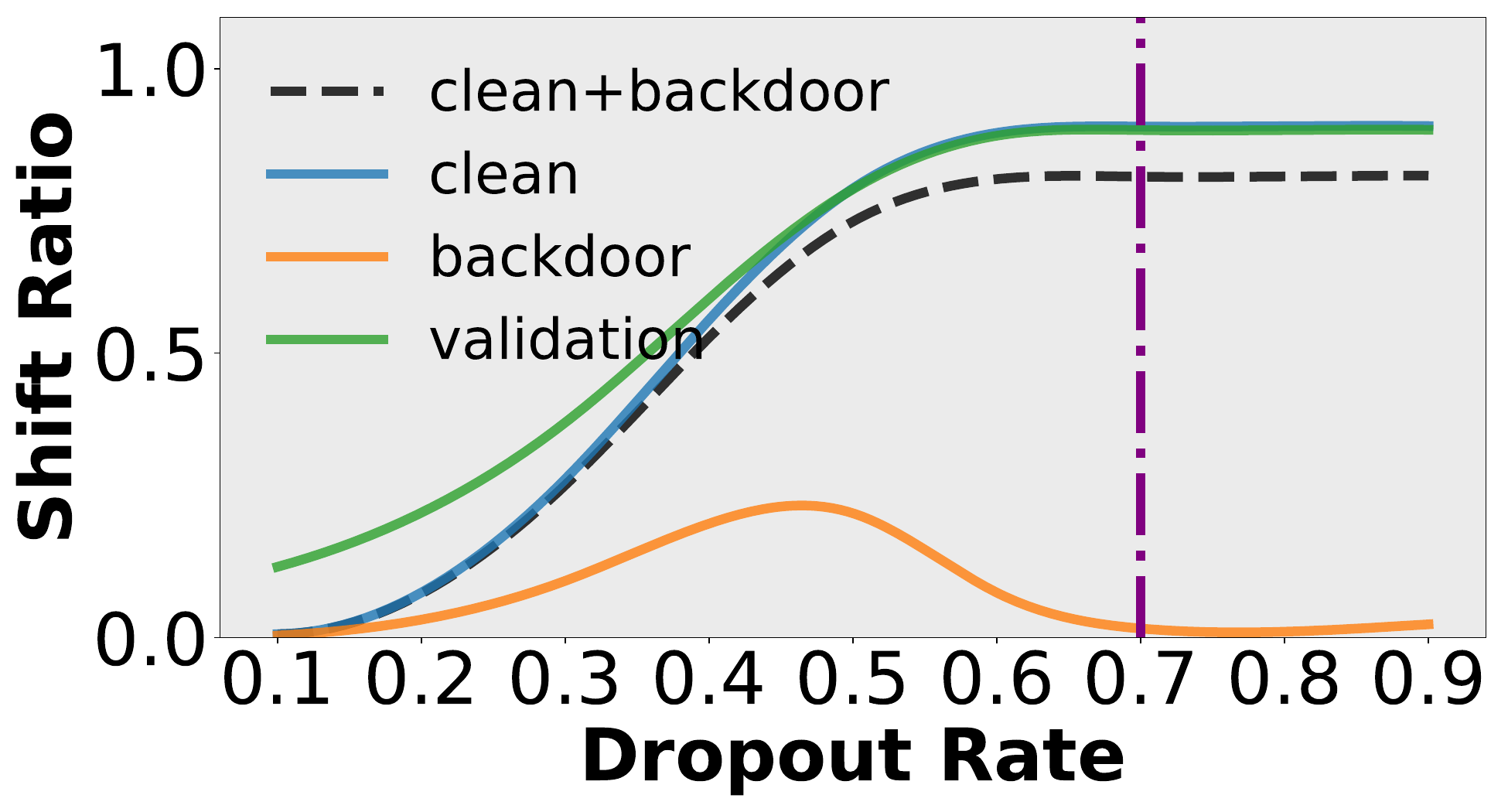}
     \end{subfigure}
     \vfill
     \begin{subfigure}[h]{0.3\textwidth}
         \centering
         \includegraphics[width=\textwidth]{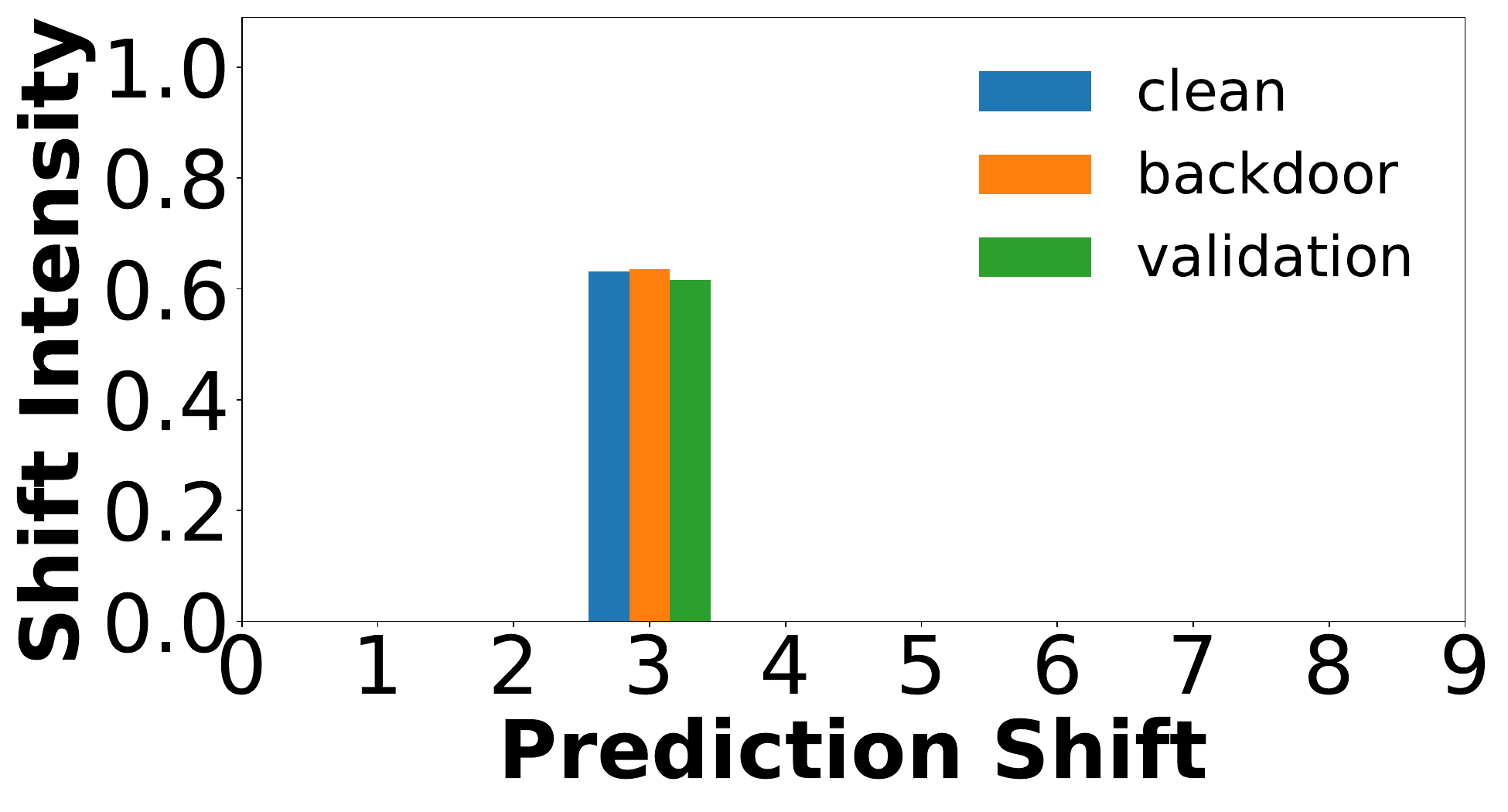}
         \caption{Benign Model}
         \label{fig:ps benign}
     \end{subfigure}
     \hfill
     \begin{subfigure}[h]{0.3\textwidth}
         \centering
         \includegraphics[width=\textwidth]{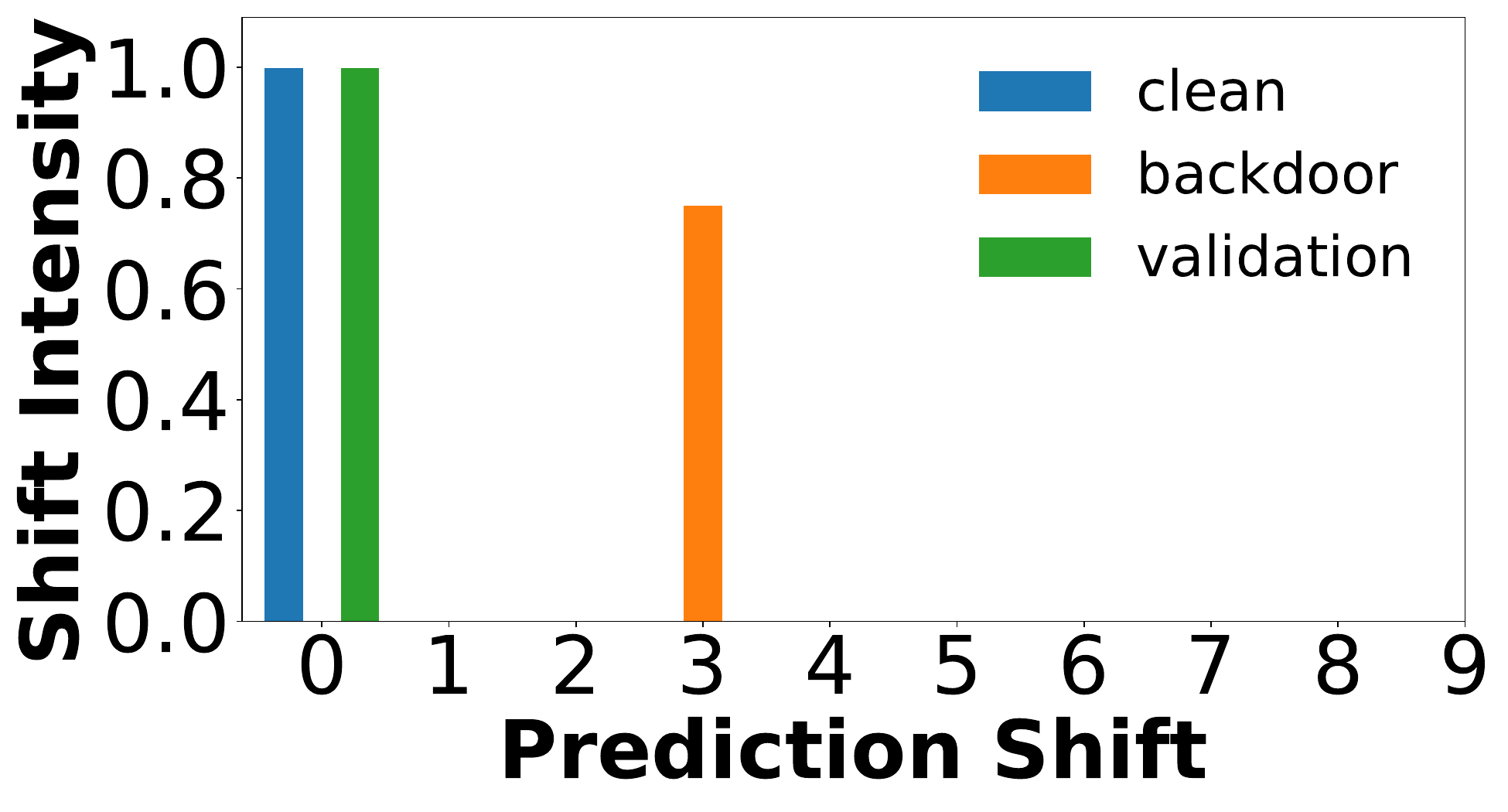}
         \caption{BadNets}
         \label{fig:ps badnets}
     \end{subfigure}
     \hfill
     \begin{subfigure}[h]{0.3\textwidth}
         \centering
         \includegraphics[width=\textwidth]{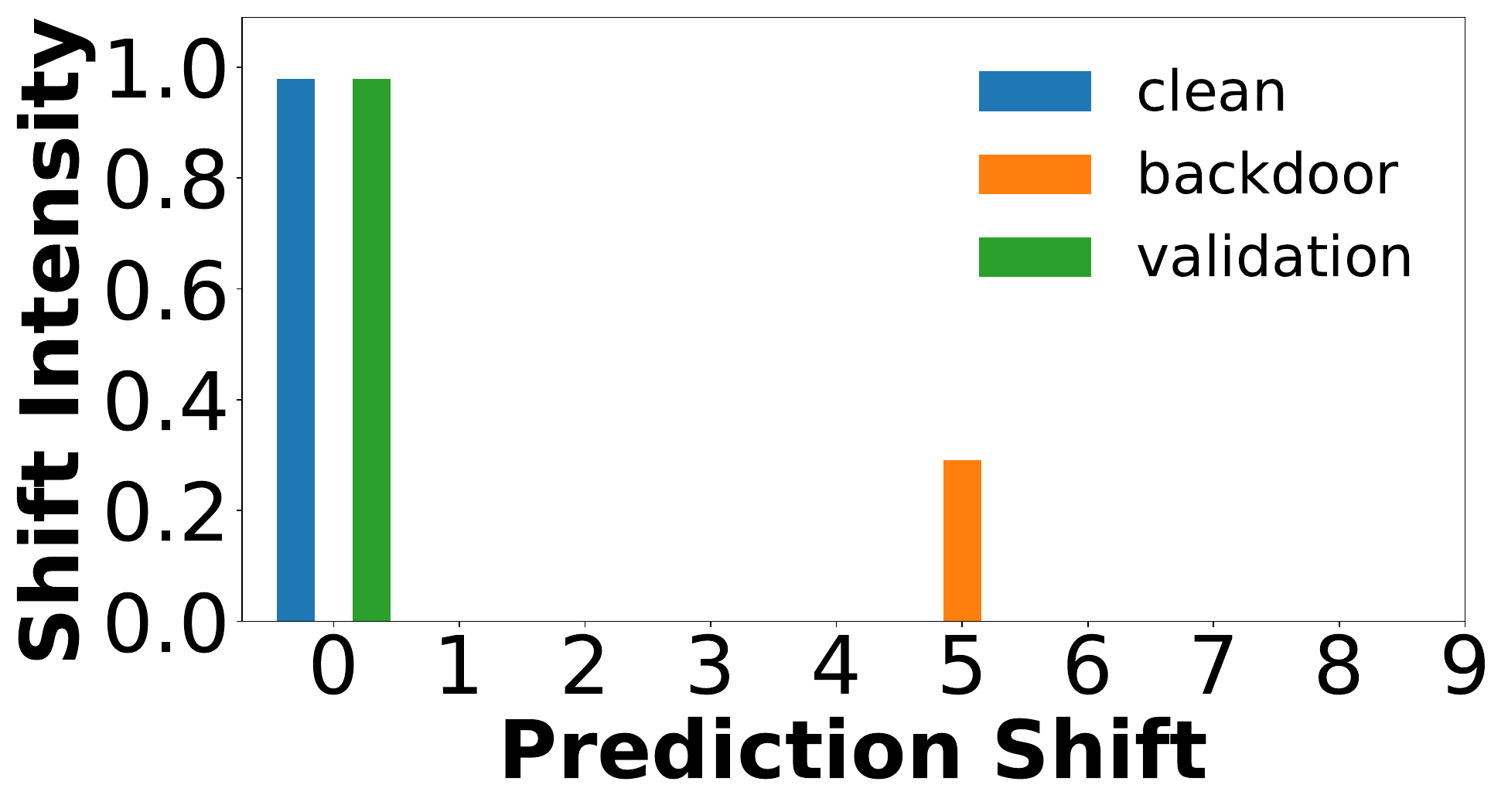}
         \caption{WaNet}
         \label{fig:ps wanet}
     \end{subfigure}
     \vspace{-3mm}
    \caption{The above row shows the shift ratio curves for the benign model, BadNets model, and WaNet model, respectively. The below row represents the prediction shift intensity for samples exhibiting PS phenomenon at the chosen $p$. The purple vertical dash line corresponds to the selected $p$ using our adaptive selection strategy.}
\end{figure*}

When applying MC-Dropout, clean validation data should show the highest uncertainty, as it's unseen during training. Clean training data should follow closely, with a smaller gap between them than between clean and backdoor training data. Backdoor data should have the lowest uncertainty, as it's easier for the model to learn. If these patterns hold, backdoor training data can be treated as outliers, enabling their detection using outlier methods.

\paragraph{Settings.}
\label{pilot study 1 settings}
We adopt BadNets and WaNet as examples for our discussion. We conduct experiments on the CIFAR-10 dataset \cite{krizhevsky2009learning} and ResNet-18 \cite{He_2016_CVPR}, trained for 100 epochs. For both attacks, we set the poisoning ratio to 10\%, i.e. replaced 10\% of total training data with malicious backdoor training data. Without sacrificing generality, the target class $y_t$ of backdoor data is class 0 in our all examples.
We randomly select the clean validation dataset from the original CIFAR-10 test set, and its size is 5\% of the total size of the training set. 
We calculate the average MC-Dropout uncertainty of three types of data with models obtained from all 100 epochs to compare the difference in their uncertainty. 

\paragraph{Results.}
Figure~\ref{fig:pilot study 1 badnets} shows that, over 100 epochs, the average uncertainty of backdoor training data under BadNets is significantly lower than that of clean training and validation data, with this difference stabilizing in later training stages. This aligns with our expectation that backdoor examples have smaller uncertainty. However, in Figure~\ref{fig:pilot study 1 wanet}, the uncertainty of backdoor data under WaNet sometimes matches or exceeds that of clean data. Additional results are available in Appendix~\ref{pilot 1 appendix}. We also tested a variant of the MC-Dropout method, which showed some improvement in detection but still failed in certain cases (details in Appendix~\ref{pilot 2 appendix}). These findings suggest that using uncertainty based on standard deviation may be insufficient for detecting backdoor data across different attack scenarios. Additionally, determining the appropriate dropout rate $p$ is challenging without detailed knowledge of backdoor attacks. 

\subsection{The Enlightening Eureka Moment: Prediction Shift Phenomenon}
\label{sec:ps and nb}
Contrary to the indications from pilot study, relying solely on the simple MC-Dropout predictive uncertainty proves insufficient for distinguishing between clean and backdoor data. Although frustrating, we can still observe that the model's mapping from trigger to target label in backdoor data is more salient and robust compared to general image features. Informed by these preliminary findings, we delved further into the impact of employing dropout during the model inference phase on the model's behavior.

\paragraph{Prediction Shift.}
\label{PS}
To delve deeper into how dropout affects the predictive uncertainty of the model, we examined how enabling dropout during the model's forward process alters the model's classifications and  prediction confidence. We define Prediction Shift (PS) as the phenomenon where the class predicted by the model changes before and after dropout is enabled, for samples $\mathbf{x}$ in the dataset $\mathcal D$. The shift ratio $\sigma$ represents the frequency of PS occurring in all forward inferences with dropout activated across the dataset $\mathcal D$, i.e.,

\begin{align}
    \begin{split}
    \phi_{PS}(\mathbf{x}) &= \mathbb{I}\left(\mathcal{Y}(\mathbf{x};{\boldsymbol \theta})\neq\mathcal{Y}(\mathbf{x};{\boldsymbol \theta^{'}})\right),\\ \sigma(\mathcal D) &= \frac{1}{k|\mathcal D|}\sum\limits_{\mathbf{x}\in \mathcal D} \phi_{PS}(\mathbf{x})
    \end{split}
    \label{PS definition}
\end{align}

where $\mathcal D$ represents an arbitrary dataset, which could encompass the entire training set or a specific subset, such as one class of data or a poisoned/clean training set; $\mathcal{Y}(\mathbf{x};{\boldsymbol \theta})$ represents the predicted class of the model ${\boldsymbol \theta}$ without dropout for input $\mathbf{x}$ and $\mathcal{Y}(\mathbf{x};{\boldsymbol \theta^{'}})$ corresponds to the predicted class of model ${\boldsymbol \theta^{'}}$ with dropout in forward inference stage; $\phi_{PS}(\cdot)$ denotes the PS function;  $k$ denotes the number of forward iterations performed with dropout. 

\paragraph{Settings.} 
\label{ps setting}
Firstly, the model is trained on the poisoned training set following the standard training procedure, which excludes the use of dropout, data augmentation, and data normalization. After that, we apply dropout without using data augmentation and data normalization during model inference. This allows us to completely control the model's ability to extract data features, thereby influencing the uncertainty of its predictions by adjusting the dropout rate $p$. We perform forward inference $k=3$ times and record the value of PS in the three types of data. 
Specifically, dropout layers are applied after each residual connection in the residual basic block, before the activation function, as this can significantly influence the model's predictions with the dropout.

\begin{figure*}[t]
    \centering
     \begin{subfigure}[h]{0.24\textwidth}
         \centering
         \includegraphics[width=\textwidth]{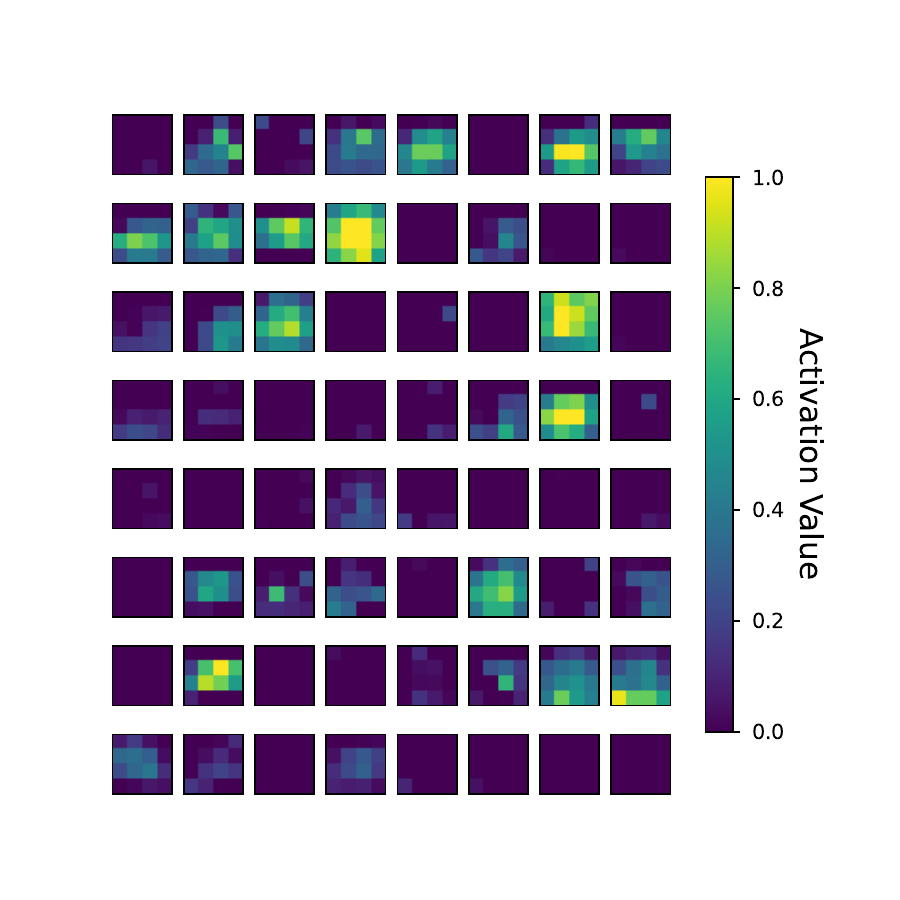}
         \caption{Clean image w/o dropout}
         \label{fig:feature clean w/o drop}
     \end{subfigure}
     \begin{subfigure}[h]{0.24\textwidth}
         \centering
         \includegraphics[width=\textwidth]{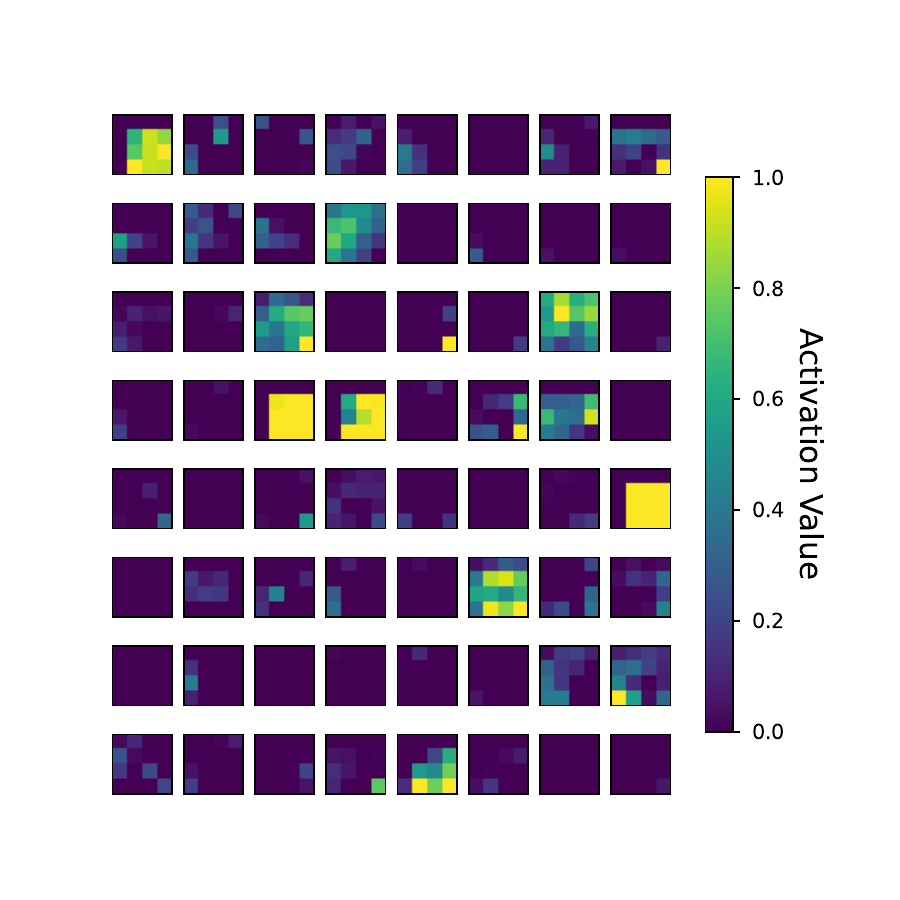}
         \caption{{Backdoor image w/o dropout}}
         \label{fig:feature backdoor w/o drop}
     \end{subfigure}
     \begin{subfigure}[h]{0.24\textwidth}
         \centering
         \includegraphics[width=\textwidth]{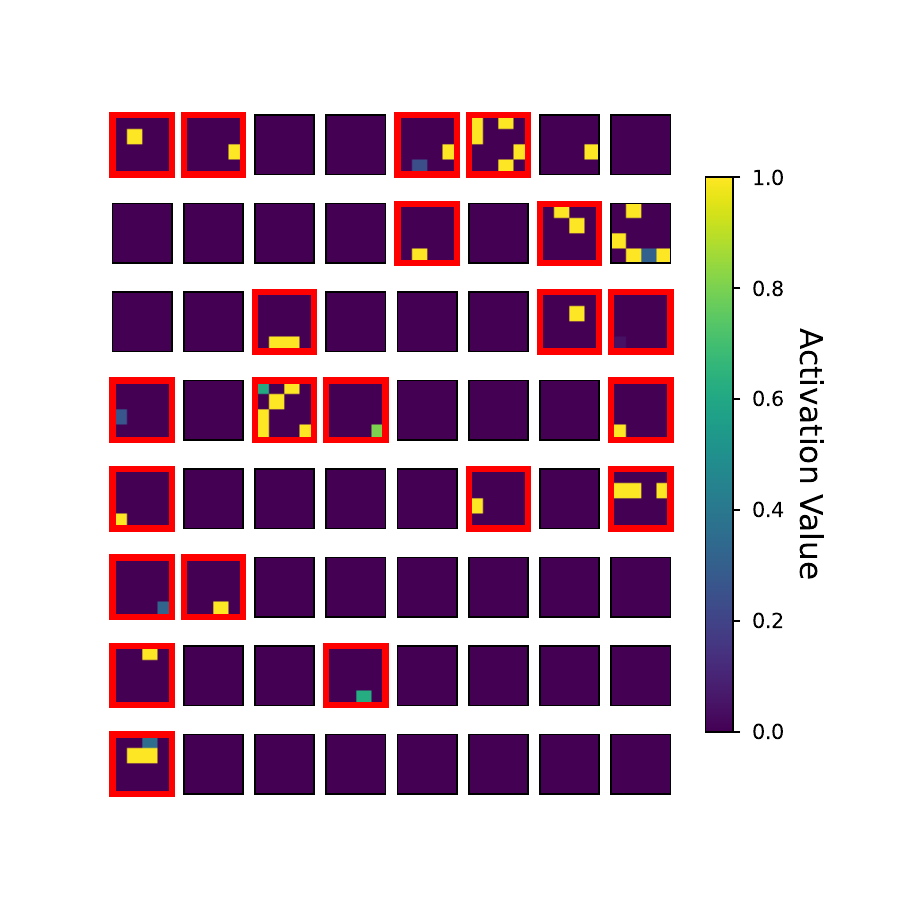}
         \caption{{Clean image w/ dropout}}
         \label{fig:feature clean w/ drop}
     \end{subfigure}
     \begin{subfigure}[h]{0.24\textwidth}
         \centering
         \includegraphics[width=\textwidth]{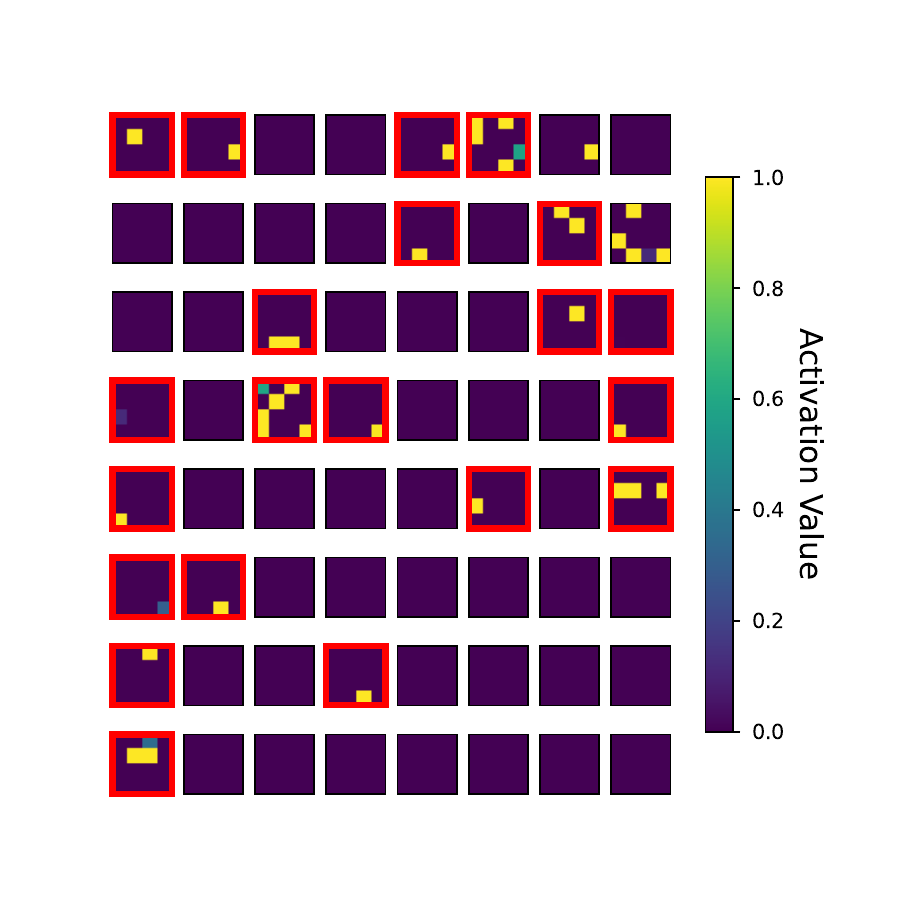}
         \caption{{Backdoor image w/ dropout}}
         \label{fig:feature backdoor w/ drop}
     \end{subfigure}
     \vspace{-3mm}
    \caption{The first 64 feature maps out of the 512 extracted by the top layer of the model. The red boxes represent the feature map values are non-zero and the difference between each activation value in the clean and backdoor feature maps is no greater than 1. The features of clean and backdoor image become almost identical with dropout, verifying the existence of neuron bias effect.}
    \label{fig:activations w/ and w/o dropout}
\end{figure*}

\paragraph{Results.}
In the bottom row of Figure~\ref{fig:ps benign}, about 60\% of clean training data that experience Prediction Shift (PS) under the benign model shift to class 3. The x-axis shows the shifted labels, while the y-axis represents the intensity of the shift—the proportion of times a sample was predicted to a particular class during PS. This pattern is also seen in backdoor training and clean validation data, suggesting that PS is a universal characteristic of DNNs. In the top row of Figure~\ref{fig:ps benign}, both clean and backdoor training data exhibit similar shift ratio trends, supporting the conclusion from Section~\ref{pilot study 1} that the benign model treats backdoor data as perturbed clean data, classifying them mainly based on natural image features.

As illustrated in Figure~\ref{fig:ps badnets} and~\ref{fig:ps wanet}, in the BadNets and WaNet scenarios, we observe that the shift ratio curve for clean data still follows an increasing trend as $p$ increases, eventually stabilizing. However, when $p$ reaches a certain special value, the $\sigma$ for backdoor data approaches 0, while the $\sigma$ for clean data reaches a relatively high value (around 0.8). The most important thing is, among the samples experiencing PS, almost all clean data shifts to the target class $y_t$ (class 0 in our experiments). The same phenomenon has been observed in other attack scenarios on CIFAR-10 as well. This indicates that training with backdoor samples enhances the PS phenomenon of clean data while suppresses that of backdoor data. 
This is likely due to significant differences in the internal behavior of the model towards clean data and backdoor data under appropriate dropout $p$. 
Other poisoned model's results can be found in the Appendix~\ref{PS on more poison appendix}.

In addition, we continued to observe similar patterns on the expanded and intricate Tiny ImageNet dataset. However, the shift classes of the clean training data and the clean validation data exhibit a predominant inclination towards a certain class rather than the target class. Nevertheless, there is still a certain proportion that exhibits a bias towards the target class. Despite assuming that the defender can freely choose the model architecture, we also conducted experiments using the VGG\cite{simonyan2014very} to demonstrate that our method is not dependent on any specific model architecture. For all results, please refer to the Appendix~\ref{PS on tiny appendix} and~\ref{PS on vgg appendix}.

\paragraph{``Neuron Bias'' - An Explanation to Prediction Shift.}
\label{subsection Neuron Bias}
We posit that the PS phenomenon arises from the neuron bias effect in the network during training, where neurons become predisposed to features highly representative of certain classes. This bias intensifies as the network establishes strong associations between specific data features and particular classes, especially in the case of backdoor features linked to the target class. In the absence of dropout, backdoored models typically predict the correct class for clean data, as they possess sufficient features to make accurate predictions. However, under dropout conditions, many key distinguishing features in clean data are discarded. Consequently, the model relies more heavily on the neuron bias established during training, leading it to classify clean data to the label associated with this bias. In contrast, the model learns backdoor data patterns more effectively and rapidly, resulting in a more stable and pronounced neuron bias. This enhanced bias allows the model to correctly classify backdoor data even when some features are omitted due to dropout.

{To validate our hypothesis, we analyzed the features extracted by the BadNets model from both clean image and its corresponding backdoor version, comparing the results with and without the application of dropout. We presented the first 64 feature maps out of the 512 extracted by the top layer of the model. As illustrated in Figure~\ref{fig:feature clean w/o drop} and~\ref{fig:feature backdoor w/o drop}, without the dropout, the features of clean and backdoor version exhibit minimal similarity, which partly explains the model's distinct behavior towards these two types of data. However, under an appropriate dropout rate, Figure~\ref{fig:feature clean w/ drop} and~\ref{fig:feature backdoor w/ drop} clearly shows that the features of clean and backdoor version become almost identical with dropout. The red boxes in the figure highlight regions where the feature map values are non-zero and the difference between each activation value in the corresponding feature maps is no greater than 1. This finding successfully confirms the validity of our neuron bias effect hypothesis. Detailed results are available in Appendix~\ref{features w/ and w/o dropout}.}

\subsection{From Insight to Innovation: Prediction Shift Backdoor Detection}
\label{sec:PSU}
Even with dropout enabled, the predicted labels for some clean data remain unchanged before and after applying dropout, although their prediction confidence changes significantly. To quantify the change in prediction confidence rather than the change in labels as defined in Equation~\eqref{PS definition}, we introduce a new and more fine-grained measure of predictive uncertainty, Prediction Shift Uncertainty (PSU). PSU computes the difference between the predicted class confidence without dropout and the average predicted class confidence across $k$ dropout inferences to quantify the intensity of PS:
\begin{align}
    \begin{split}
        \phi_{PSU}(\mathbf{x}) &= {P}_{c}(\mathbf{x};{\boldsymbol \theta})-\frac{1}{k}\sum\limits_{i=1}^{k} {P}_{c}(\mathbf{x};p,{\boldsymbol \theta_{i}^{'}}), \\
        c &= \mathop{\arg\max}_{c\in \mathcal C} {P}(\mathbf{x};{\boldsymbol \theta})
    \end{split}
    \label{PSU definition}
\end{align}
where ${c}$ represents the class with the highest predicted confidence for data $\mathbf{x}$ without dropout during the inference stage; ${P}_{c}(\mathbf{x};{\boldsymbol \theta})$ represents the predicted confidence of class $c$ by the model without using dropout for input $\mathbf{x}$, and ${P}_{c}(\mathbf{x};p,{\boldsymbol \theta_{i}^{'}})$ corresponds to the confidence with dropout at the $i$th forward pass; ${\boldsymbol \theta}$ represents the origin model parameters; ${\boldsymbol \theta_{i}^{'}}$ represents the $i$th dropout model parameters across all $k$ inferences. Here $p$ is the dropout rate.

Similar to pilot studies, the optimal dropout rate $p$ is a crucial factor in the dropout-based uncertainty method and is challenging to determine without knowledge of backdoor attacks. A reasonable $p$ is selected when the PS of clean data 
achieves a relatively strong intensity, and that of backdoor data remains relatively weak. However, due to a lack of backdoor knowledge, we cannot directly compute the PS of backdoor data. Thus, based on the definition of $\sigma$ provided in Equation~\eqref{PS definition}, we propose an adaptive selection strategy for $p$. Specifically, we identify the $p$ where the $\sigma$ of clean validation data approach to a high value (0.8 in our experiments), while the difference between the $\sigma$ of the entire training data and that of the clean validation data reaches its maximum. 

\paragraph{Prediction Shift Backdoor Detection.}
As we mentioned above, clean data always shift from the origin prediction class to another specific class, while backdoor data often remain static. Consequently, the PSU of clean training data and clean validation data will be close and high, whereas the PSU of backdoor data will be small under an appropriate $p$. For suspicious data $\mathbf{x}$, it can be determined as malicious based on a defender-specified threshold $T$. If $PSU(\mathbf{x})< T$, it is classified as a backdoor sample. We set $T$ based on the close proximity of PSU values between clean training data and extra clean validation data. In other words, in the absence of knowledge regarding the backdoor attack, $T$ can be roughly regarded as the tolerable loss rate for clean training data. In all our experiments, $T$ is set to the 25th percentile PSU value of the whole clean validation data.
Furthermore, we found that using data augmentation in model training significantly outperforms the non-augmented training approach on Tiny ImageNet.
It indicates that the use of data augmentation can intensify neuron bias, especially when the model has a lack of generalization ability to recognize the more sophisticated features. Hence, we incorporate data augmentation during model training when the model's generality is lacking. The specific workflow of the prediction shift backdoor detection (PSBD) method is given as follows:
\begin{itemize}
    \item {First, we train the model using a standard supervised learning algorithm on the suspicious training dataset, employing common data augmentation techniques when the model lacks generalization ability.} 
    \item {Next, we select the dropout rate $p$ based on the adaptive selection strategy. Then, we select a late-stage model to calculate the PSU values for the suspicious training data and clean validation data, due to its enhanced data fitting capability and robust neuron bias paths.}
    \item {Finally, for suspicious data $\mathbf{x}$, we can determine it is malicious based on defender-specified threshold $T$. If $PSU(\mathbf{x})<T$, we view it as a backdoor sample. $T$ is set to the 25th percentile value of the PSU of clean validation data in all our experiments.}
\end{itemize}
\label{PSBD pipline}

%% file: sec/5_experiments.tex
\section{Experiments}
\label{sec:exp_results}

\subsection{Experiment Settings}
\label{exp settings}
\paragraph{Dataset and DNN Model.}
We conduct all experiments on the CIFAR-10 \cite{krizhevsky2009learning}, GTSRB \cite{stallkamp2011german} and Tiny ImageNet \cite{russakovsky2015imagenet} datasets using the ResNet-18 \cite{He_2016_CVPR} architecture. \textbf{Please note that a defender is free to choose any architecture, as the sole objective is to detect any potential backdoor data that may exist within the dataset.} We randomly select 5\% of the total quantity of whole poisoned training dataset from the original test sets as our extra clean validation data. Further details can be found in the Appendix~\ref{dataset detail appendix}.
\vspace{-3mm}
\paragraph{Backdoor Attack Settings.}
We evaluate our PSBD method against seven representative backdoor attacks, namely BadNets \cite{gu2017badnets}, Blend \cite{chen2017targeted}, TrojanNN\cite{liu2018trojaning}, Label-Consistent \cite{turner2019label}, WaNet \cite{nguyen2021wanet}, ISSBA \cite{li2021invisible} and Adaptive-Blend \cite{qi2022revisiting}.  
{We examined two scenarios with poisoning ratios of 5\% and 10\%. The main paper discusses the 10\% poisoning ratio in detail, while the results for the 5\% poisoning ratio are presented in Appendix~\ref{low pr appendix}.}
Without sacrificing generality, in all our experiments, the target class $y_t$ is set to class 0. Data augmentation during the model training was employed exclusively for Adaptive-Blend on CIFAR-10, GTSRB, and all experiments on Tiny ImageNet to achieve an attack success rate exceeding 85\%. 
More detailed settings are presented in Appendix~\ref{attack details appendix}.
{We also verify the robustness of PSBD against potential adaptive attacks in Appendix ~\ref{adaptive_attack appendix}.}

\vspace{-3mm}
\paragraph{Backdoor Detection Baseline.}
We compare PSBD with five classic and state-of-the-art backdoor data detection methods, namely Spectral Signature(SS) \cite{tran2018spectral}, Strip \cite{gao2019strip}, Spectre \cite{hayase2021spectre}, SCAN \cite{tang2021demon}, SCP \cite{guo2023scale} and CD-L \cite{huang2023distilling}. 
All six methods were implemented and evaluated on the CIFAR-10 dataset. For GTSRB and Tiny ImageNet datasets, SCAN was excluded due to its computationally intensive matrix eigenvalue computations, which significantly increased processing time.
We run 10 trials for each experiment of all methods and report the average results across all cases as the final result. We found that the variances are relatively small, so we ignored them. Please refer to the Appendix~\ref{baseline details appendix} for the implementation details.

\vspace{-3mm}
\paragraph{Metric.}
To assess the effectiveness of detection methods, we employ common classification metrics: True Positive Rate (TPR) and False Positive Rate (FPR). Our evaluation prioritizes achieving a high TPR to ensure effective identification of backdoor samples, while simultaneously maintaining a low FPR to minimize erroneous deletion of clean samples. Values inside brackets represent standard deviations (SD). 
{Moreover, the results for the area under receiver operating curve (AUROC) can be found in the Appendix~\ref{AUROC appendix}.}

\begin{table*}[!t]
\centering
\caption{
\footnotesize{
The performance (TPR/FPR) on CIFAR-10, GTSRB and Tiny ImageNet. We mark the \textbf{best result} in boldface while the value with underline denotes the \underline{second-best}. The \textcolor{lightgray}{failed cases} (i.e., TPR < 0.8) are marked in gray. Adaptive-Blend attack has a 1\%/1\%/2\% poisoning ratio on CIFAR-10/GTSRB/Tiny ImageNet, while other attacks have a 10\% poisoning ratio. OOT indicates that the method did not finish within the allocated time limit. 
}
}
\vspace{-2mm}
\label{tab: poison rate 0.1}
\resizebox{0.85\textwidth}{!}{
\begin{tabular}{c|c|c|c|c|c|c|c}
\toprule[1pt]
\midrule
\multirow{1}{*}{Defenses$\rightarrow$} & \multirow{2}{*}{PSBD (\textbf{Ours})} & \multirow{2}{*}{SS} & \multirow{2}{*}{Strip} & \multirow{2}{*}{Spectre} & \multirow{2}{*}{SCAN} & \multirow{2}{*}{SCP} & \multirow{2}{*}{CD-L} \\
\multirow{1}{*}{Attacks$\downarrow$} & & & & & & \\
\rowcolor{LightCyan}
\midrule
\multicolumn{8}{c}{\hspace{20mm}\textbf{CIFAR-10}\hspace{8mm}} \\
\midrule
Badnet & \underline{1.000/0.104} & \textcolor{lightgray}{0.389/0.512} & 1.000/0.113 & 0.953/0.450 & \textbf{1.000/0.009} & 1.000/0.205 & 
0.998/0.158\\ 
Blend  & \textbf{1.000/0.135} & \textcolor{lightgray}{0.438/0.507} & \underline{0.993/0.118} & 0.953/0.450 & 0.991/0.000 & 0.939/0.244 & 0.976/0.156\\
TrojanNN & 0.983/0.171 & \textcolor{lightgray}{0.302/0.509} & {0.996/0.112} & 0.950/0.450 & \textbf{1.000/0.000} & 0.921/0.227 & \underline{0.999/0.161}\\
Label-Consistent & \underline{0.992/0.130} & \textcolor{lightgray}{0.447/0.506} & \textbf{0.994/0.117} & 0.953/0.450 & 0.979/0.014 & 0.889/0.237 & 0.962/0.159\\ 
WaNet & \textbf{1.000/0.116} & \textcolor{lightgray}{0.456/0.505} & \textcolor{lightgray}{0.050/0.101} & \underline{0.951/0.450} & 0.891/0.034 & 0.869/0.251 & 0.863/0.144 \\  
ISSBA & \textbf{1.000/0.113} & \textcolor{lightgray}{0.436/0.507} & \textcolor{lightgray}{0.774/0.120} & 0.950/0.450 & {0.963/0.011} & 0.939/0.290 & \underline{0.965/0.157}\\
Adaptive-Blend & \textbf{0.982/0.184} & \textcolor{lightgray}{0.608/0.145} & \textcolor{lightgray}{0.014/0.069} & \textcolor{lightgray}{0.753/0.144} & \textcolor{lightgray}{0.000/0.023} & \textcolor{lightgray}{0.721/0.257} &
\textcolor{lightgray}{0.432/0.167} \\
\rowcolor{Gray}
    Average & \textbf{0.994/0.136} & 0.439/0.456 & 0.689/0.107 & \underline{0.923/0.406} & 0.832/0.013 & 0.899/0.244 &
    0.855/0.157\\ 
\rowcolor{LightCyan}
\midrule
\multicolumn{8}{c}{\hspace{20mm}\textbf{GTSRB}\hspace{8mm}} \\
\midrule
Badnet & 0.987/0.202 & \textcolor{lightgray}{0.476/0.502} & \underline{0.999/0.096} & 0.524/0.497 & OOT & \textbf{1.000/0.344} & 
0.911/0.193\\ 
Blend  & \textbf{0.910/0.207} & \textcolor{lightgray}{0.476/0.502} & \underline{0.897/0.093} & \textcolor{lightgray}{0.524/0.497} & OOT & \textcolor{lightgray}{0.286/0.337} & \textcolor{lightgray}{0.462/0.199}\\
TrojanNN & \underline{0.952/0.212} & \textcolor{lightgray}{0.476/0.502} & \textcolor{lightgray}{0.639/0.096} & \textcolor{lightgray}{0.524/0.497} & OOT & \textcolor{lightgray}{0.113/0.345} & \textbf{0.967/0.194}\\
Label-Consistent & 0.944/0.203 & \textcolor{lightgray}{0.476/0.502} & \textbf{1.000/0.115} & \textcolor{lightgray}{0.524/0.497} & OOT & \underline{0.998/0.362} & \textcolor{lightgray}{0.416/0.175}\\ 
WaNet & \textbf{0.996/0.115} & \textcolor{lightgray}{0.476/0.502} & \textcolor{lightgray}{0.037/0.109} & \textcolor{lightgray}{0.524/0.497} & OOT & \textcolor{lightgray}{0.129/0.306} & \textcolor{lightgray}{0.031/0.182} \\  
ISSBA & \textbf{0.999/0.211} & \textcolor{lightgray}{0.476/0.502} & \textcolor{lightgray}{0.725/0.092} & \textcolor{lightgray}{0.524/0.497} & OOT & \textcolor{lightgray}{0.584/0.339} & \textcolor{lightgray}{0.705/0.197}\\
Adaptive-Blend & \textbf{0.899/0.194} & \textcolor{lightgray}{0.299/0.392} & \textcolor{lightgray}{0.004/0.094} & \textcolor{lightgray}{0.750/0.388} & {OOT} & \textcolor{lightgray}{0.071/0.332} &
\textcolor{lightgray}{0.028/0.158} \\
\rowcolor{Gray}
    Average & \textbf{0.955/0.192} & 0.451/0.486 & \underline{0.614/0.099} & 0.556/0.481 & OOT & {0.454/0.338} &
    {0.503/0.185}\\
\rowcolor{LightCyan}
\midrule
\multicolumn{8}{c}{\hspace{20mm}\textbf{Tiny ImageNet}\hspace{3mm}} \\
\midrule
Badnet & \underline{0.989/0.088} & \textcolor{lightgray}{0.480/0.502} & 0.841/0.108 & \textcolor{lightgray}{0.522/0.497} & OOT & \textbf{0.999/0.271} &
\textcolor{lightgray}{0.462/0.176}\\
Blend  & \textbf{0.919/0.108} & \textcolor{lightgray}{0.478/0.502} & \textcolor{lightgray}{0.249/0.086} & \textcolor{lightgray}{0.522/0.496} & OOT & \textcolor{lightgray}{0.551/0.260} &
\underline{0.874/0.175}\\
TrojanNN & 0.961/0.222 & \textcolor{lightgray}{0.478/0.502} & {0.963/0.104} & \textcolor{lightgray}{0.522/0.497} & OOT & \underline{0.972/0.301} &
\textbf{0.985/0.150}\\
Label-Consistent & \underline{0.839/0.039} & \textcolor{lightgray}{0.478/0.502} & \textcolor{lightgray}{0.460/0.088} & \textcolor{lightgray}{0.522/0.497} & OOT & \textcolor{lightgray}{0.741/0.187} &
\textbf{0.931/0.203}\\
WaNet & \textbf{0.959/0.086} & \textcolor{lightgray}{0.478/0.502} & \textcolor{lightgray}{0.087/0.082} & \textcolor{lightgray}{0.522/0.497} & OOT & \textcolor{lightgray}{0.446/0.254} &
\textcolor{lightgray}{0.577/0.151}
\\
ISSBA & {0.886/0.209} & \textcolor{lightgray}{0.478/0.502} & \underline{0.954/0.097} & \textcolor{lightgray}{0.522/0.497} & OOT & \textcolor{lightgray}{0.691/0.297} & 
\textbf{0.978/0.137}\\
Adaptive-Blend & \textbf{0.949/0.095} & \textcolor{lightgray}{0.392/0.502} & \textcolor{lightgray}{0.210/0.099} & \textcolor{lightgray}{0.621/0.497} & OOT & \textcolor{lightgray}{0.651/0.190} &
\textcolor{lightgray}{0.331/0.176}\\
\rowcolor{Gray}
    Average & \textbf{0.929/0.121} & 0.466/0.502 & 0.538/0.095 & 0.536/0.497 & OOT & {0.722/0.251} &
    \underline{0.734/0.167}\\
\midrule
\bottomrule[1pt]
\end{tabular}
}
\vspace*{-4mm}
\end{table*}
\subsection{Experiment Results}
\paragraph{Effectiveness and Efficiency of PSBD.} 

As shown in Table~\ref{tab: poison rate 0.1}, PSBD demonstrates excellent backdoor detection performance across a wide range of attack scenarios, while effectively preserving a substantial amount of clean data. The results also demonstrate that PSBD achieves a substantial improvement in detection performance compared to the defense baselines. \textit{In contrast, all baseline methods fail in some evaluation attacks.}

\begin{figure}[t]
\centering{
    \begin{subfigure}{0.23\textwidth}
         \centering
         \includegraphics[width=\textwidth]{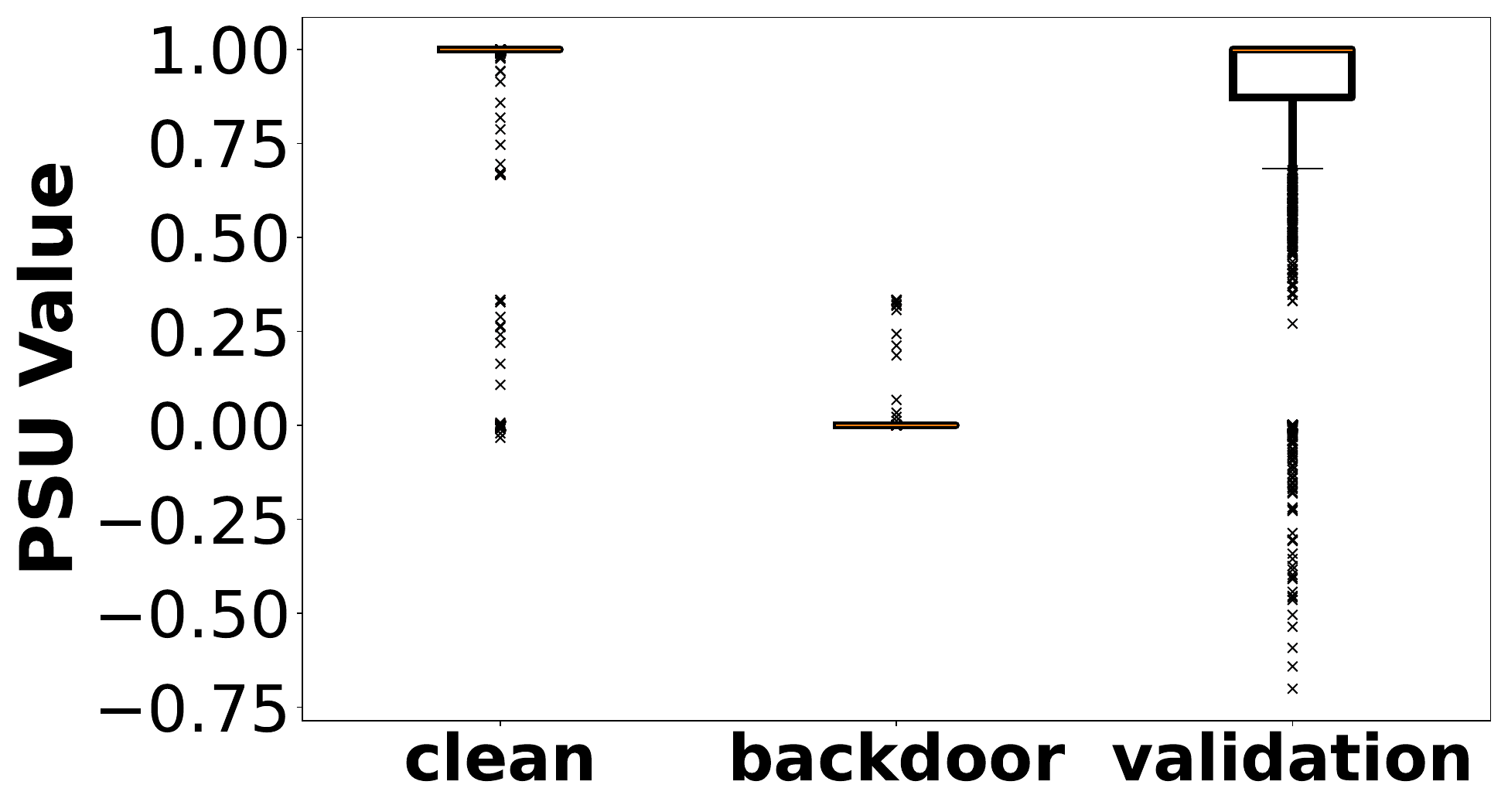}
         \caption{BadNets}
     \end{subfigure}
     \begin{subfigure}{0.23\textwidth}
         \centering
         \includegraphics[width=\textwidth]{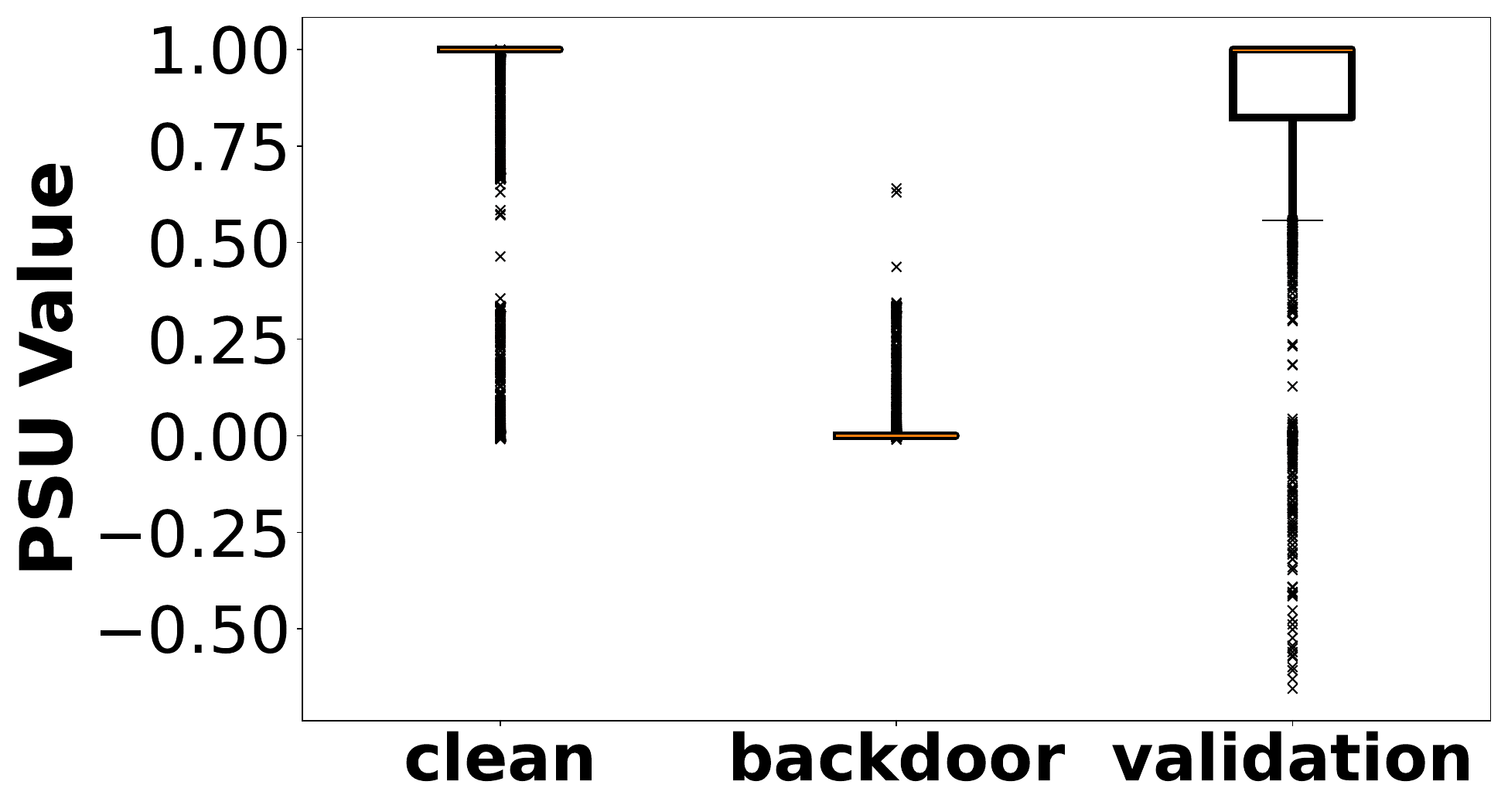}
         \caption{WaNet}
     \end{subfigure}
}
\vspace*{-2mm}
\caption{\footnotesize{The PSU values of BadNets and WaNet in CIFAR-10. The poisoning ratio is 10\%. PSBD exhibits strong capability to effectively differentiate clean data from backdoor data.
}}
  \label{fig:box fig}
\end{figure}

Specifically, on the CIFAR-10 dataset, the Spectral Signature method failed to detect backdoor data under all attack scenarios, while simultaneously misclassifying a substantial amount of clean data as backdoor data. This suggests that when the trigger pattern is relatively large or complex, the spectral signature property may be difficult to capture. 
The Spectre method demonstrated relatively effective detection capabilities across most attack scenarios, but it also has a high FPR. This is not desirable in practice as it filters out a significant amount of clean training data, which can lead to issues like overfitting due to insufficient training data. Although the SCAN, Strip, SCP, and CD-L methods exhibited relatively acceptable performance across most attack scenarios, achieving a relatively high TPR and a low FPR, their effectiveness deteriorated when confronted with attacks employing confusion strategy such as WaNet and Adaptive-Blend. This confusion strategy aims to disrupt the model by retaining an equal or greater proportion of confounding samples that contain the trigger pattern but are correctly labeled. On the GTSRB dataset, which contains a larger number of classes, the Spectral Signature and Spectre methods completely fail to detect backdoor data. The Strip, SCP, and CD-L methods exhibit good detection performance only in a few specific attack scenarios. In contrast, our PSBD method demonstrates strong detection capabilities across all attack scenarios, highlighting its robustness and generalizability.

On the more challenging Tiny ImageNet, all baseline methods failed in most attack scenarios. This failure is likely due to the increased complexity of image features, which weakened their ability to capture the mapping between trigger pattern and target label. Encouragingly, our PSBD method maintained its effectiveness, ranking in the top two for all attacks except for TrojanNN (where it still achieved a TPR of 0.961){ and ISSBA. The slight performance degradation under ISSBA is likely due to the model's insufficient ability to extract features from the data. This is reflected in the significantly lower clean accuracy of the model compared to that of the clean model. The clean accuracy of models can be found in the Appendix Table~\ref{tab: asr and ca appendix}.} By employing dropout to diminish prominent image features and utilizing robust neuron bias paths, PSBD effectively discerned the mapping from trigger pattern to target label.

\vspace{-5mm}
\paragraph{The Strong Discriminative Capability of PSBD.}
PSBD excels in its critical ability to effectively differentiate between clean and backdoor training data. By leveraging the PSU values, we have developed some informative box plots that clearly and vividly demonstrate the remarkable discriminative power of our approach. 
As shown in Figure~\ref{fig:box fig}, provide a visual representation of how PSBD separates clean data from backdoor data.
In these plots, it is prominently visible that backdoor data is characterized by lower PSU values, distinguishing it from clean training and validation data, which generally exhibit higher PSU values. This distinction is crucial for effective backdoor detection, as it highlights the different behavioral patterns of the model when exposed to clean versus poisoned data. The lower PSU values in backdoor data indicate the model's ability to maintain confident predictions, a direct consequence of the embedded trigger in these samples with neuron bias effect.

%% file: sec/6_conclusion.tex
\section{Conclusion}
\label{conlusion}
In our study, we developed PSBD, a simple and effective method to detect backdoor samples in the training dataset by focusing on Prediction Shift phenomenon under dropout conditions, leading to the concept of neuron bias effect.
By analyzing changes in prediction confidence with and without dropout, PSBD effectively distinguishes between clean and backdoor data across multiple datasets and attack types. 
This research contributes a practical and effective solution to the challenge of backdoor attacks in DNNs, marking a notable advancement in the field of neural network security. Future efforts could explore extending the PSBD method to a broader range of domains, such as natural language processing or time-series analysis.

%% file: sec/7_appendix.tex
\newpage
\clearpage
\begin{center}
\large
\appendix
\large
\section*{Appendix}
\label{sec:appendix}
\end{center}

\setcounter{figure}{0}
\makeatletter 
\renewcommand{\thefigure}{A\arabic{figure}}
\renewcommand{\theHfigure}{A\arabic{figure}}
\renewcommand{\thetable}{A\arabic{table}}
\renewcommand{\theHtable}{A\arabic{table}}

\renewcommand{\thefigure}{A\@arabic\c@figure}
\makeatother
\setcounter{table}{0}

\section{Pilot Studies on More Attacks}
In this section, we will show the more results of our pilot study in Section~\ref{pilot study 1} and a combined method between MC-Dropout\cite{gal2016dropout} and SCP\cite{guo2023scale}. We maintain the same experimental setup with pilot study that conduct experiments on the CIFAR-10 dataset\cite{krizhevsky2009learning} with ResNet-18\cite{He_2016_CVPR} trained 100 epochs. Apart from Adaptive-Blend\cite{qi2022revisiting} attack, which has a poisoning ratio of 1\%, all other attacks maintain a poisoning ratio of 10\%.

\subsection{Pilot Study: MC-Dropout Predictive Uncertainty.}
\label{pilot 1 appendix}
\begin{figure*}[h]
    \centering
    \hspace{15mm}
    \begin{subfigure}[h]{0.24\textwidth}
         \centering
        \includegraphics[width=\textwidth]{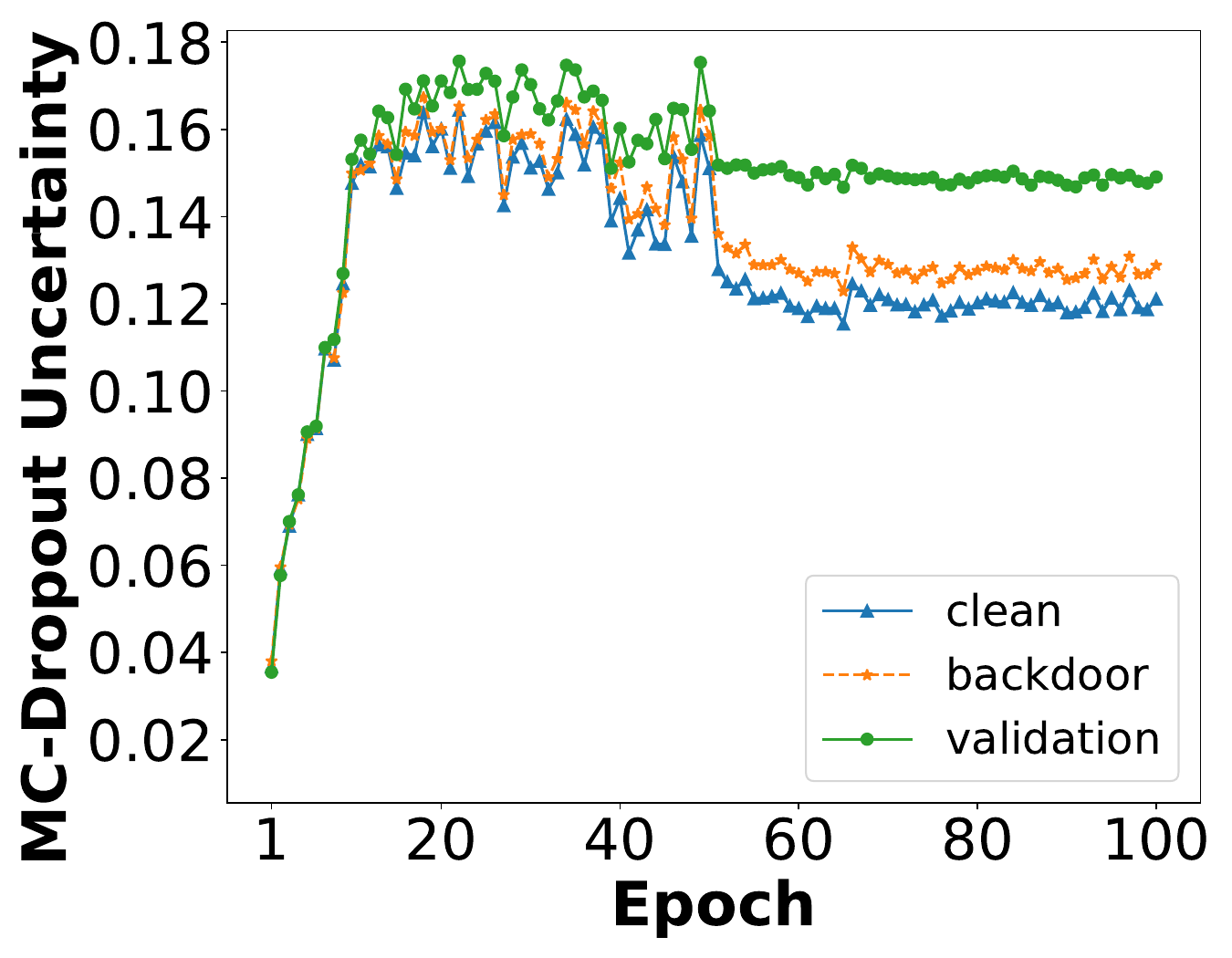}
         \caption{Benign Model}
     \end{subfigure}
    \hfill
    \begin{subfigure}[h]{0.24\textwidth}
         \centering
        \includegraphics[width=\textwidth]{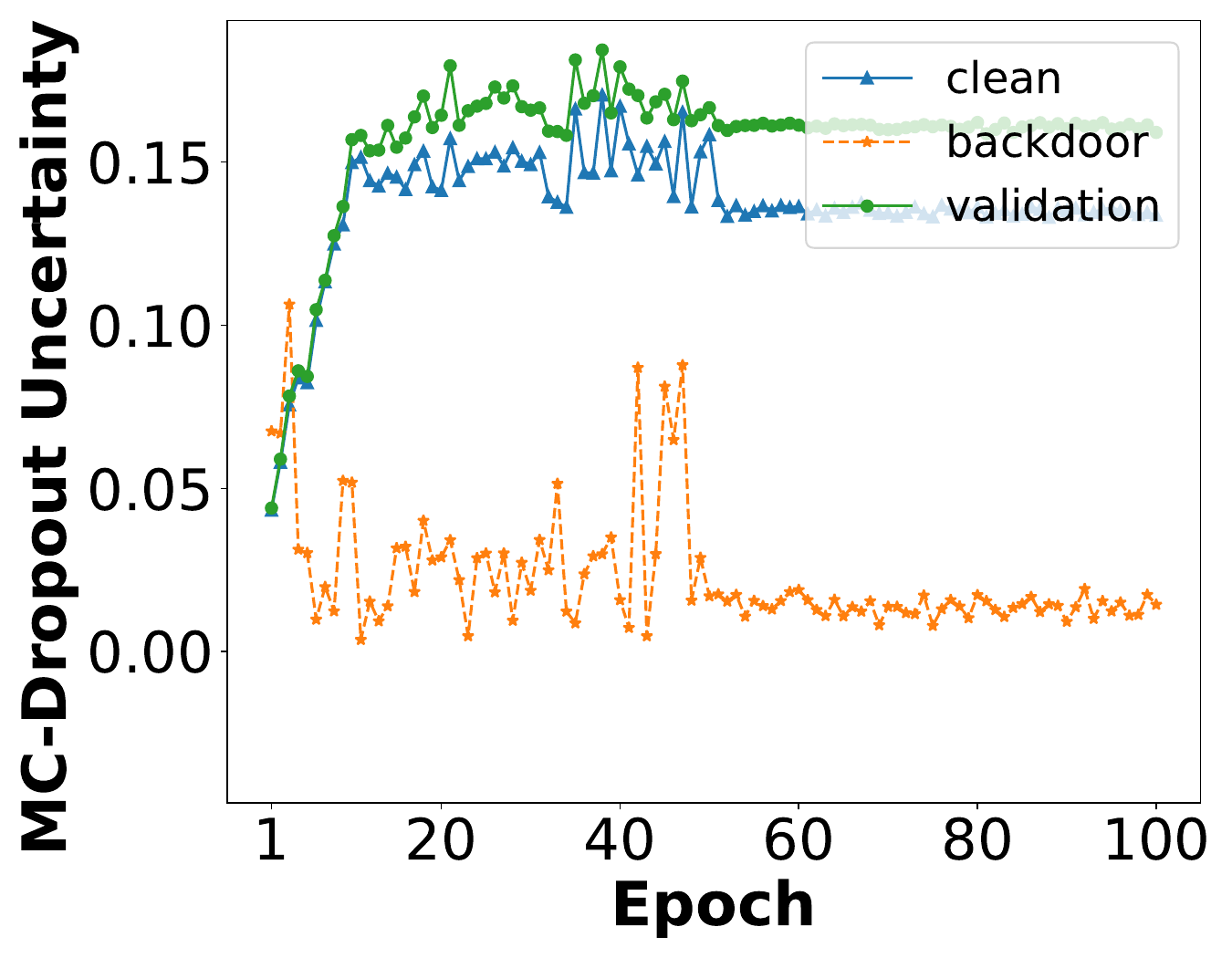}
         \caption{Blend}
     \end{subfigure}
     \hfill
     \begin{subfigure}[h]{0.24\textwidth}
         \centering
         \includegraphics[width=\textwidth]{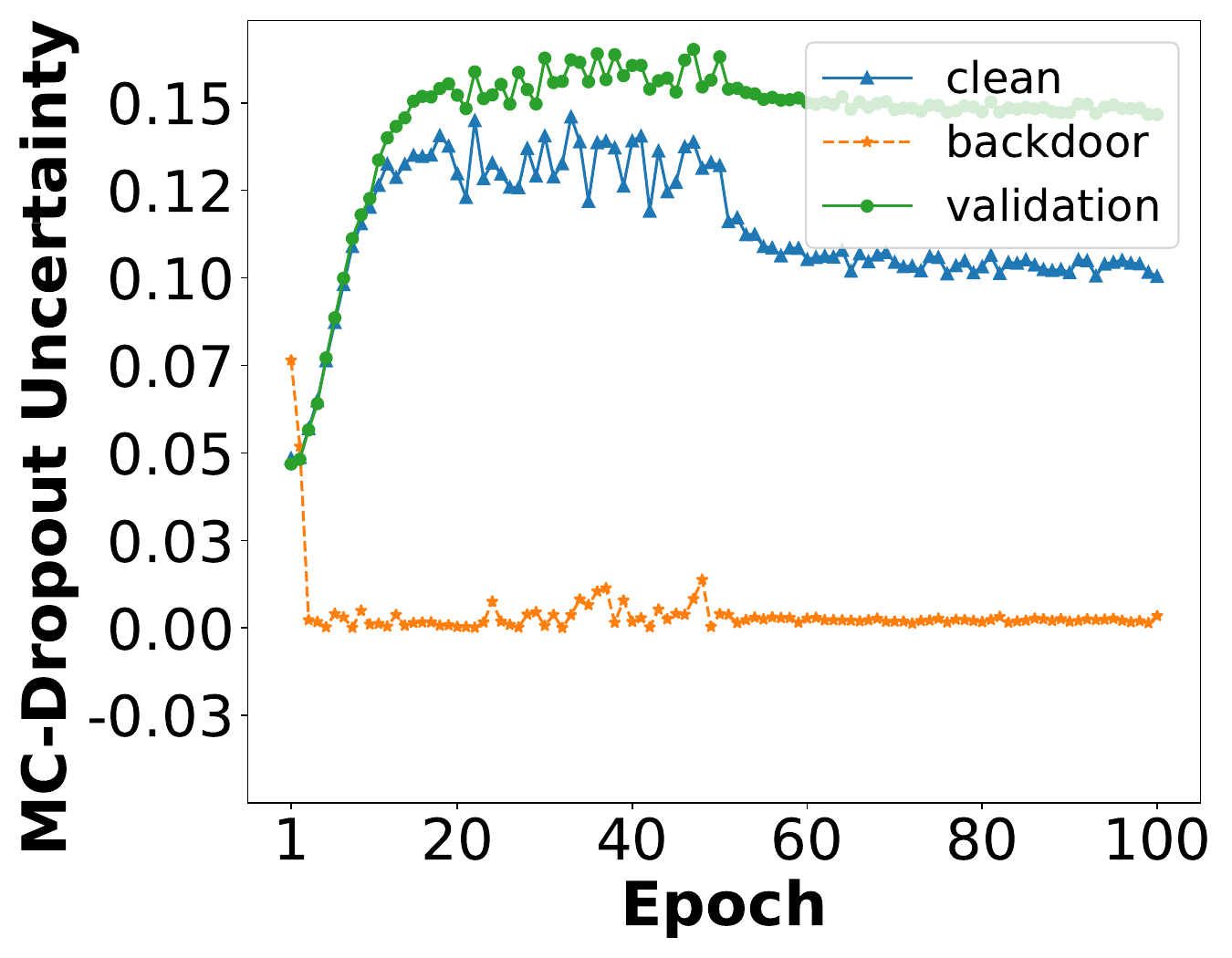}
         \caption{TrojanNN}
     \end{subfigure}
     \hspace{15mm}
     \vfill
     \hspace{15mm}
     \begin{subfigure}[h]{0.24\textwidth}
         \centering
         \includegraphics[width=\textwidth]{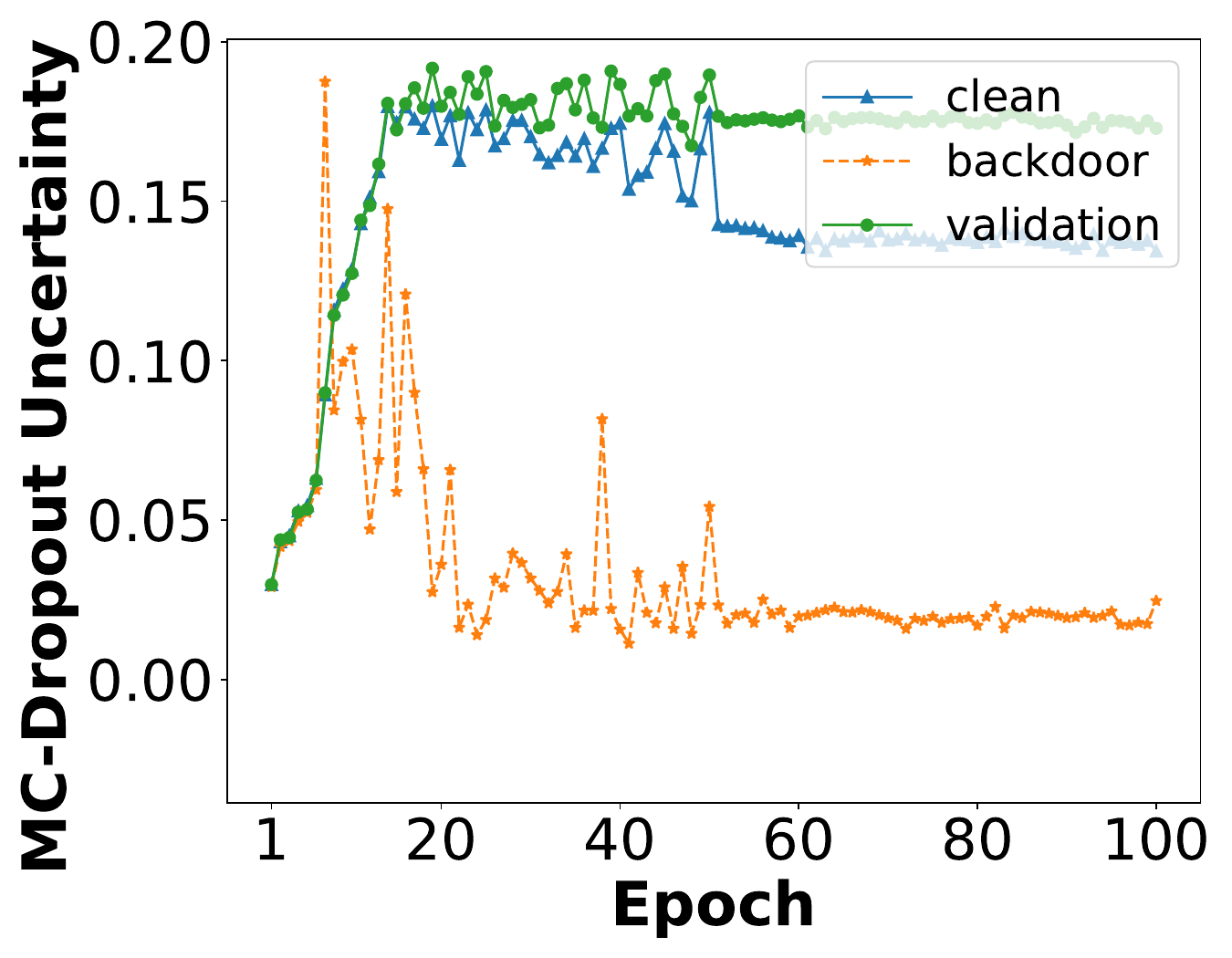}
         \caption{Label-Consistence}
     \end{subfigure}
     \hfill
     \begin{subfigure}[h]{0.24\textwidth}
         \centering
         \includegraphics[width=\textwidth]{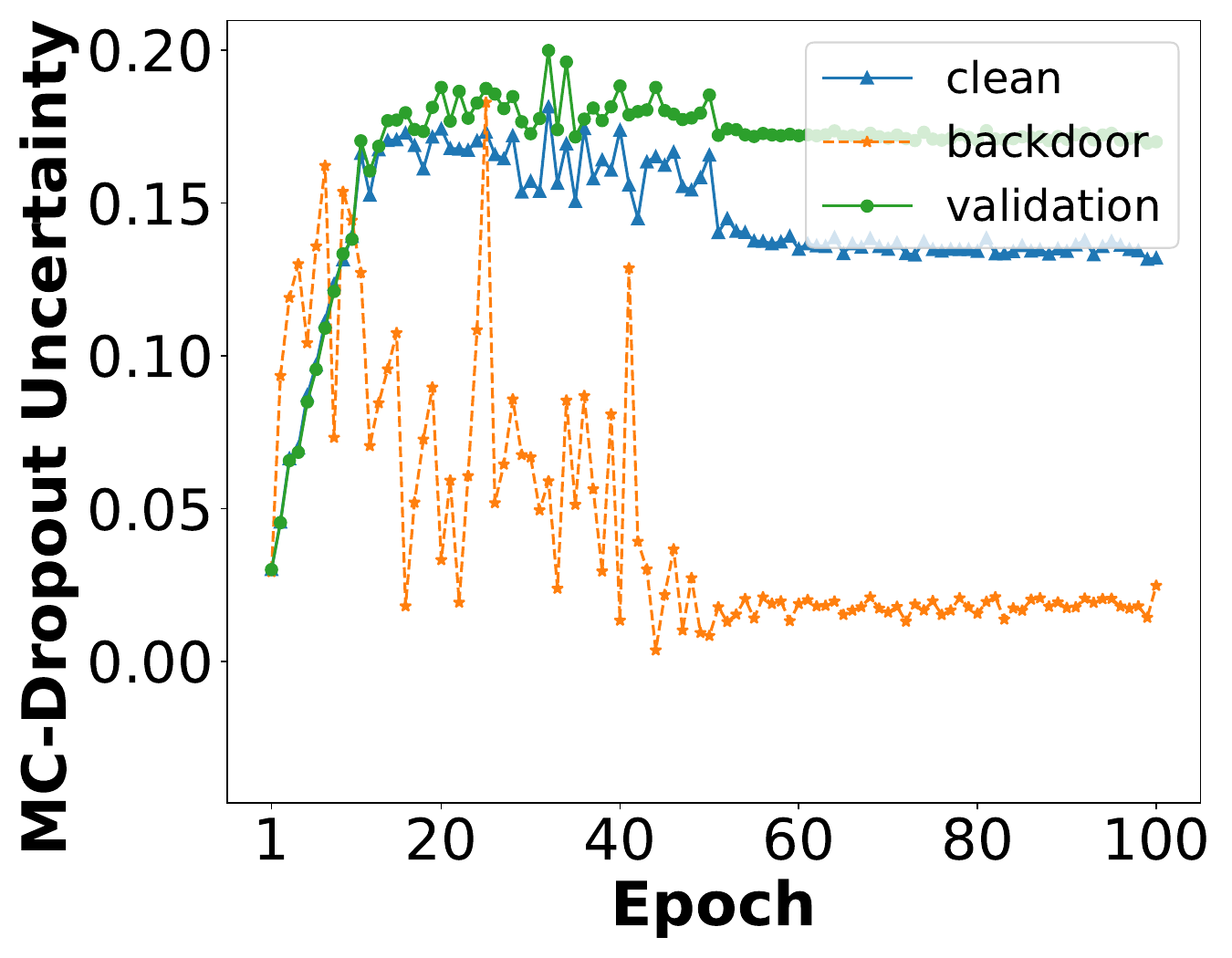}
         \caption{ISSBA}
     \end{subfigure}
     \hfill
     \begin{subfigure}[h]{0.24\textwidth}
         \centering
         \includegraphics[width=\textwidth]{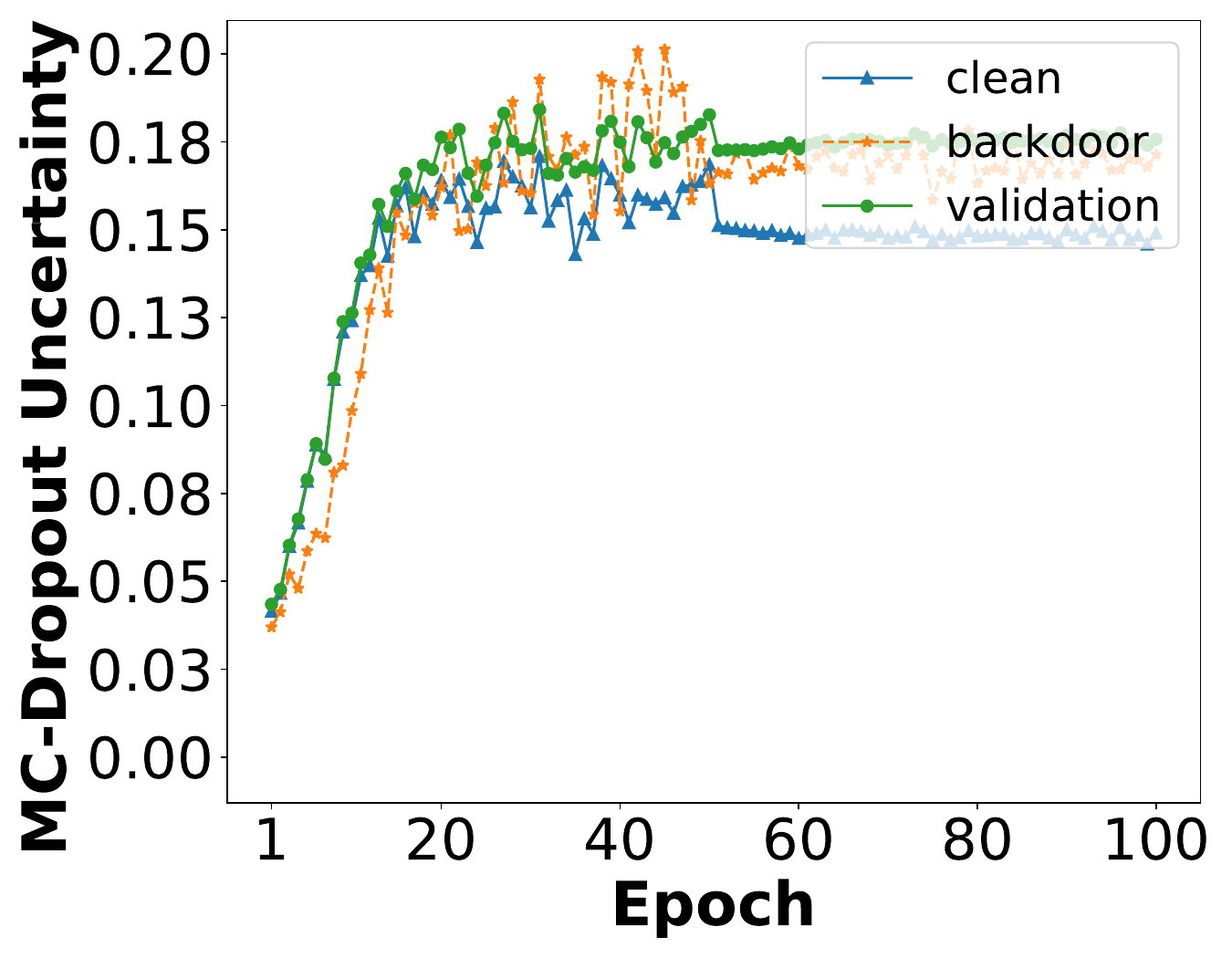}
         \caption{Adaptive-Blend}
     \end{subfigure}
     \hspace{15mm}
    \caption{The average MC-Dropout uncertainty of clean training data, backdoor training data and clean validation data under benign and various poisoned models.}
    \label{fig:pilot study 1 appendix}
\end{figure*}

As shown in Figure~\ref{fig:pilot study 1 appendix}, under most attacks, the average MC-Dropout uncertainty of backdoor training data is significantly smaller than that of clean data, both lower than clean training data and clean validation data, and this difference tends to stabilize in the later stages of model training. However, under Adaptive-Blend attack, we can observe that the uncertainty of backdoor training data is even slightly higher than clean training data. The observations align with the part of our first pilot study in the main paper Section~\ref{pilot study 1}, which suggests that using uncertainty defined by standard deviation alone may not be sufficient to detect backdoor data in general attack scenarios.

\subsection{Exploring the Potential Synergy: MC-Dropout and SCP in Combination.}
\label{pilot 2 appendix}
In the Section ~\ref{pilot study 1} of our main paper, we can observe that the model's mapping from trigger to target label in backdoor data is more salient and robust compared to general image features. In light of this observation, we hypothesize the potential factor of the proximity between backdoor training data uncertainty and clean training data under WaNet\cite{nguyen2021wanet} attack can be attributed to the absence of sufficient sabotage on the image features. We expect that the model's ability to extract feature patterns from clean data will be significantly affected, while that from backdoor data will be minimally impacted. Consequently, we can enlarge the uncertainty gap between clean data and backdoor data. One direct method is to increase the dropout rate $p$. However, due to our inability to ascertain the presence or quantity of backdoor data within the suspicious training dataset, we encounter difficulty in determining an optimal value for $p$. Thus, we incorporate elements of SCP approach to introduce a more controllable uncertainty with MC-Dropout uncertainty. SCP can be considered as an instance of input-level uncertainty. It amplifies the pixel values of input images by multiple times, aiming to disrupt the feature patterns in the image. Therefore, we conduct further experiments to see if adding input-level uncertainty further disrupts the general feature patterns in the image.

\paragraph{Settings.}
We only incorporate controllable input-level uncertainty in our experiments. Specifically, we first amplify the pixel values of input images by a factor of three following SCP. Then we compute the MC-Dropout uncertainty of these scaled input images in three data types. Our goal is that controllably increase the uncertainty of the input data, and deeply disrupt the feature patterns in the image. As an expected result, the uncertainty of the clean training data becomes closer to that of the clean validation data, whereas the backdoor training data uncertainty becomes markedly distant from them. %

\paragraph{Results.}
\begin{figure*}[h]
    \centering
    \begin{subfigure}[h]{0.24\textwidth}
         \centering
         \includegraphics[width=\textwidth]{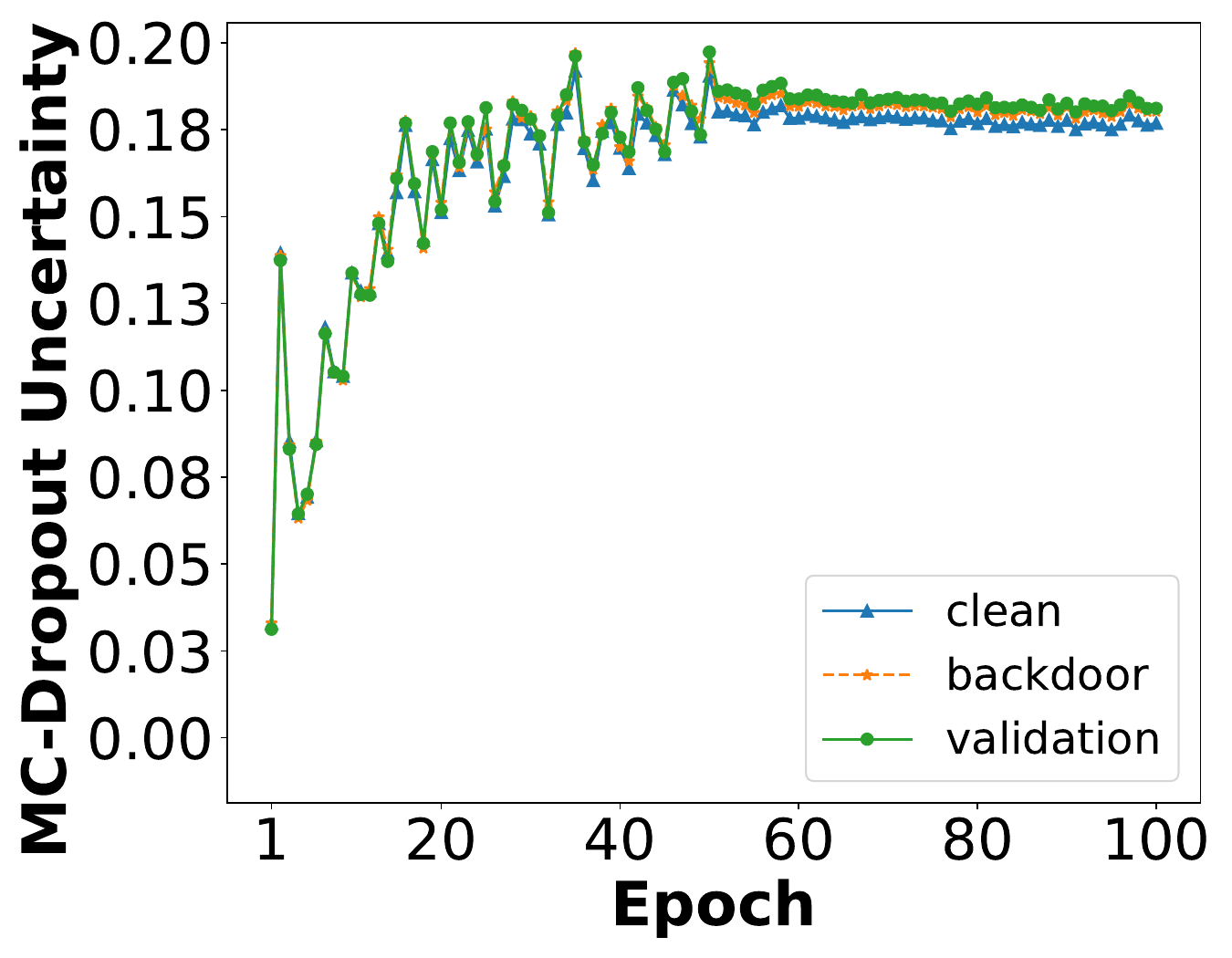}
         \caption{Benign Model}
         \label{fig:pilot study 2 benign}
     \end{subfigure}
     \hfill
     \begin{subfigure}[h]{0.24\textwidth}
         \centering
         \includegraphics[width=\textwidth]{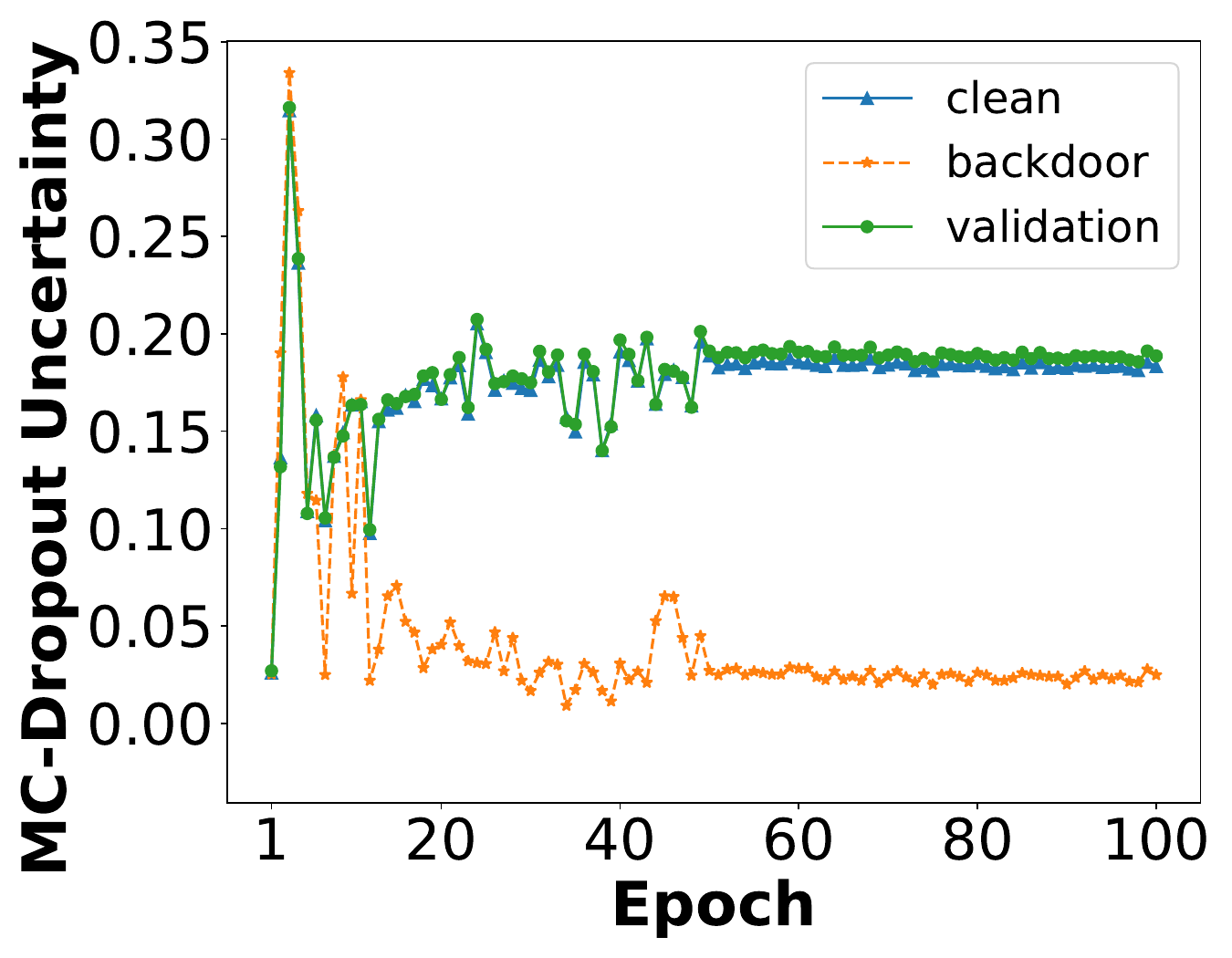}
         \caption{BadNets}
         \label{fig:pilot study 2 badnets}
     \end{subfigure}
     \hfill
     \begin{subfigure}[h]{0.24\textwidth}
         \centering
         \includegraphics[width=\textwidth]{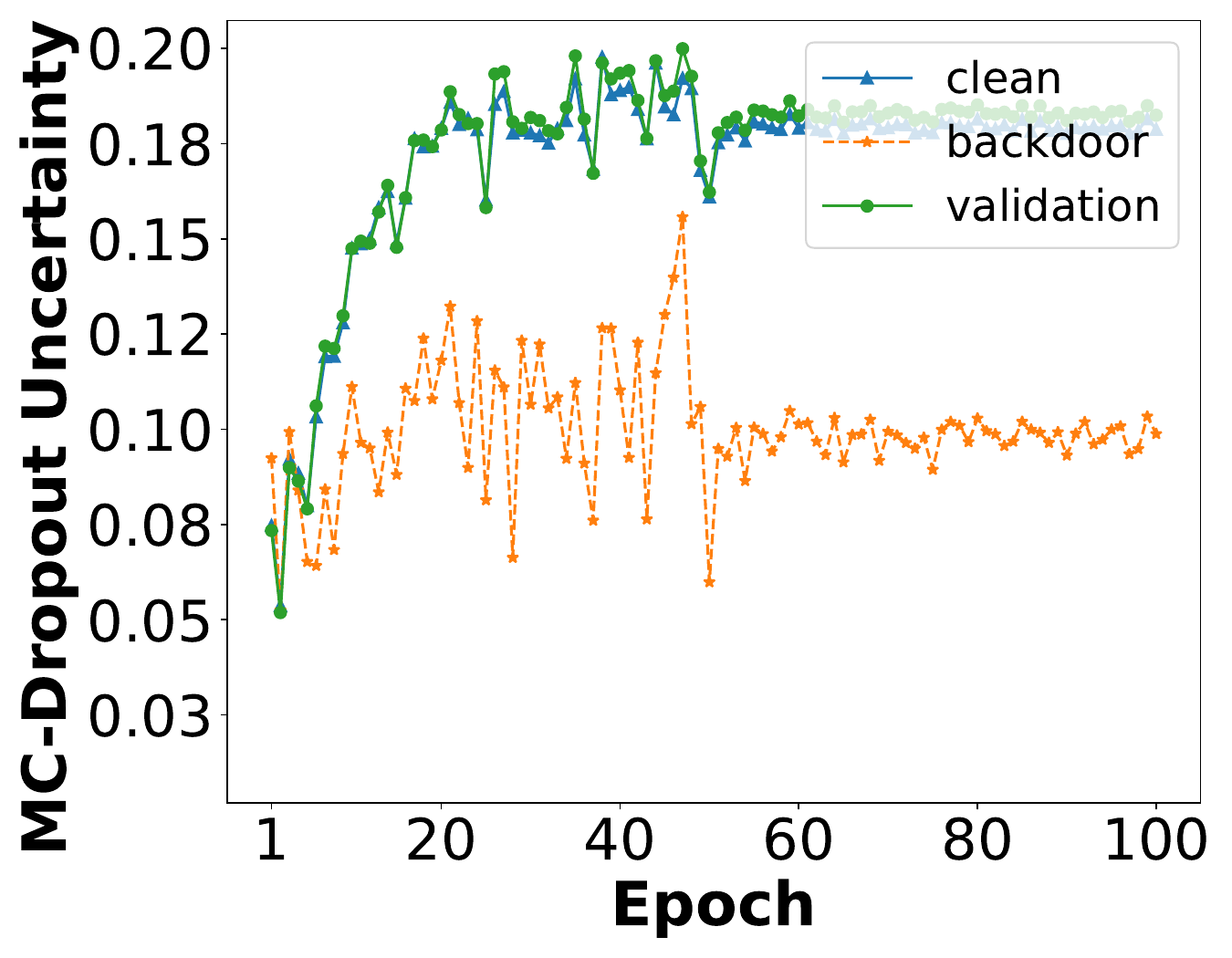}
         \caption{Blend}
         \label{fig:pilot study 2 blend}
     \end{subfigure}
     \hfill
     \begin{subfigure}[h]{0.24\textwidth}
         \centering
         \includegraphics[width=\textwidth]{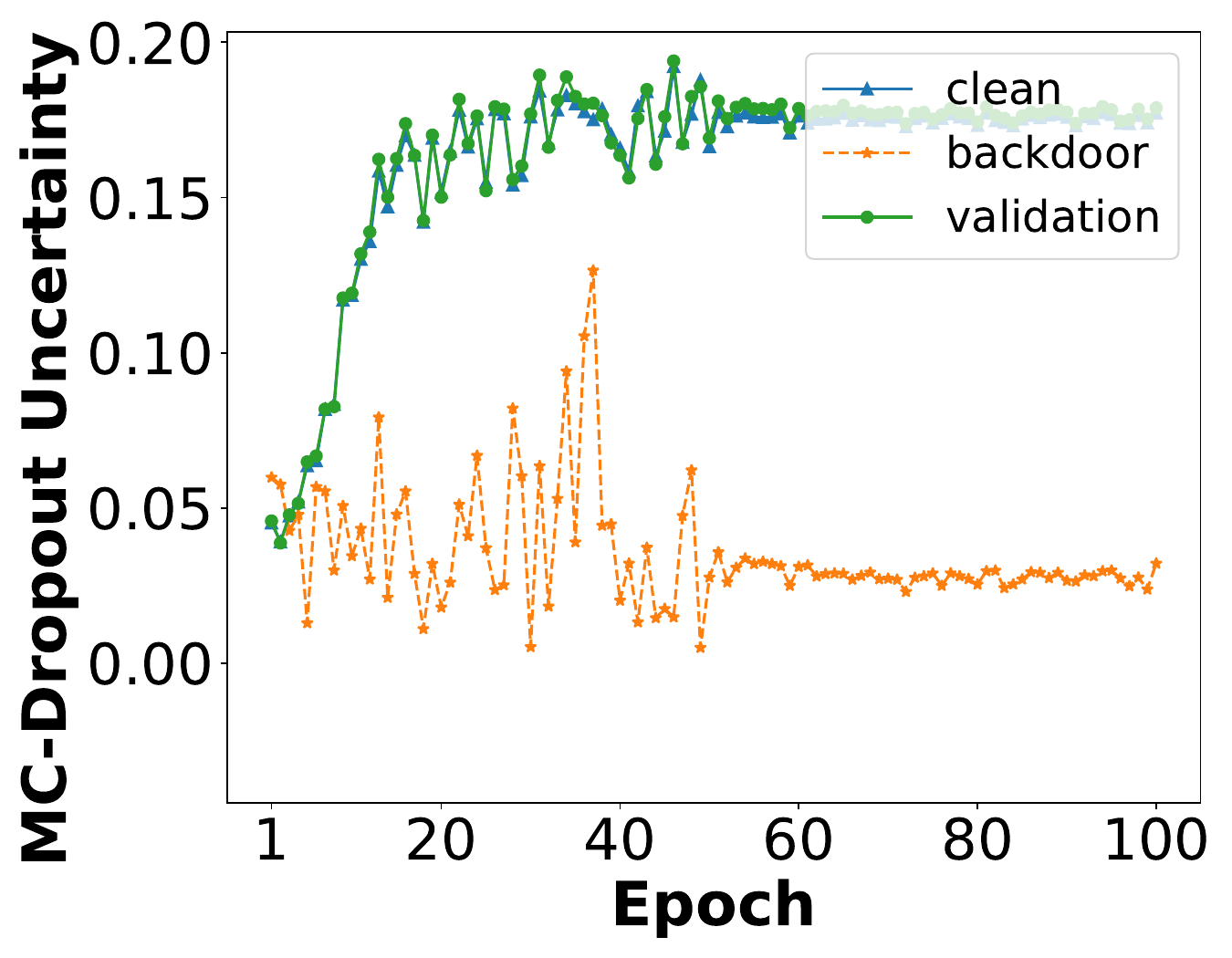}
         \caption{TrojanNN}
     \end{subfigure}
     \vfill
     \begin{subfigure}[h]{0.24\textwidth}
         \centering
         \includegraphics[width=\textwidth]{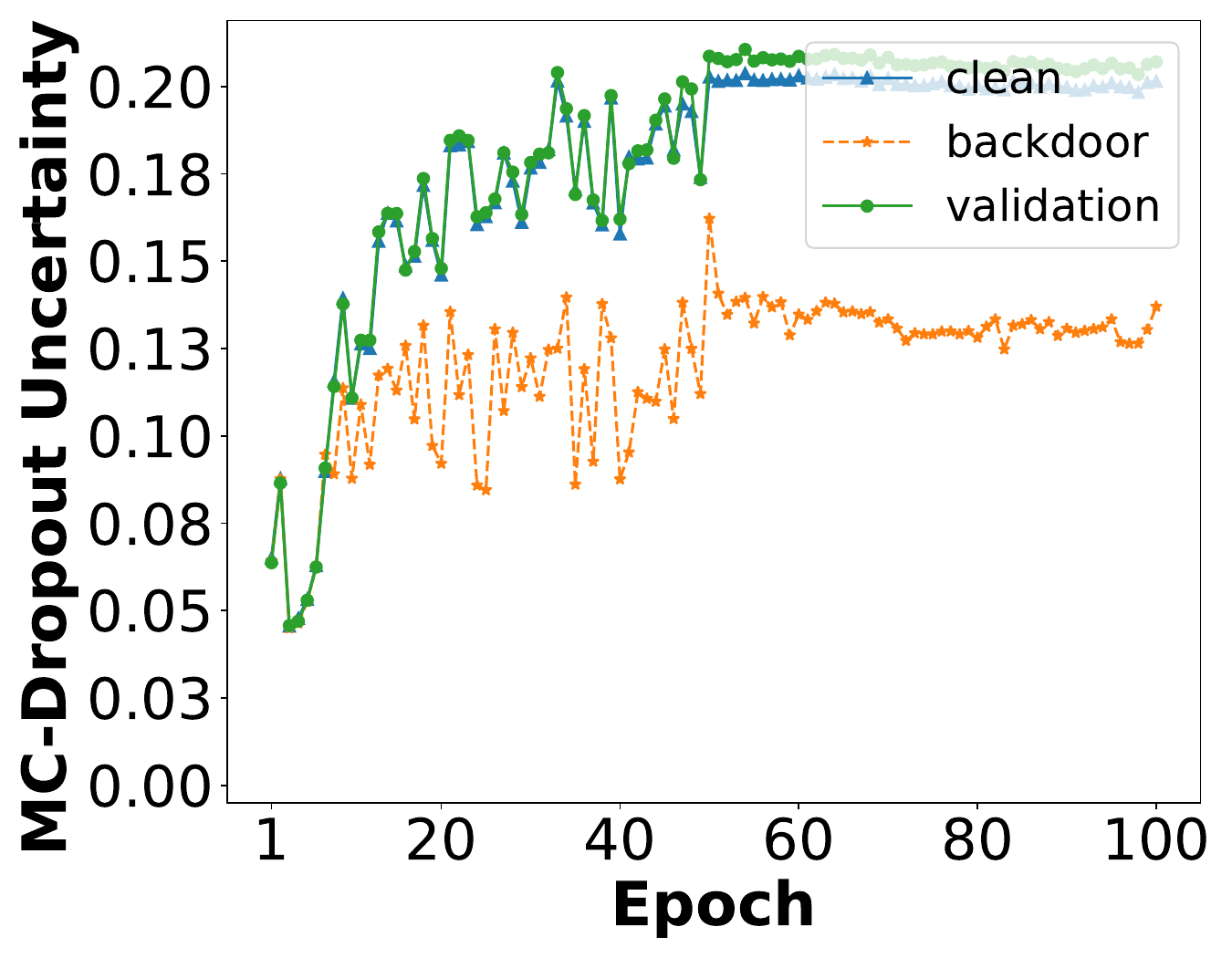}
         \caption{Label-Consistence}
         \label{fig:pilot study 2 lc}
     \end{subfigure}
     \hfill
     \begin{subfigure}[h]{0.24\textwidth}
         \centering
         \includegraphics[width=\textwidth]{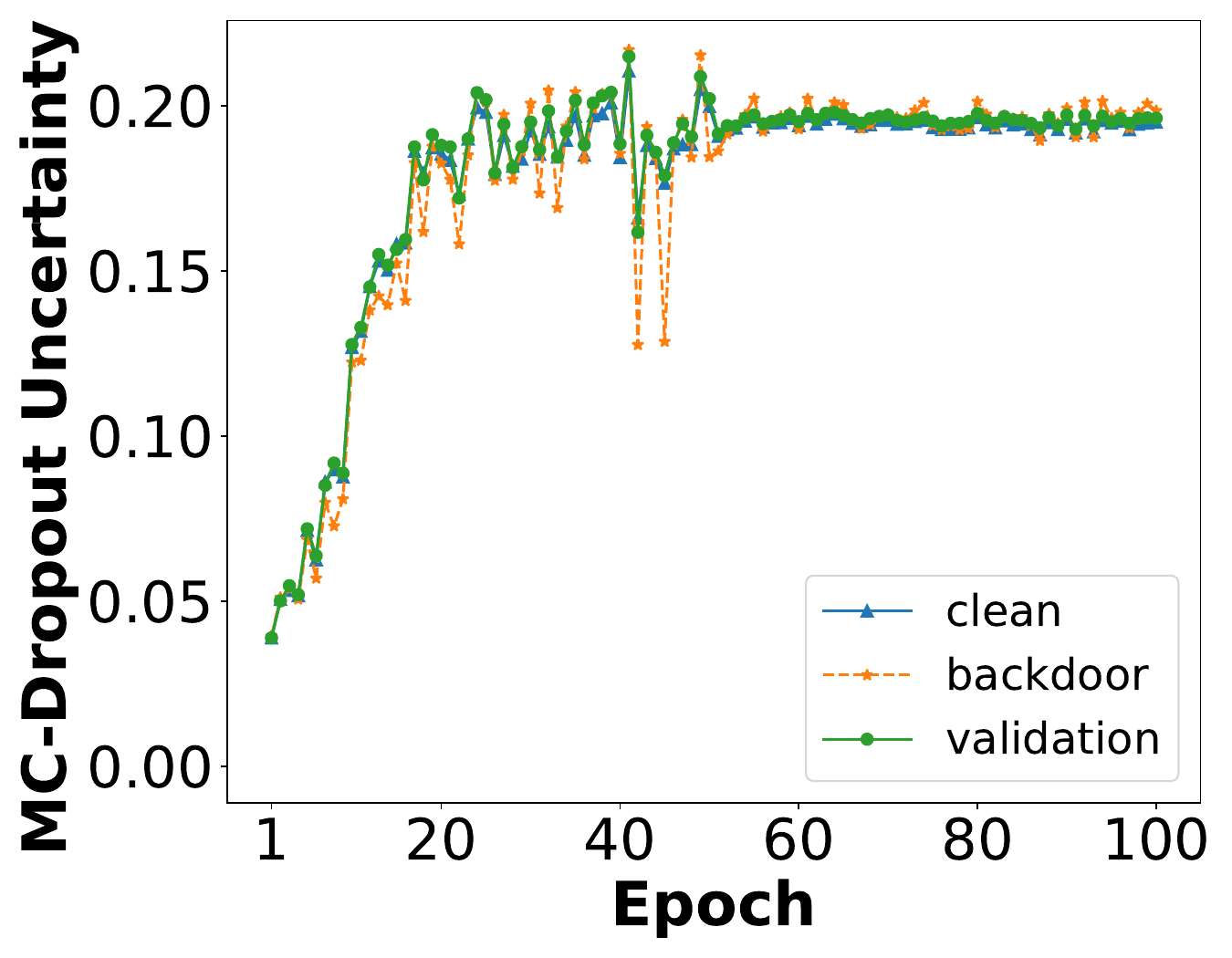}
         \caption{WaNet}
         \label{fig:pilot study 2 wanet}
     \end{subfigure}
     \hfill
     \begin{subfigure}[h]{0.24\textwidth}
         \centering
         \includegraphics[width=\textwidth]{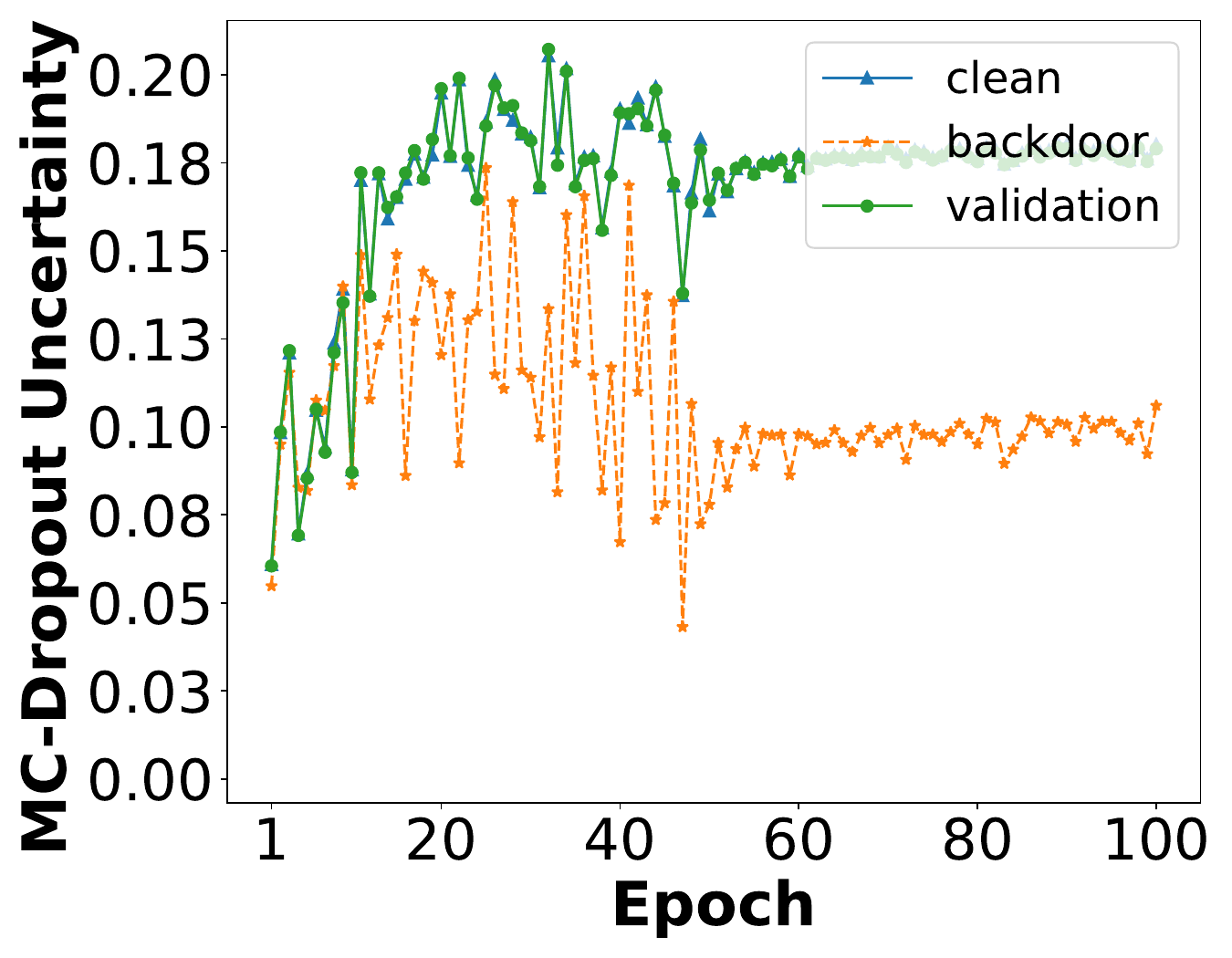}
         \caption{ISSBA}
     \end{subfigure}
     \hfill
     \begin{subfigure}[h]{0.24\textwidth}
         \centering
         \includegraphics[width=\textwidth]{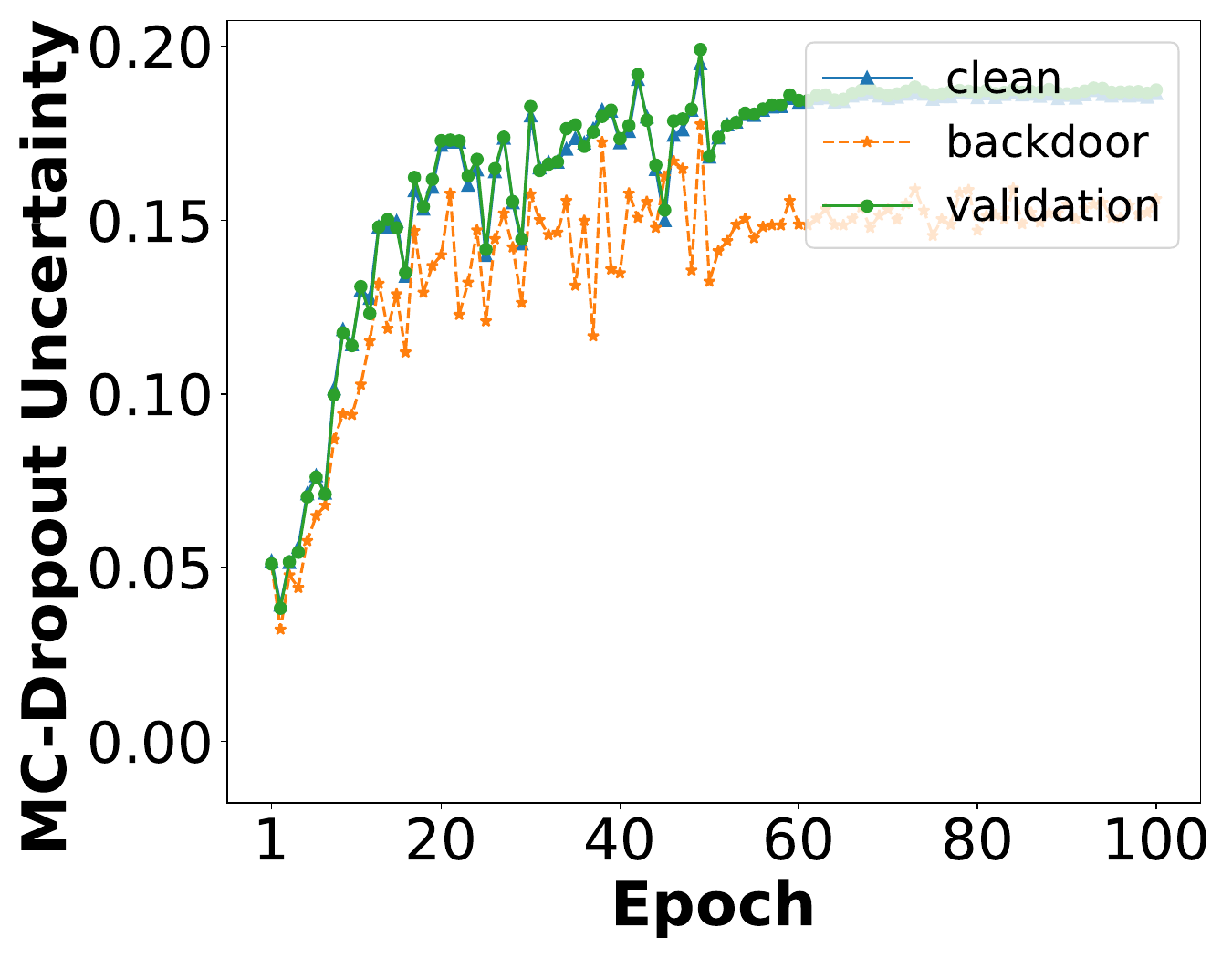}
         \caption{Adaptive-Blend}
         \label{fig:pilot study 2 adaptive-blend}
     \end{subfigure}
    \caption{The average MC-Dropout uncertainty combination with input-level uncertainty of clean training data, backdoor training data and clean validation data under benign and poisoned models.}
    \label{fig:pilot study 2}
\end{figure*}
\begin{figure*}[h]
    \centering
     \begin{subfigure}[h]{0.3\textwidth}
         \centering
         \includegraphics[width=\textwidth]{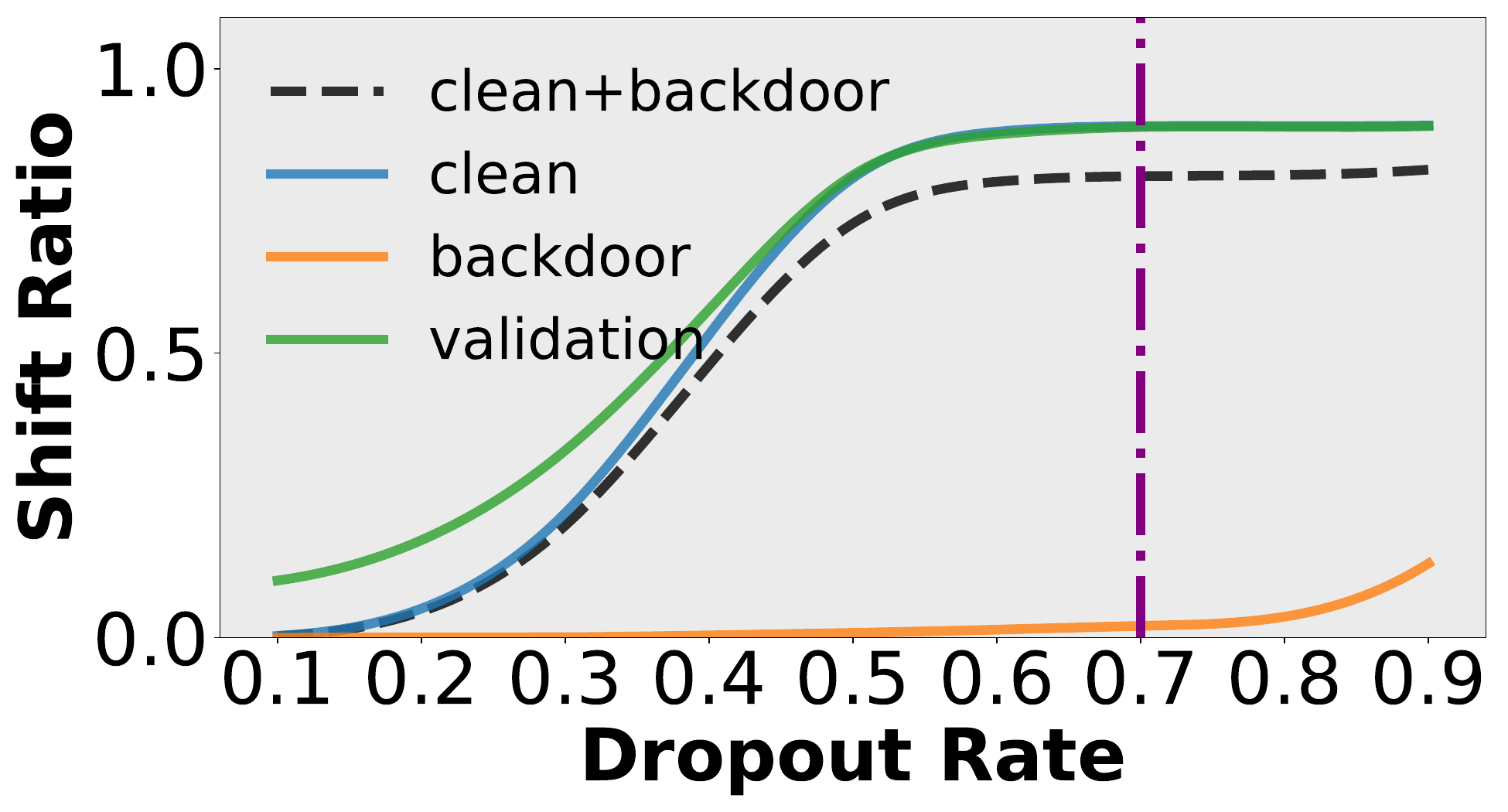}
     \end{subfigure}
     \hfill
     \begin{subfigure}[h]{0.3\textwidth}
         \centering
         \includegraphics[width=\textwidth]{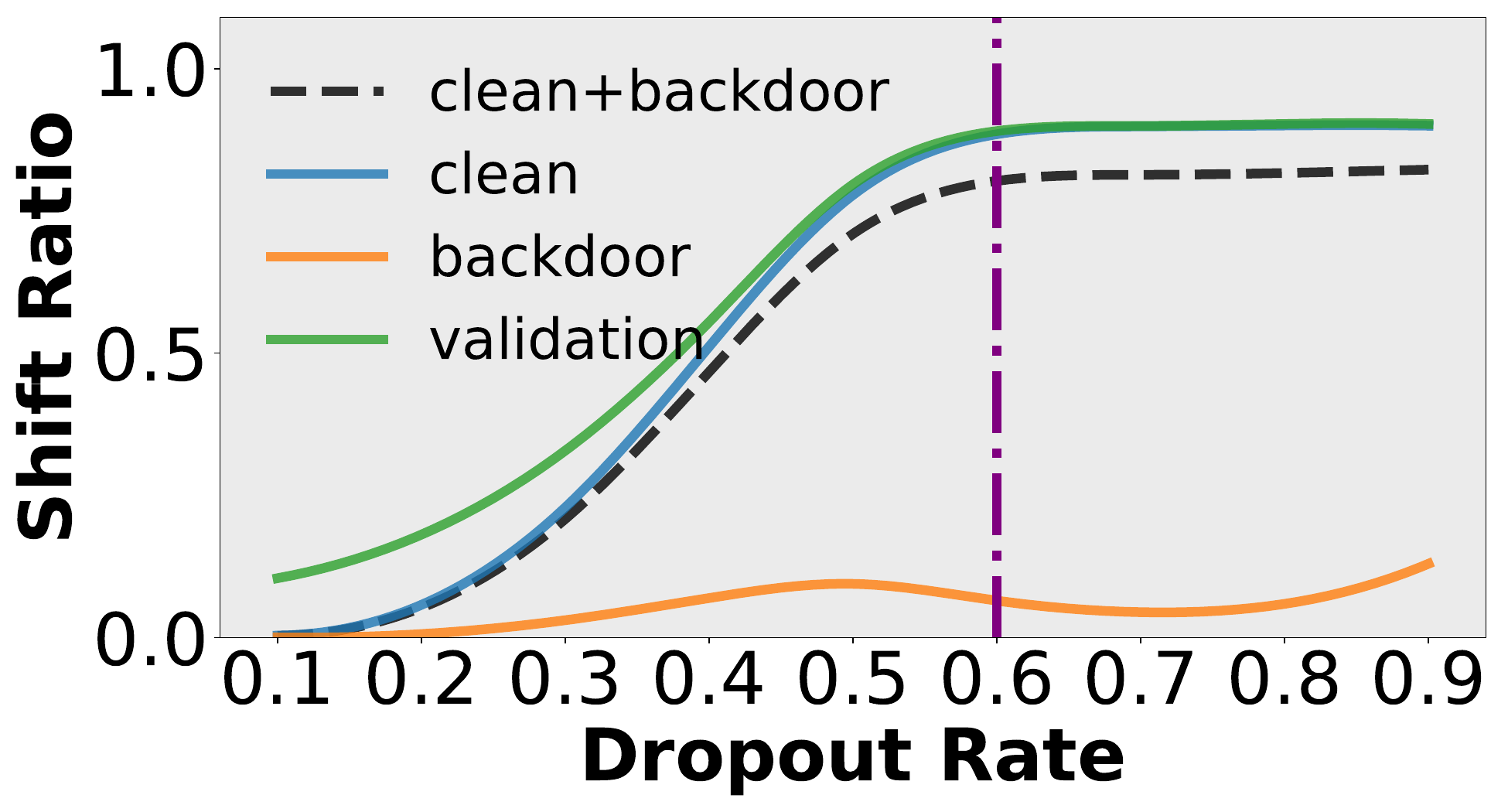}
     \end{subfigure}
     \hfill
     \begin{subfigure}[h]{0.3\textwidth}
         \centering
         \includegraphics[width=\textwidth]{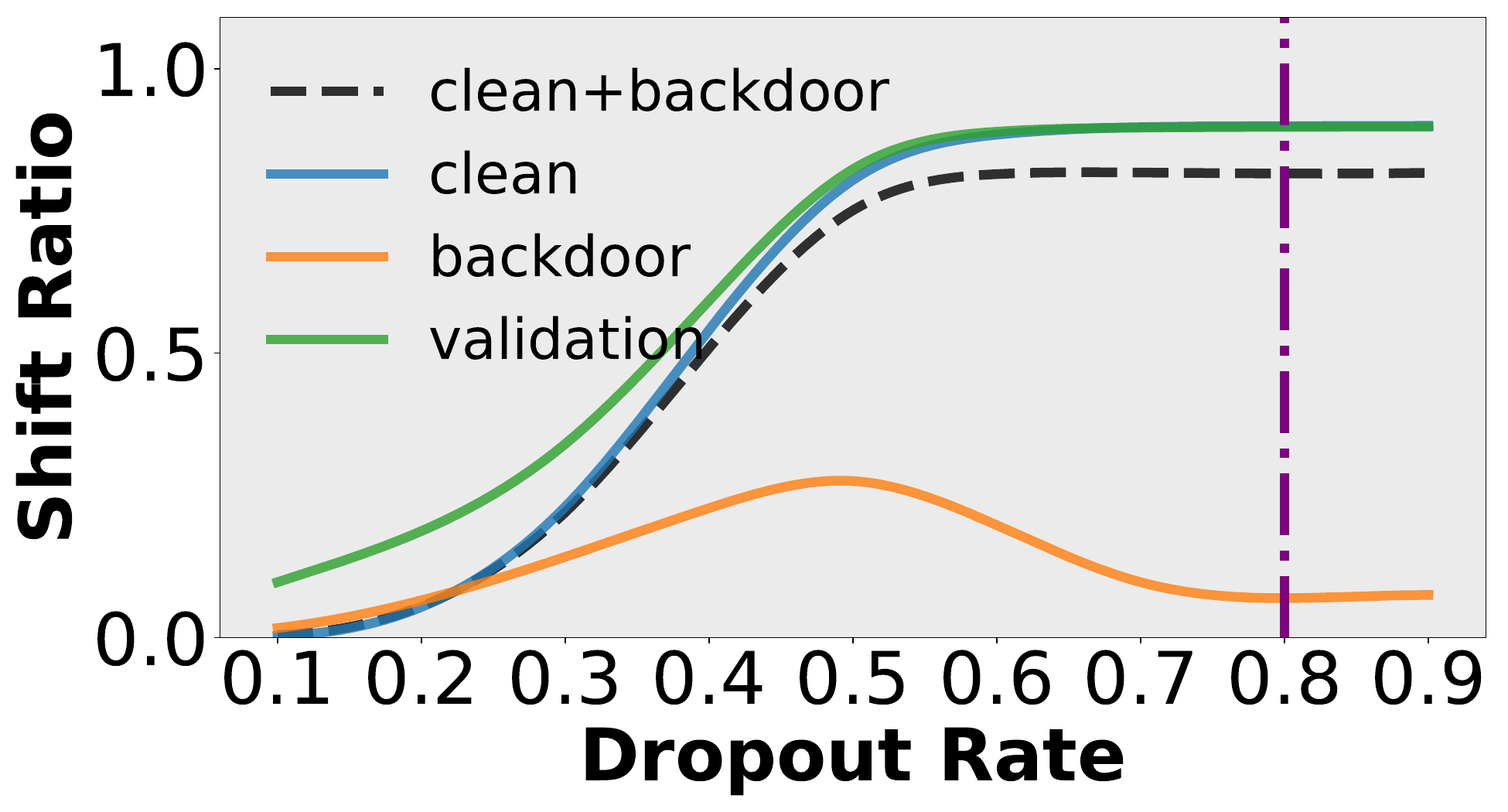}
     \end{subfigure}
     \vfill
     \begin{subfigure}[h]{0.3\textwidth}
         \centering
         \includegraphics[width=\textwidth]{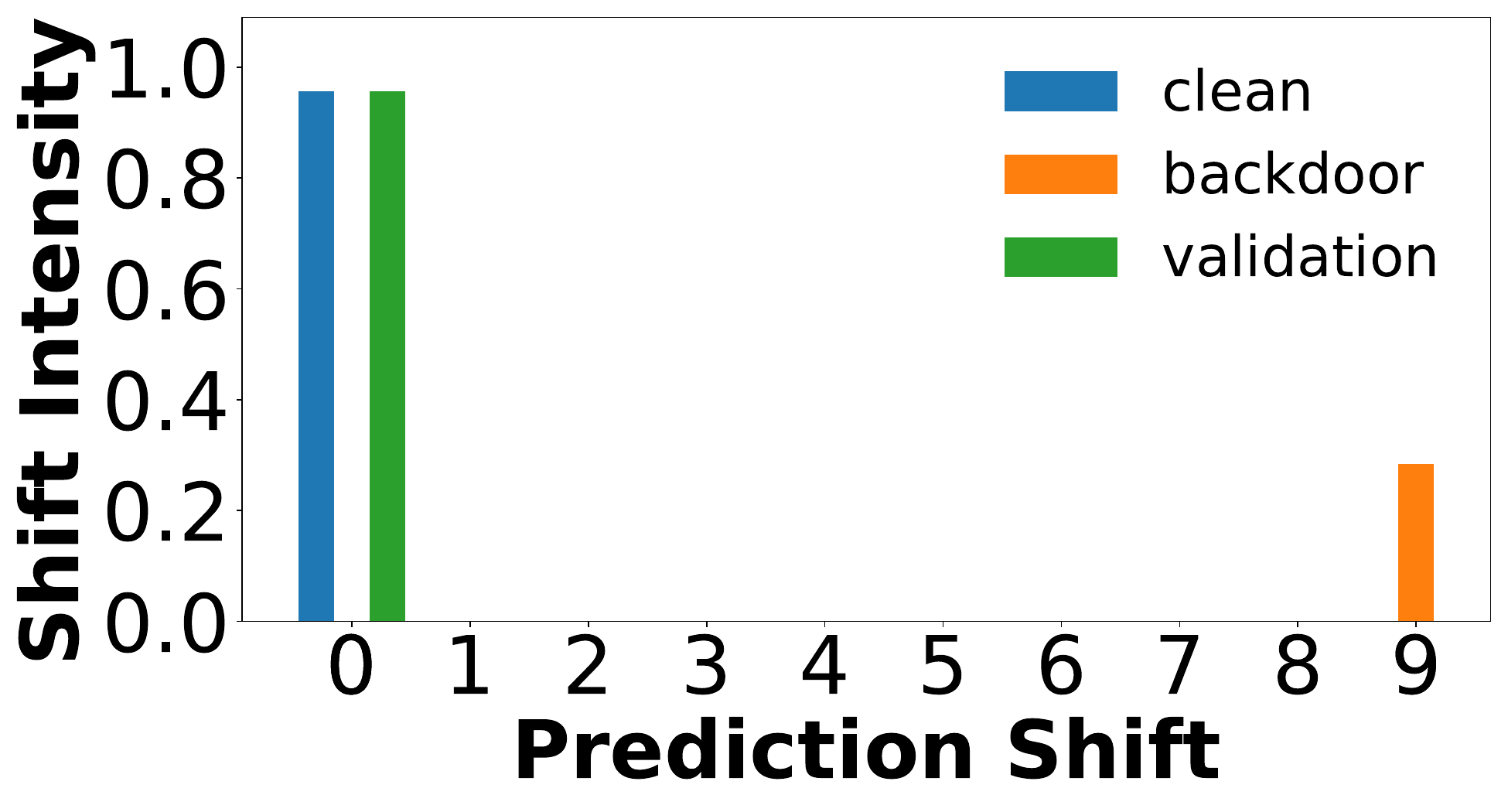}
         \caption{Blend}
     \end{subfigure}
     \hfill
     \begin{subfigure}[h]{0.3\textwidth}
         \centering
         \includegraphics[width=\textwidth]{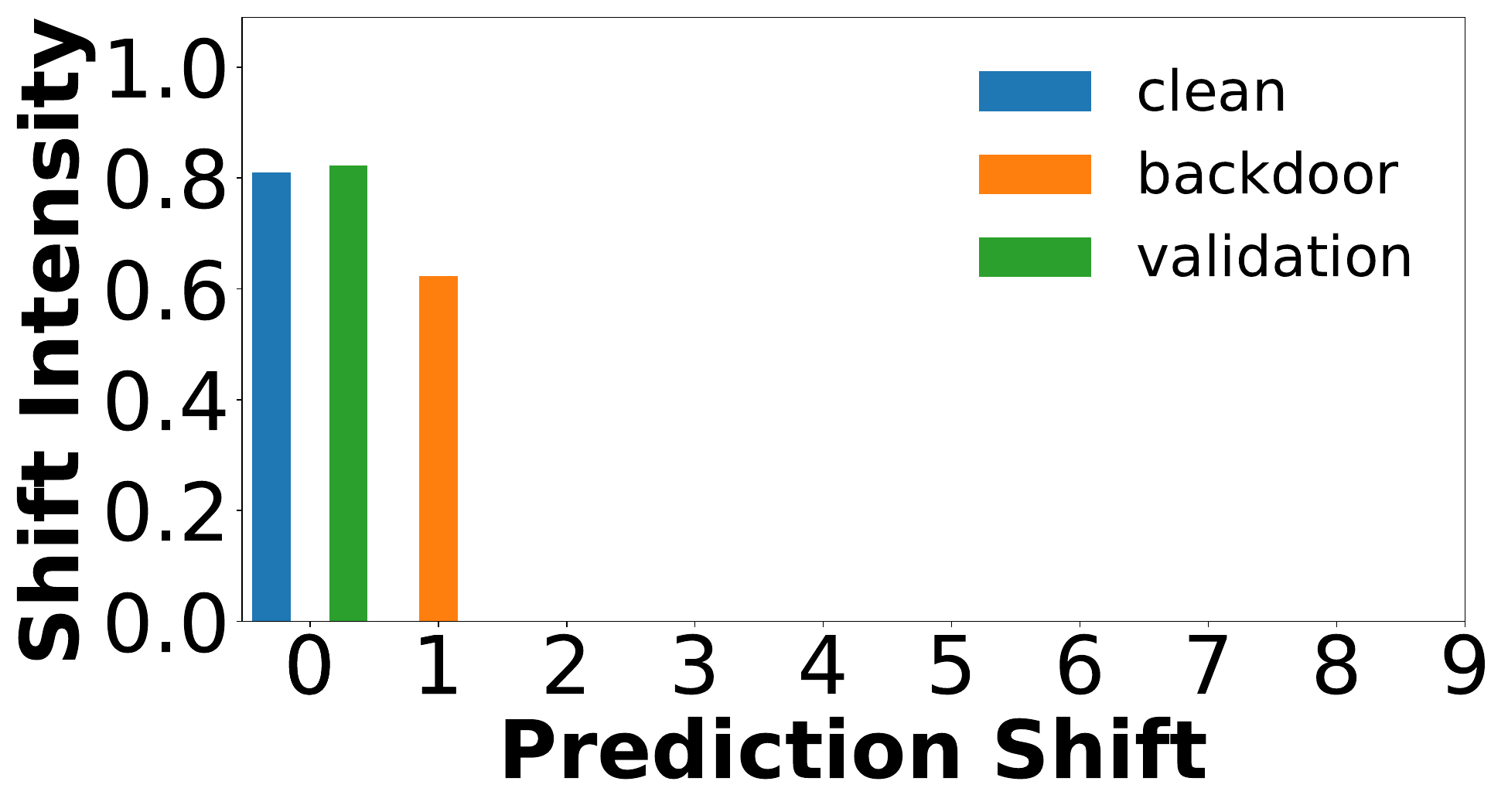}
         \caption{TrojanNN}
     \end{subfigure}
     \hfill
     \begin{subfigure}[h]{0.3\textwidth}
         \centering
         \includegraphics[width=\textwidth]{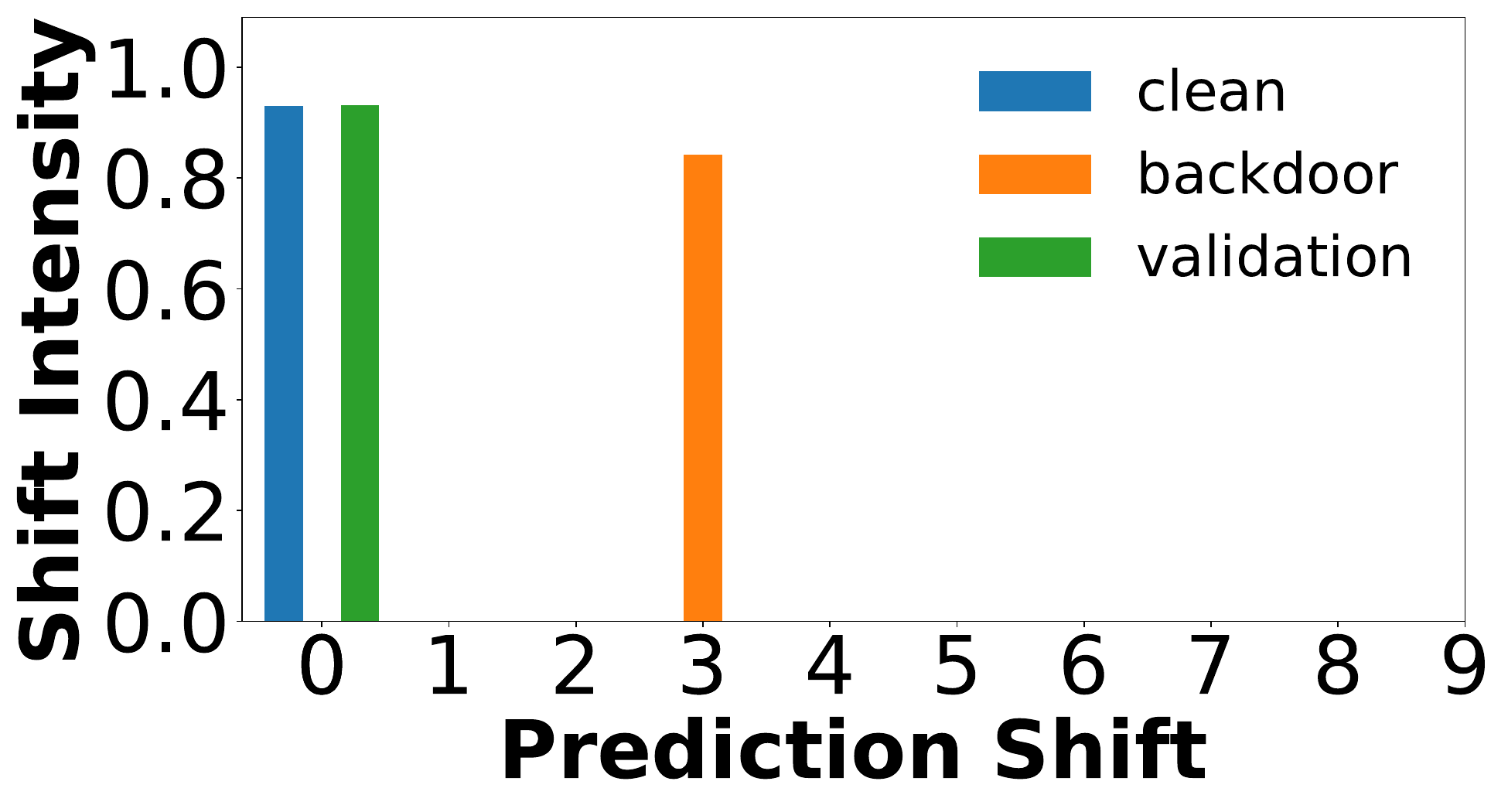}
         \caption{Label-Consistence}
     \end{subfigure}
     \vfill
     \vspace{1mm}
     \begin{subfigure}[h]{0.3\textwidth}
         \centering
         \includegraphics[width=\textwidth]{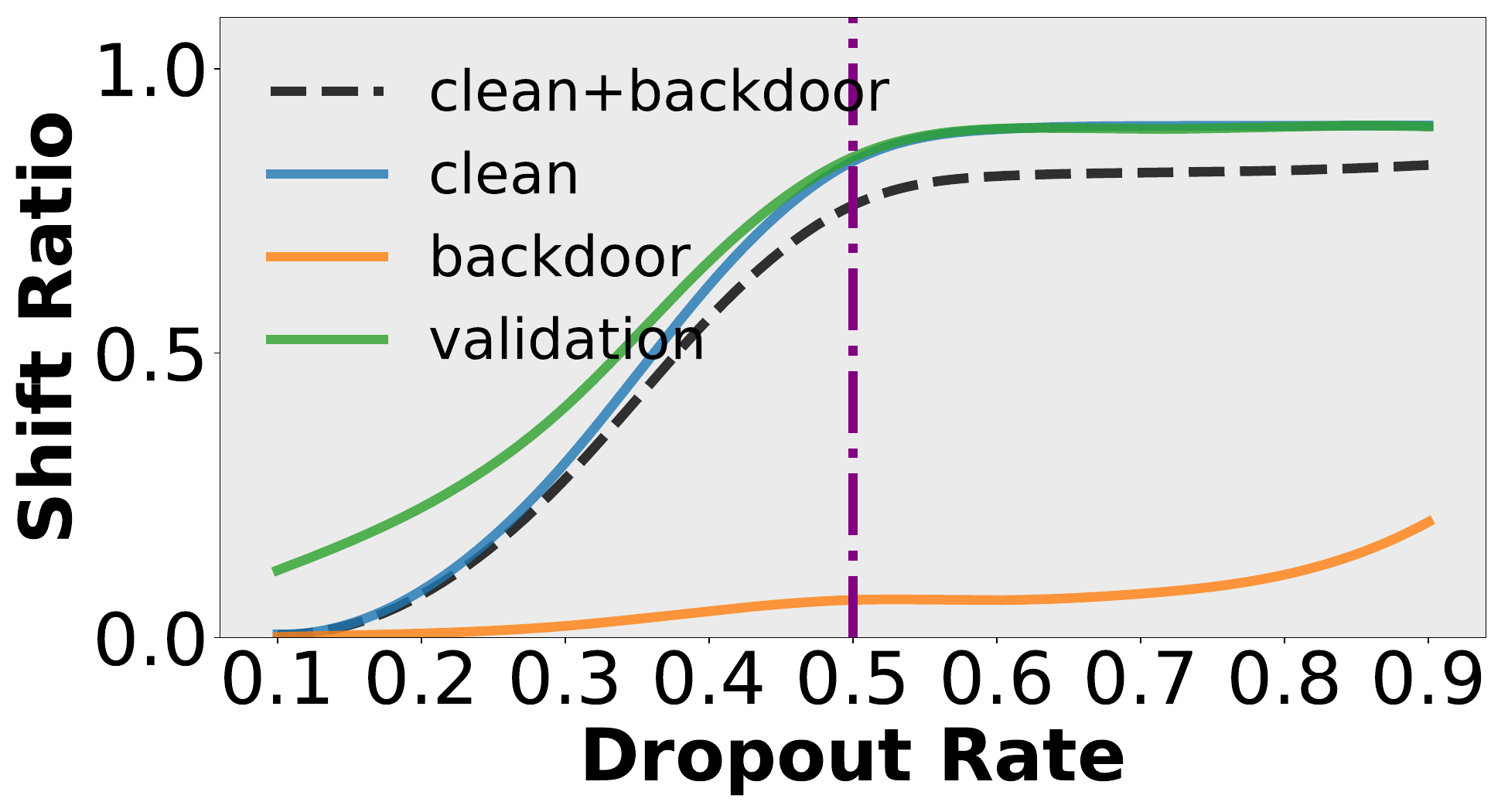}
     \end{subfigure}
     \hspace{5mm}
     \begin{subfigure}[h]{0.3\textwidth}
         \centering
         \includegraphics[width=\textwidth]{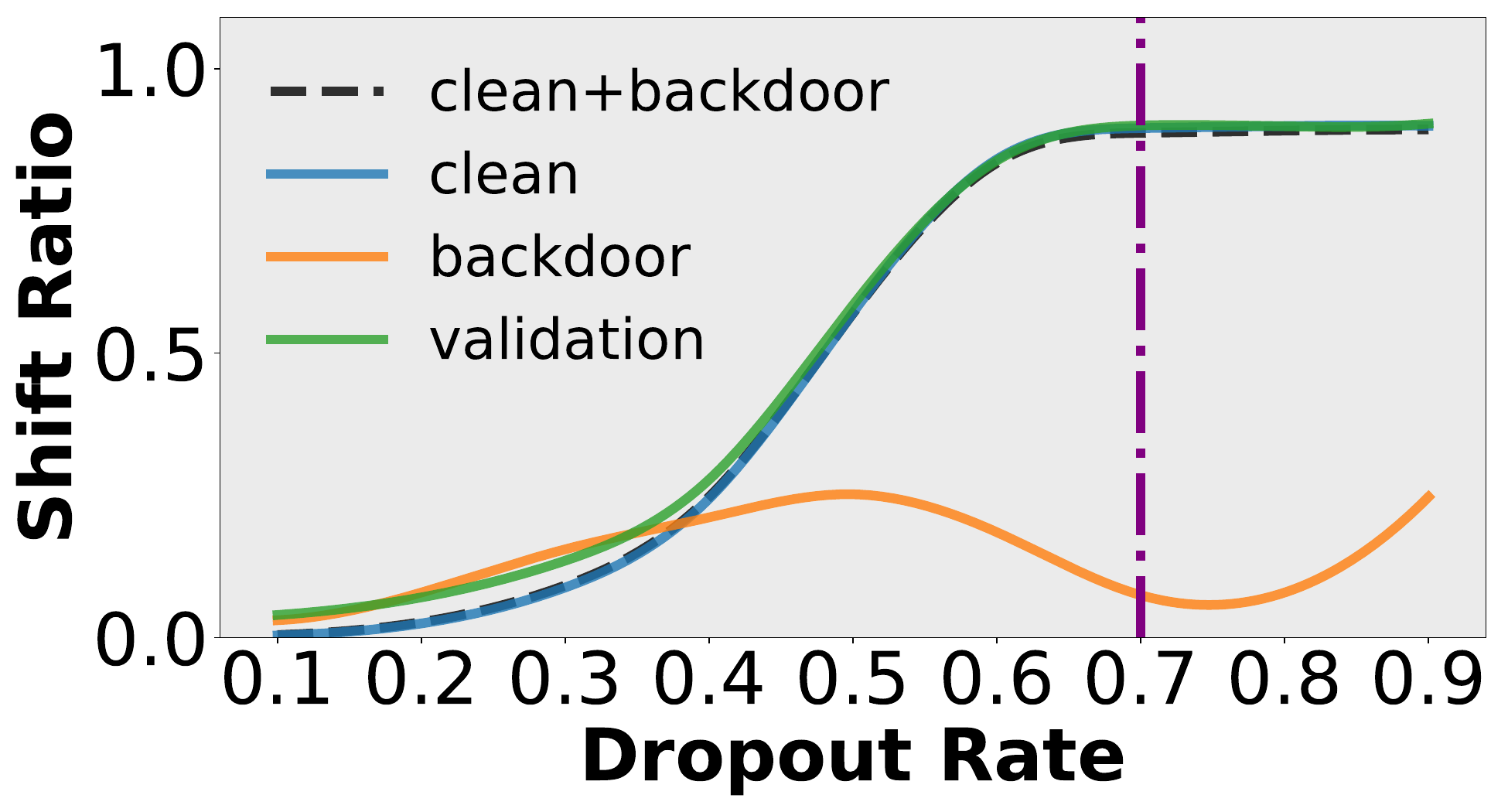}
     \end{subfigure}
     \vfill
     \begin{subfigure}[h]{0.3\textwidth}
         \centering
         \includegraphics[width=\textwidth]{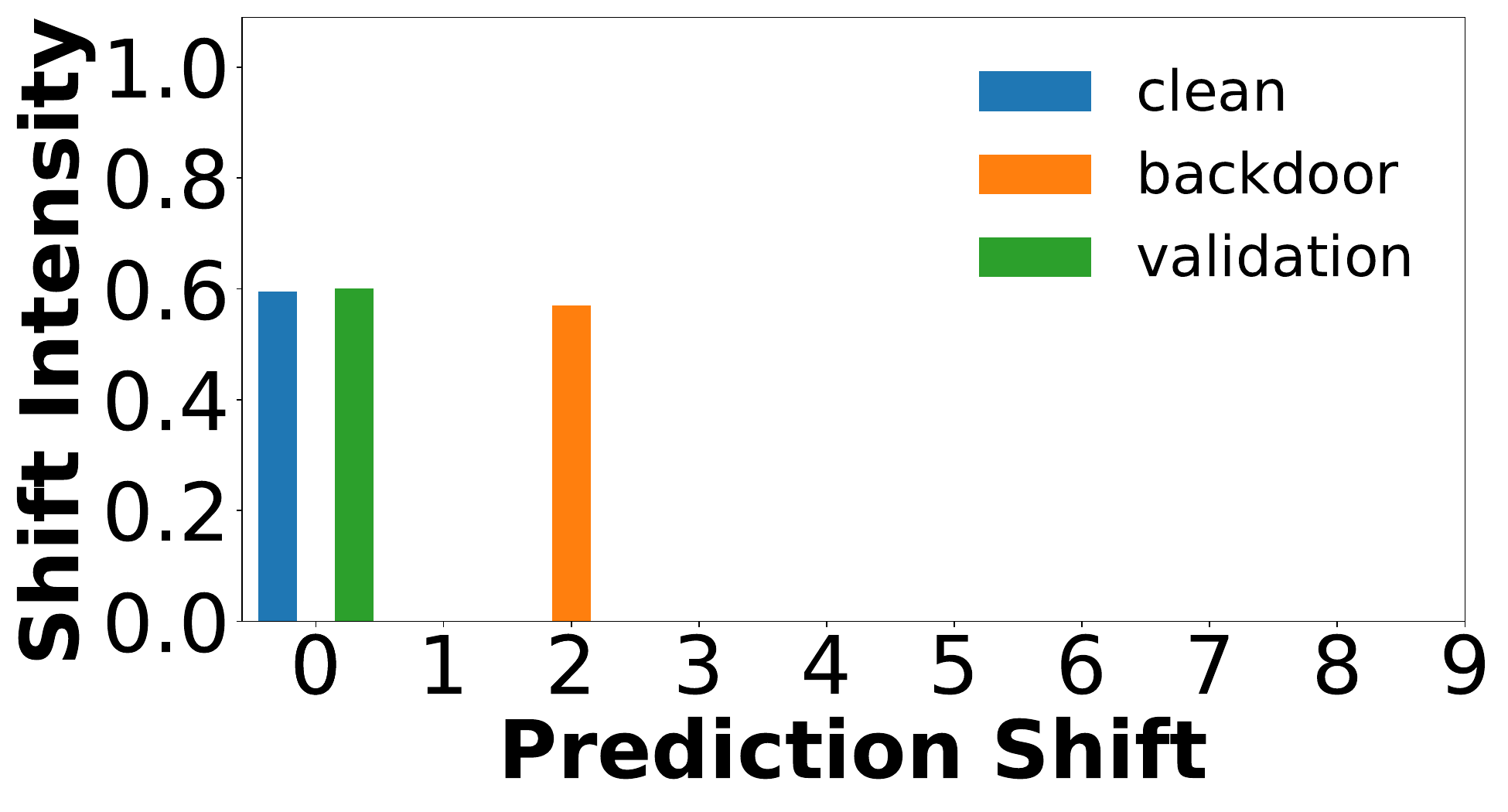}
         \caption{ISSBA}
     \end{subfigure}
     \hspace{5mm}
     \begin{subfigure}[h]{0.3\textwidth}
         \centering
         \includegraphics[width=\textwidth]{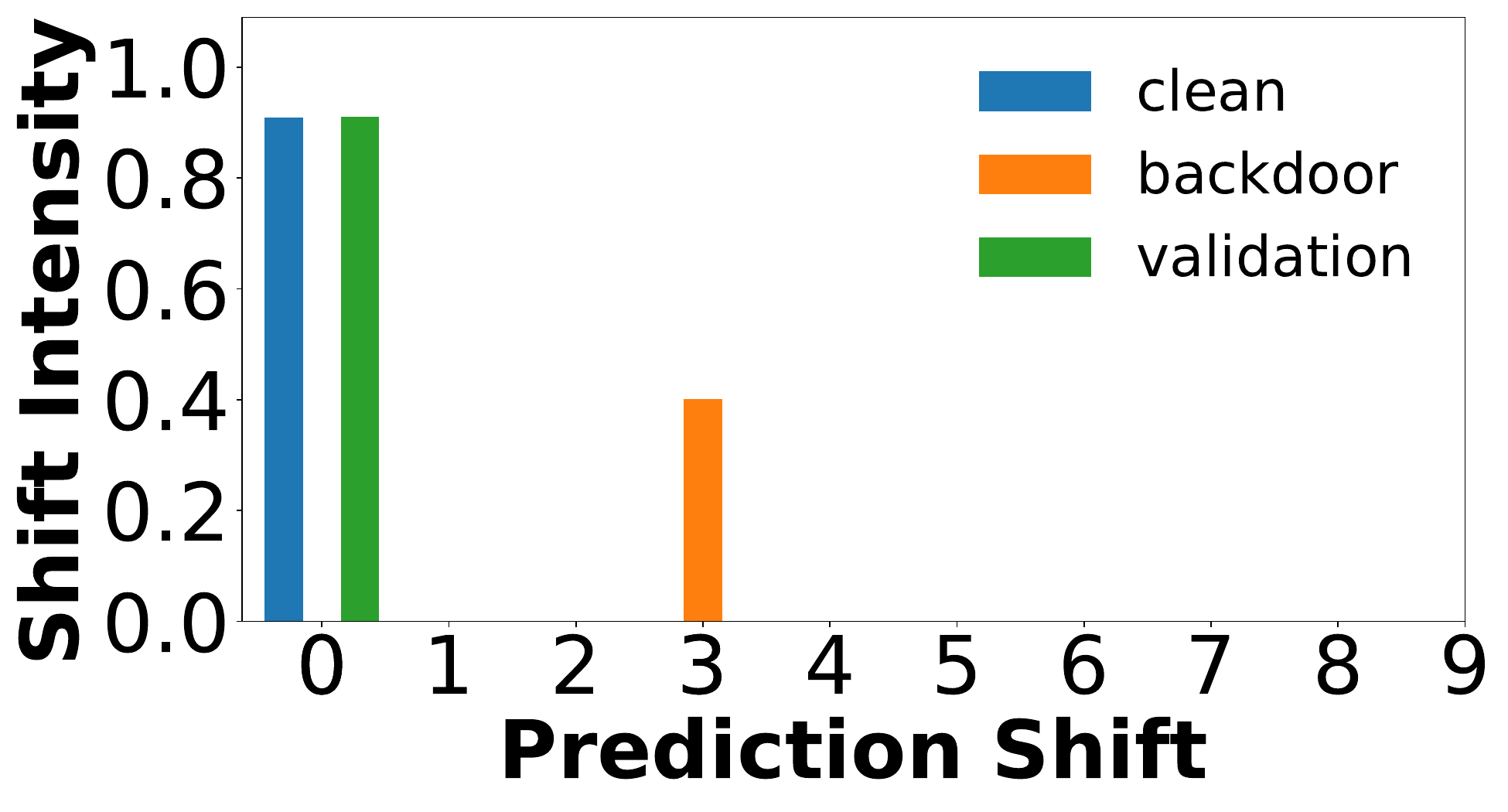}
         \caption{Adaptive-Blend}
     \end{subfigure}
    \caption{The shift ratio curves and shift intensity for more poisoned models.}
    \label{fig:PS on more poison appendix}
\end{figure*}
\begin{figure*}[h]
    \centering
     \begin{subfigure}[h]{0.3\textwidth}
         \centering
         \includegraphics[width=\textwidth]{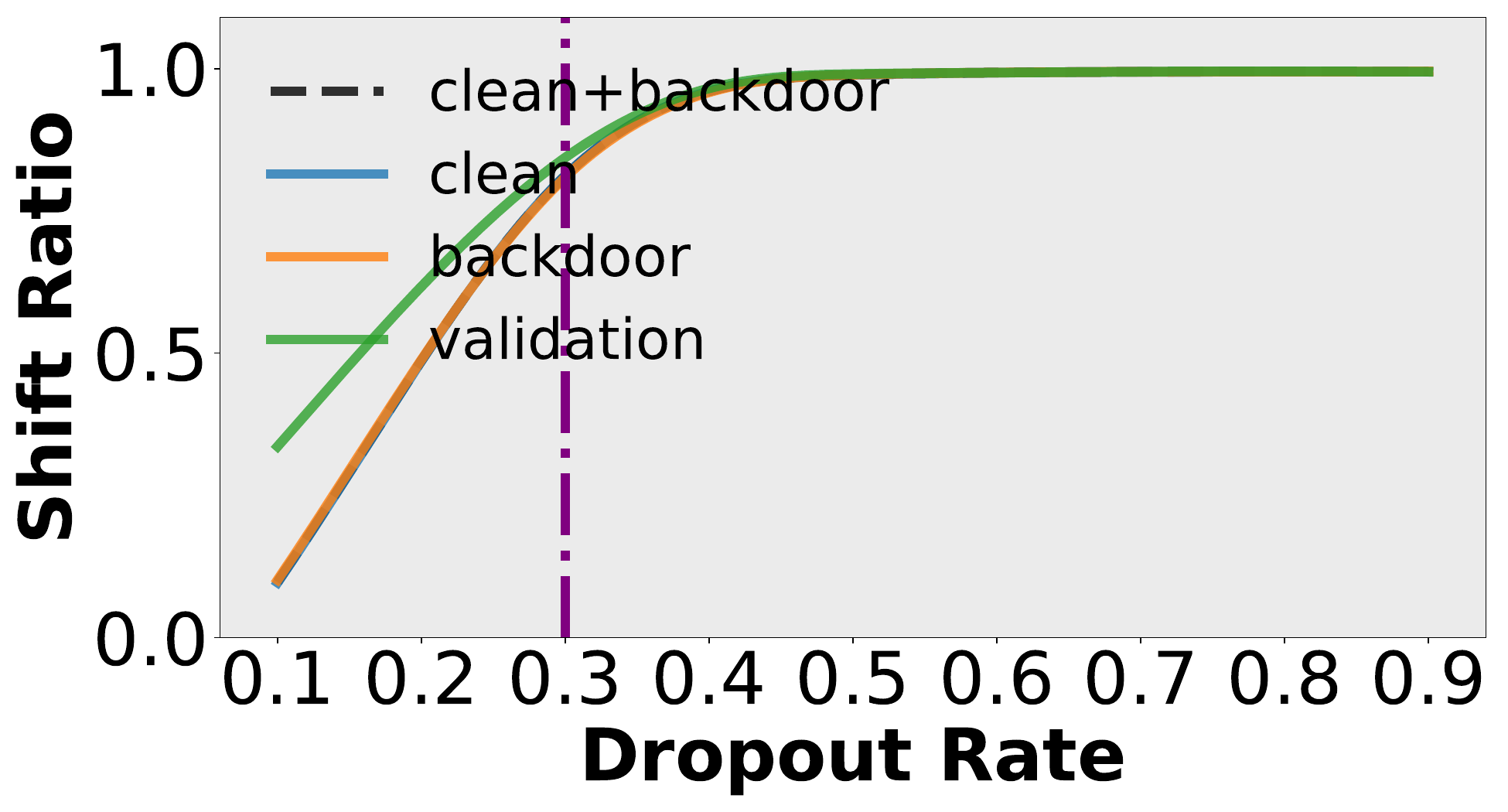}
     \end{subfigure}
     \hfill
     \begin{subfigure}[h]{0.3\textwidth}
         \centering
         \includegraphics[width=\textwidth]{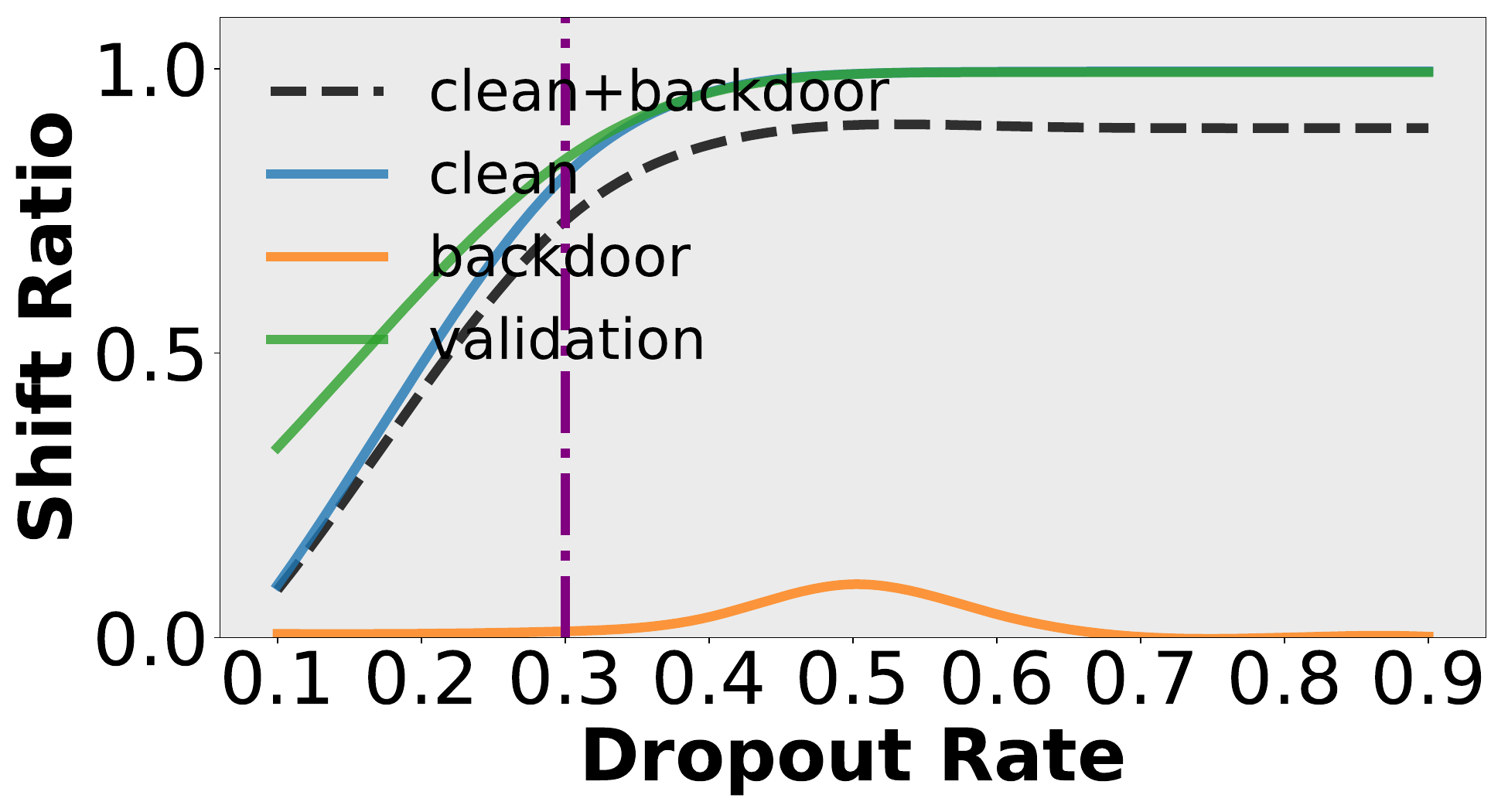}
     \end{subfigure}
     \hfill
     \begin{subfigure}[h]{0.3\textwidth}
         \centering
         \includegraphics[width=\textwidth]{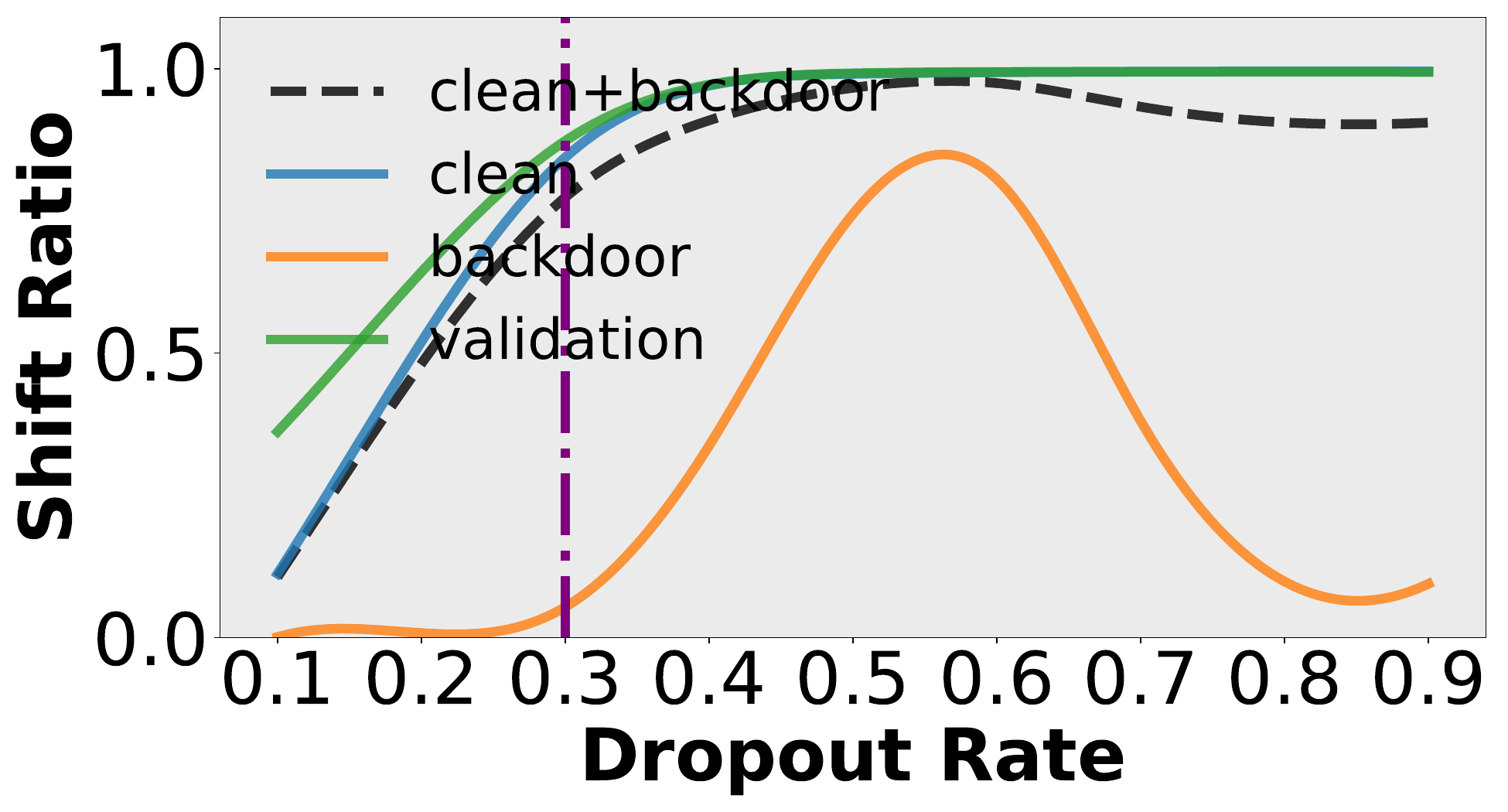}
     \end{subfigure}
     \vfill
     \begin{subfigure}[h]{0.3\textwidth}
         \centering
         \includegraphics[width=\textwidth]{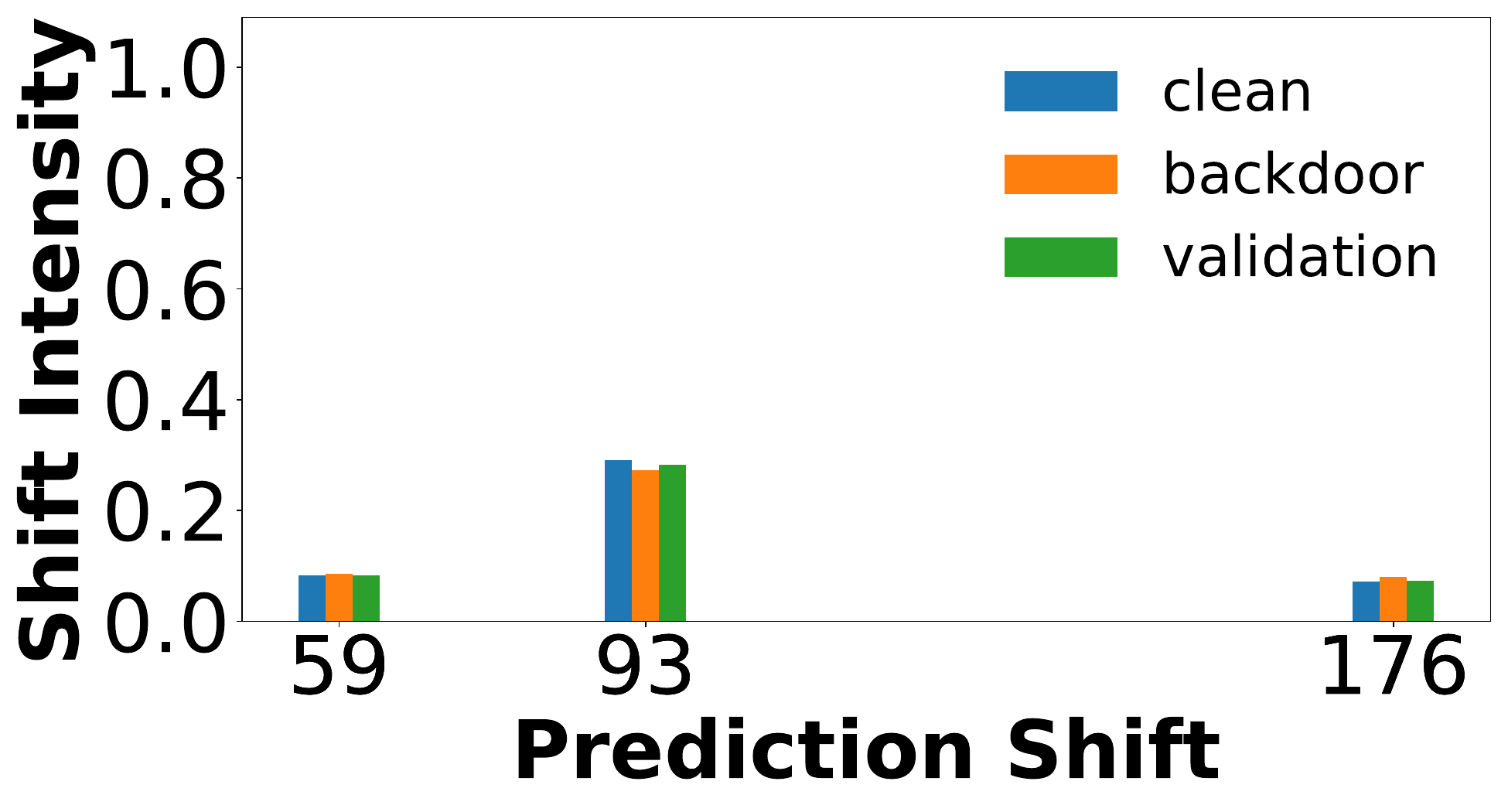}
         \caption{Benign Model}
     \end{subfigure}
     \hfill
     \begin{subfigure}[h]{0.3\textwidth}
         \centering
         \includegraphics[width=\textwidth]{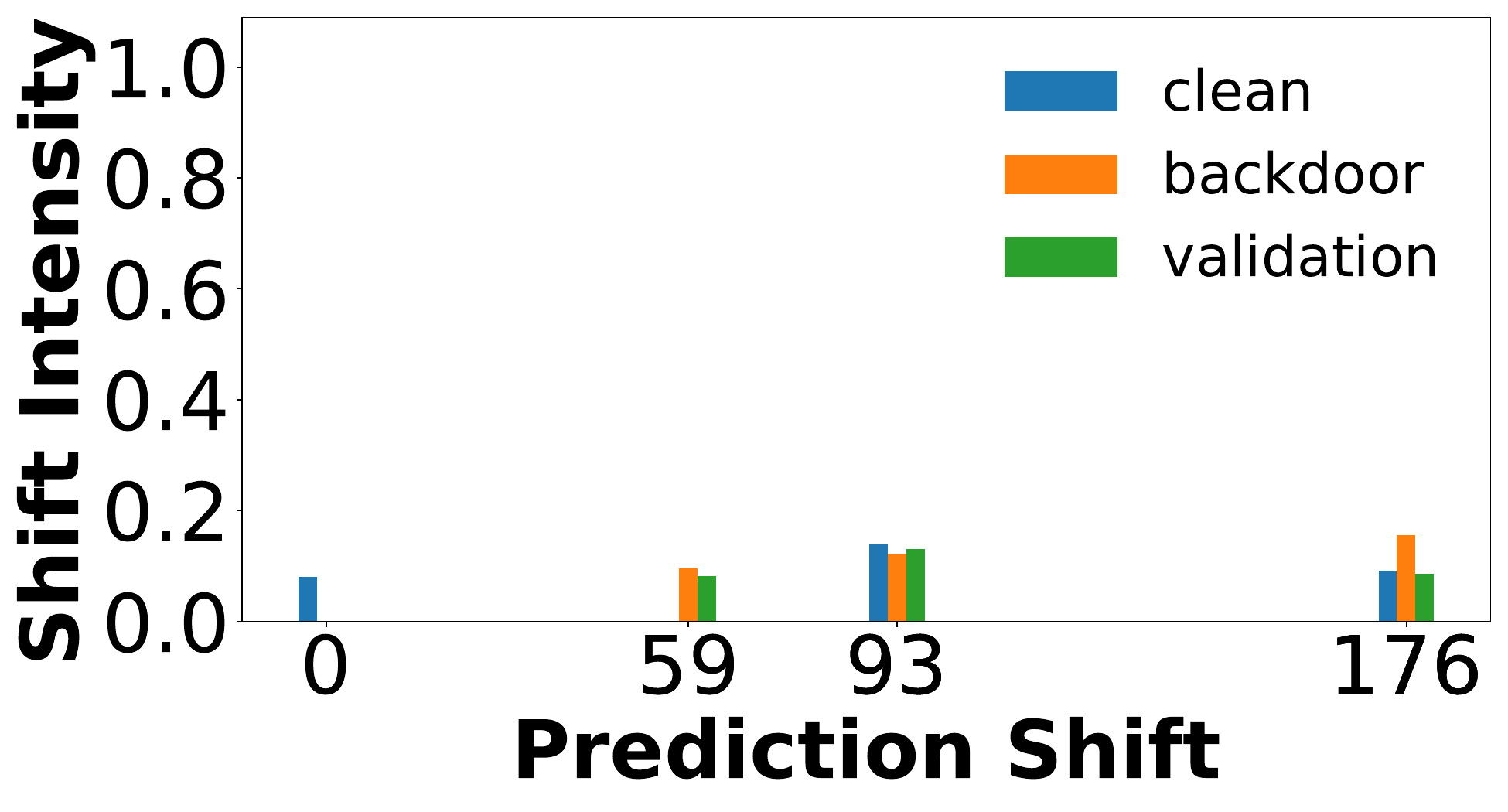}
         \caption{BadNets}
     \end{subfigure}
     \hfill
     \begin{subfigure}[h]{0.3\textwidth}
         \centering
         \includegraphics[width=\textwidth]{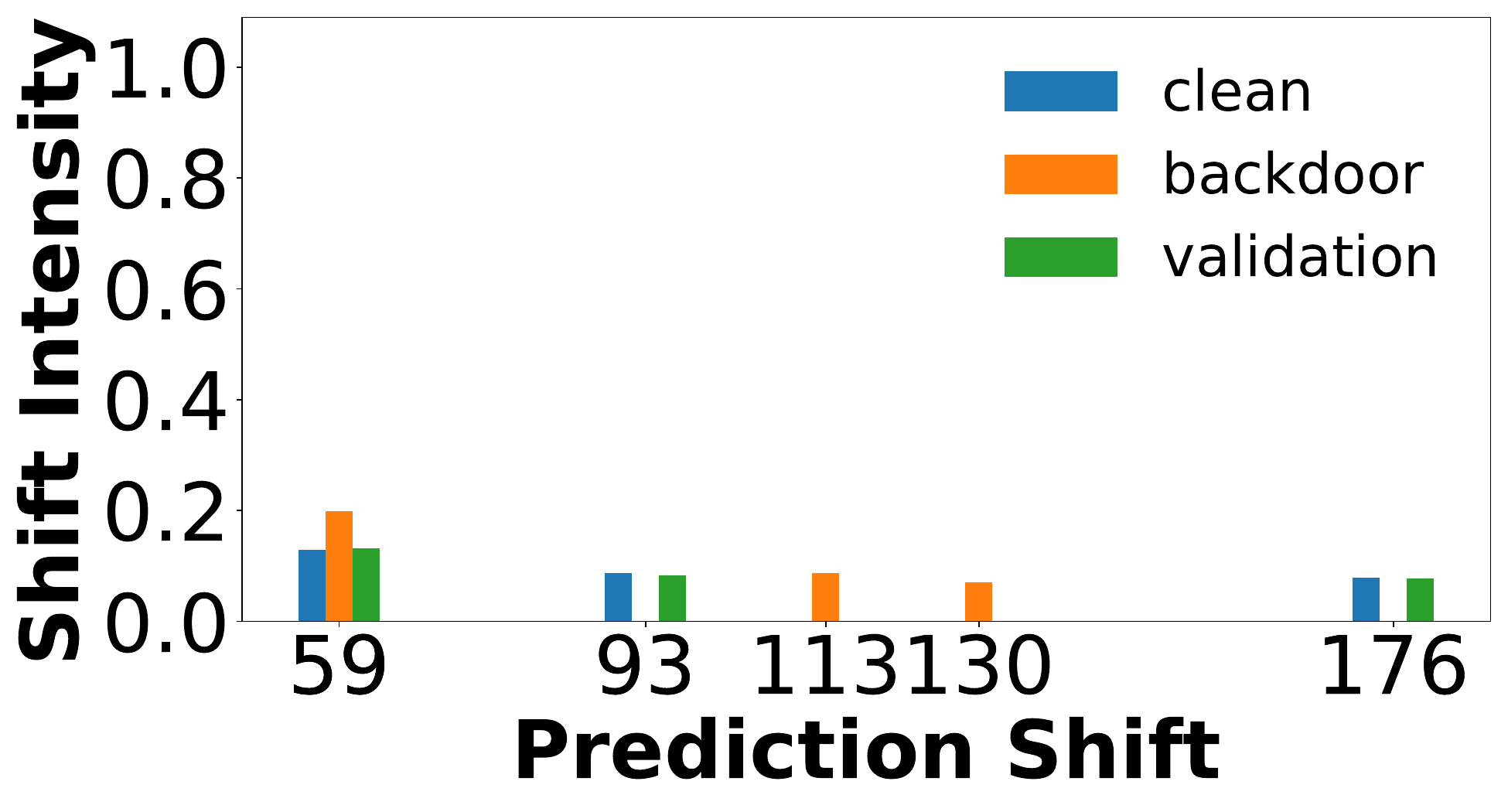}
         \caption{WaNet}
     \end{subfigure}
    \caption{The shift ratio curves and shift intensity on Tiny ImageNet for the benign model, BadNets model, and WaNet model, respectively. Please note thet we only present the results for the top three classes with the highest shift intensity values on Tiny ImageNet.}
    \label{fig:PS on tiny appendix}
\end{figure*}
\begin{figure*}[h]
    \centering
     \begin{subfigure}[h]{0.3\textwidth}
         \centering
         \includegraphics[width=\textwidth]{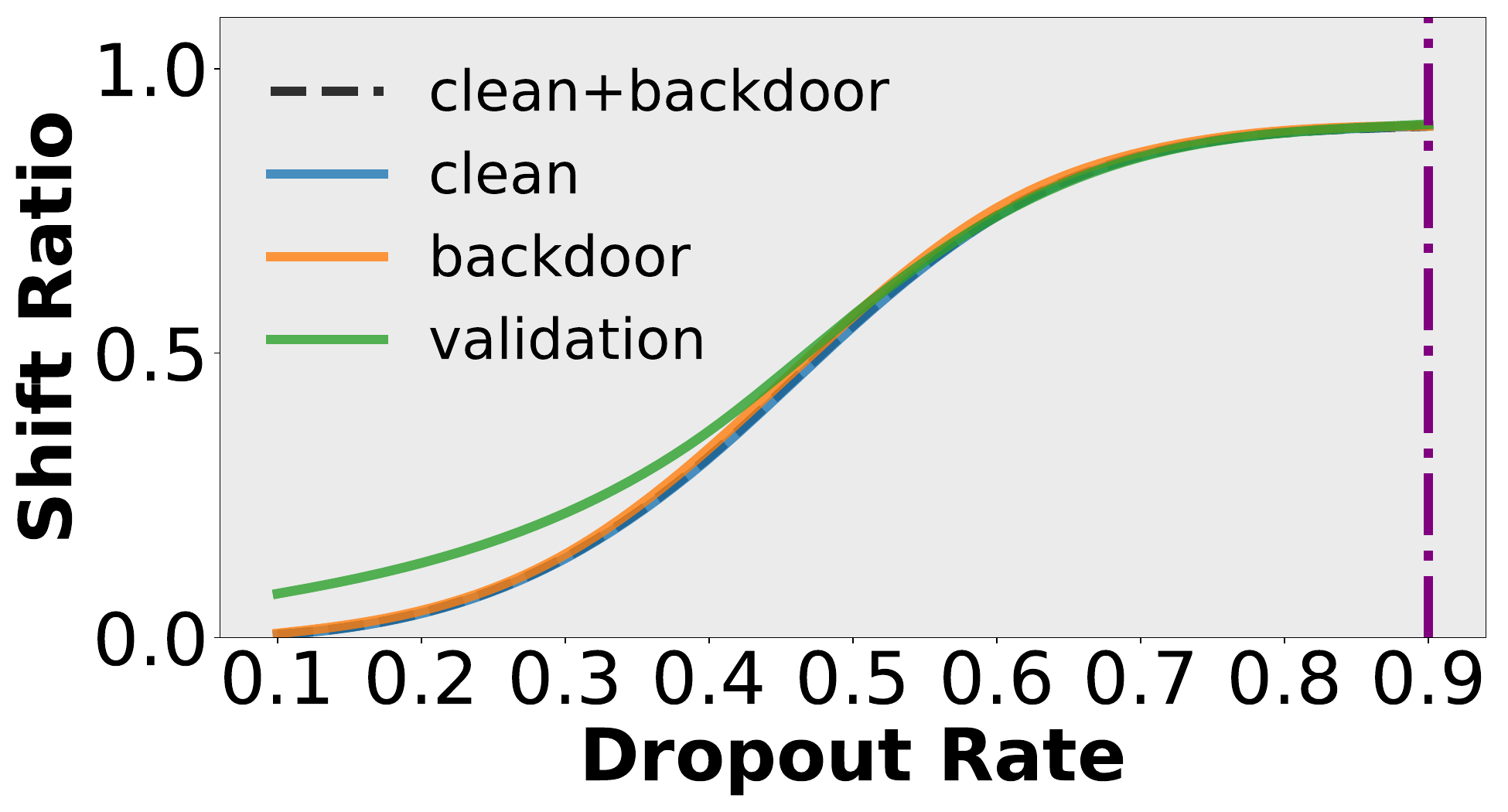}
     \end{subfigure}
     \hfill
     \begin{subfigure}[h]{0.3\textwidth}
         \centering
         \includegraphics[width=\textwidth]{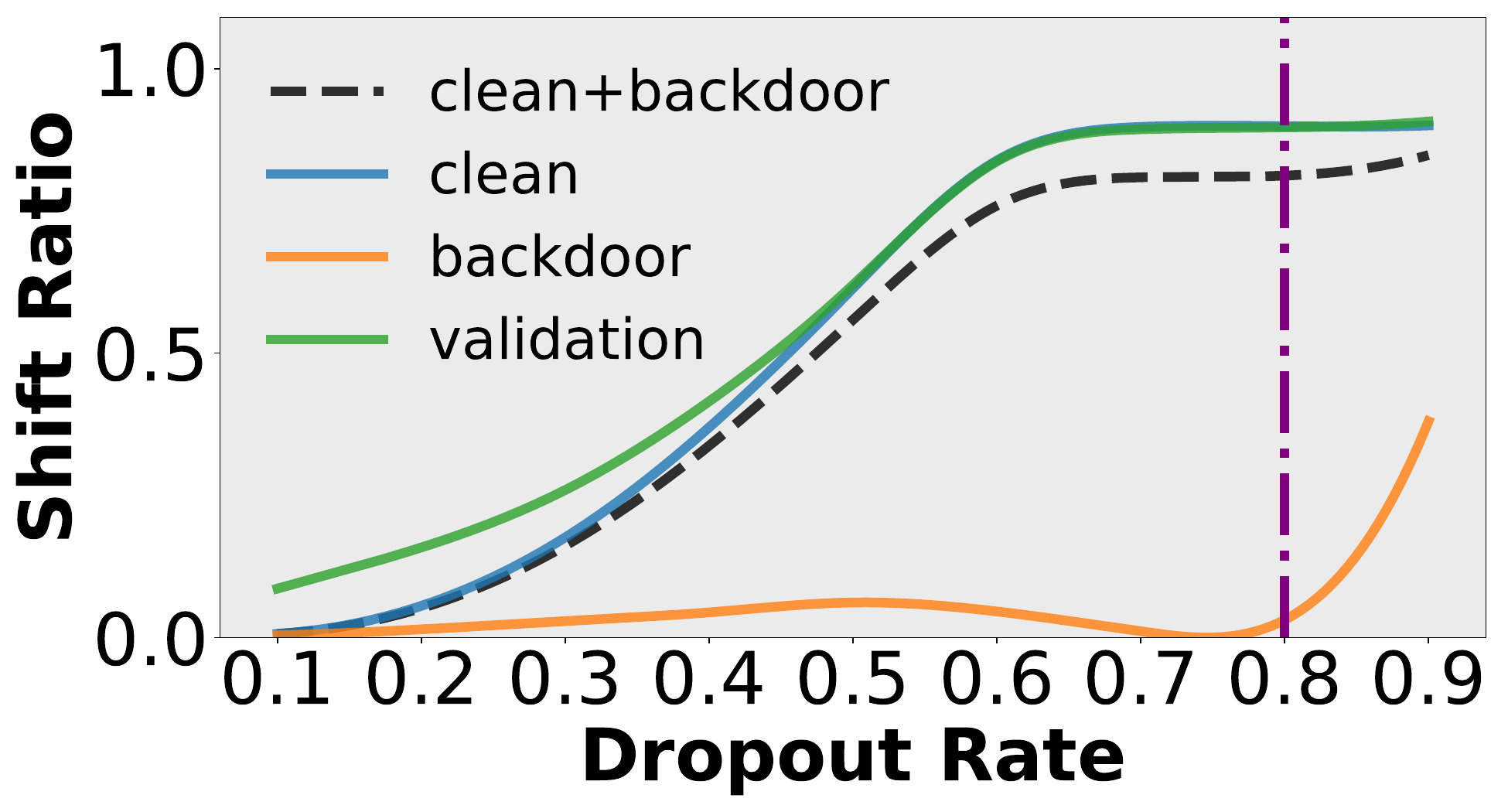}
     \end{subfigure}
     \hfill
     \begin{subfigure}[h]{0.3\textwidth}
         \centering
         \includegraphics[width=\textwidth]{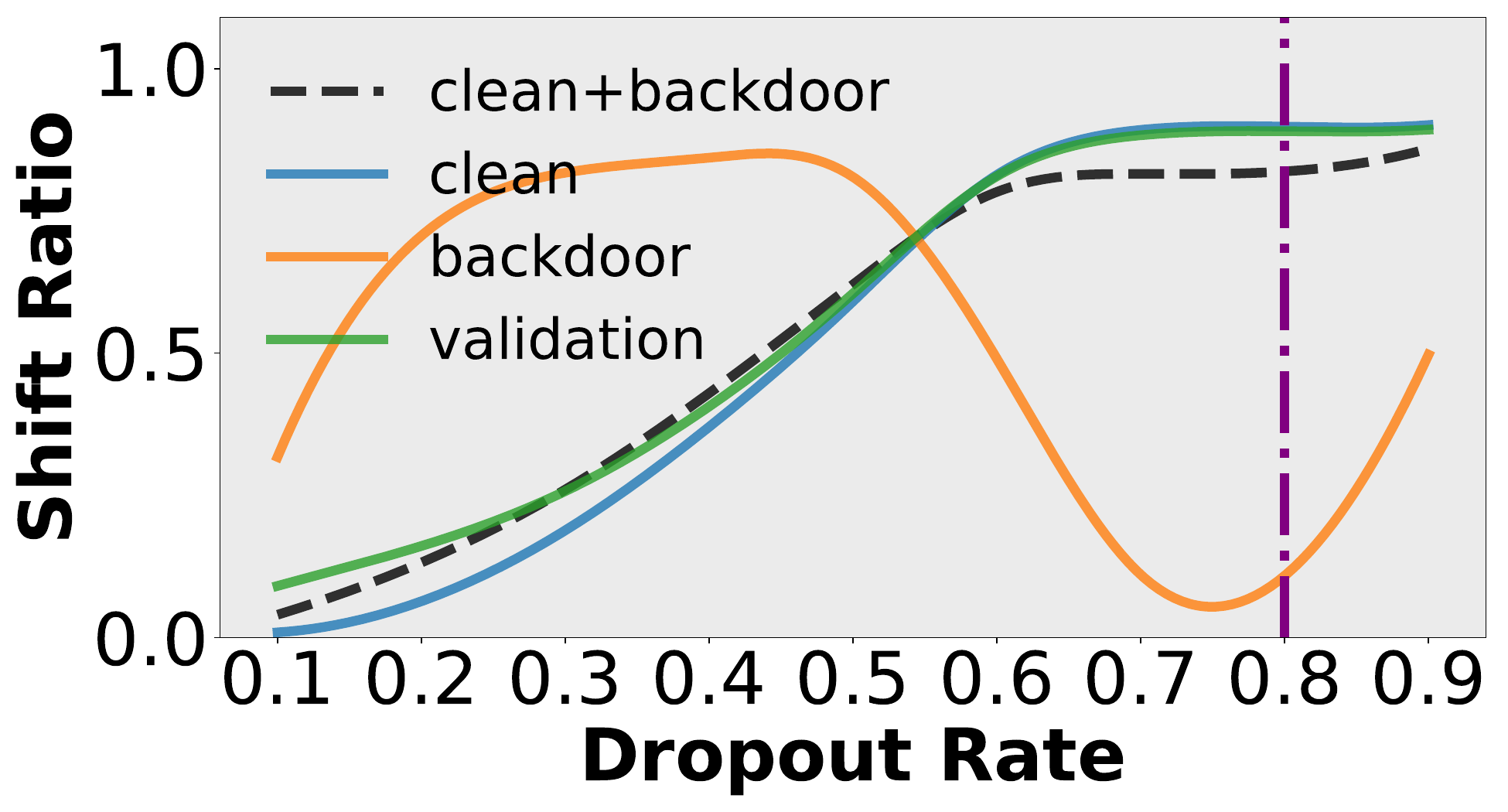}
     \end{subfigure}
     \vfill
     \begin{subfigure}[h]{0.3\textwidth}
         \centering
         \includegraphics[width=\textwidth]{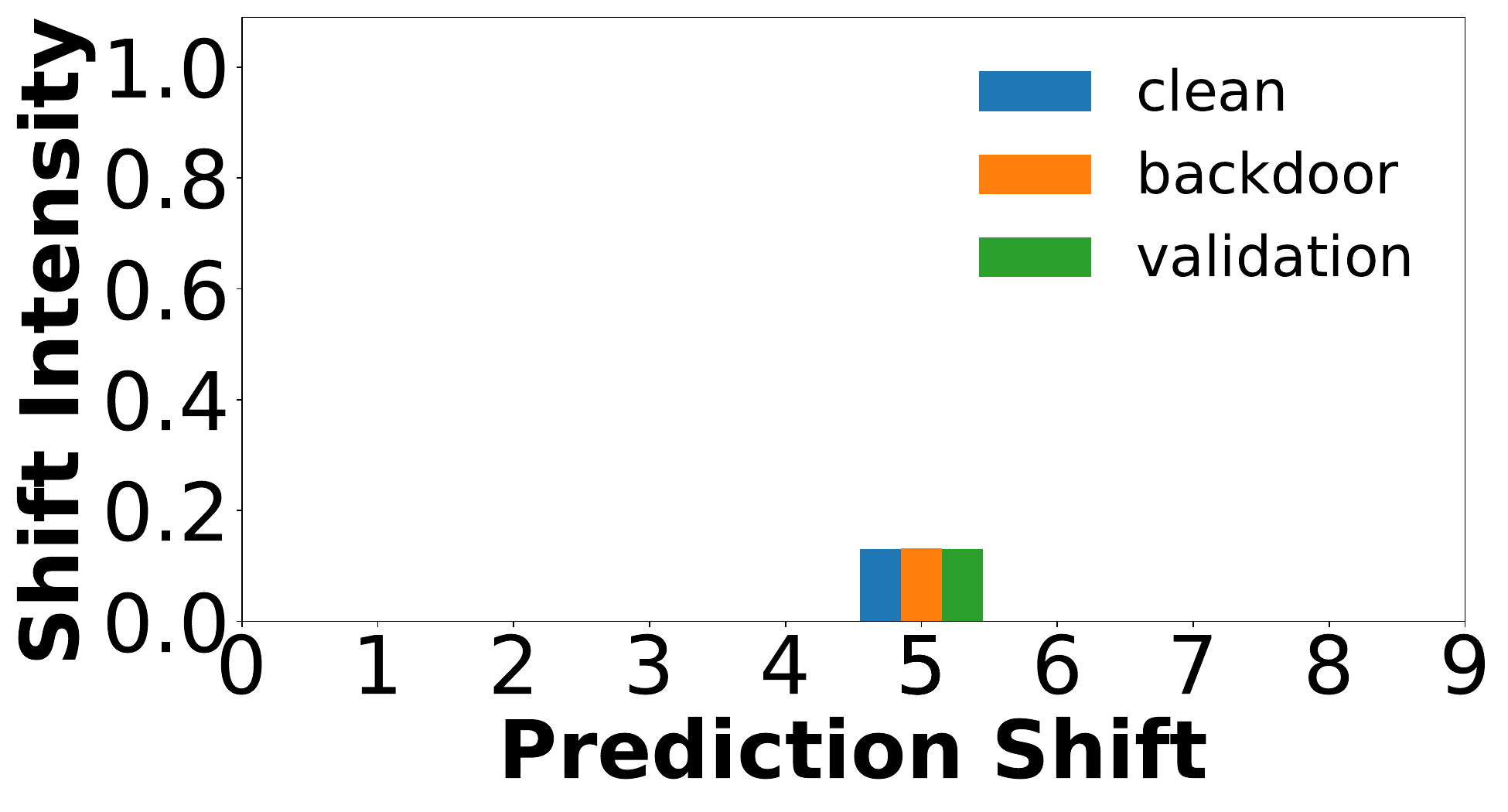}
         \caption{Benign Model}
     \end{subfigure}
     \hfill
     \begin{subfigure}[h]{0.3\textwidth}
         \centering
         \includegraphics[width=\textwidth]{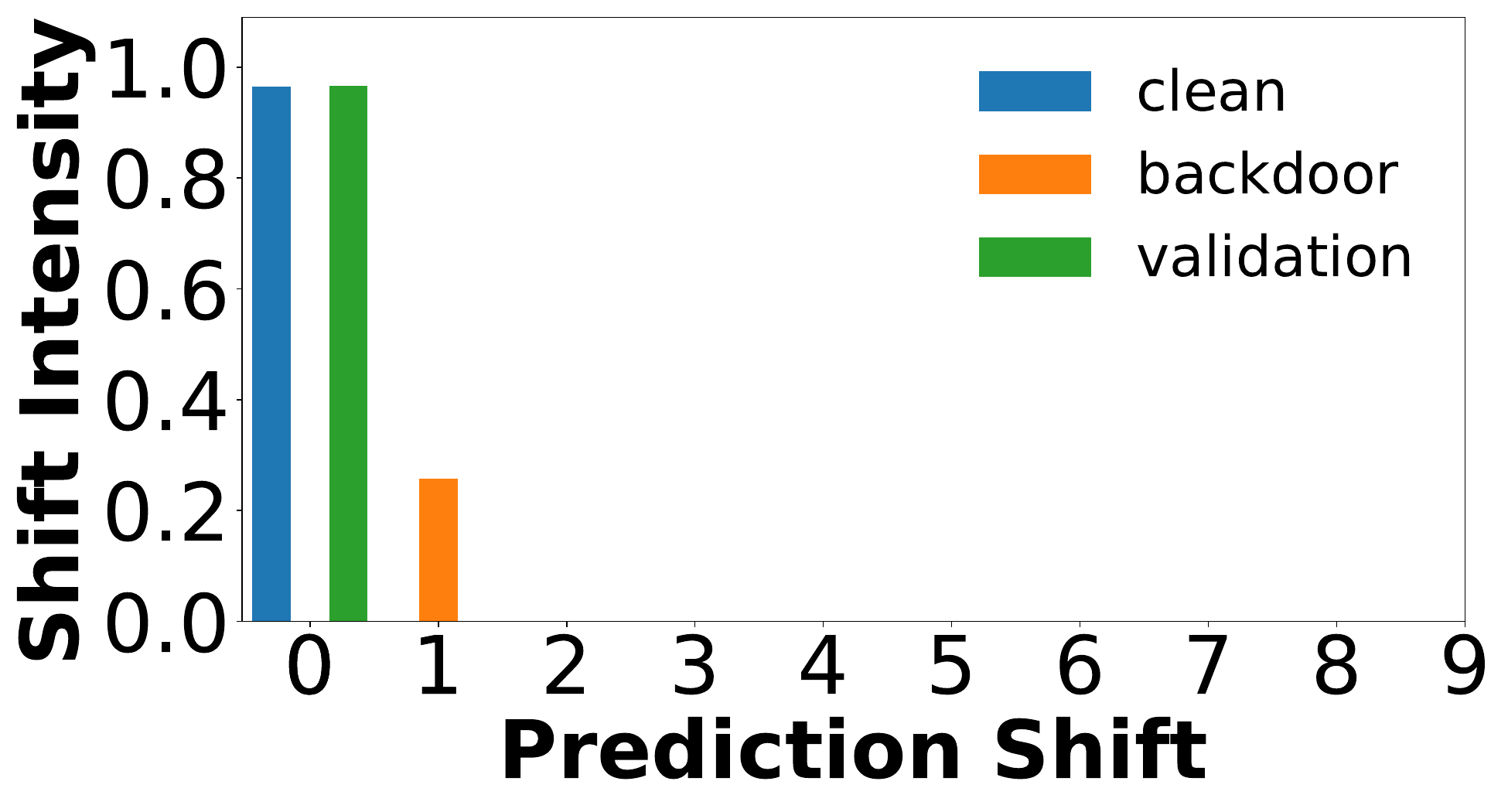}
         \caption{BadNets}
     \end{subfigure}
     \hfill
     \begin{subfigure}[h]{0.3\textwidth}
         \centering
         \includegraphics[width=\textwidth]{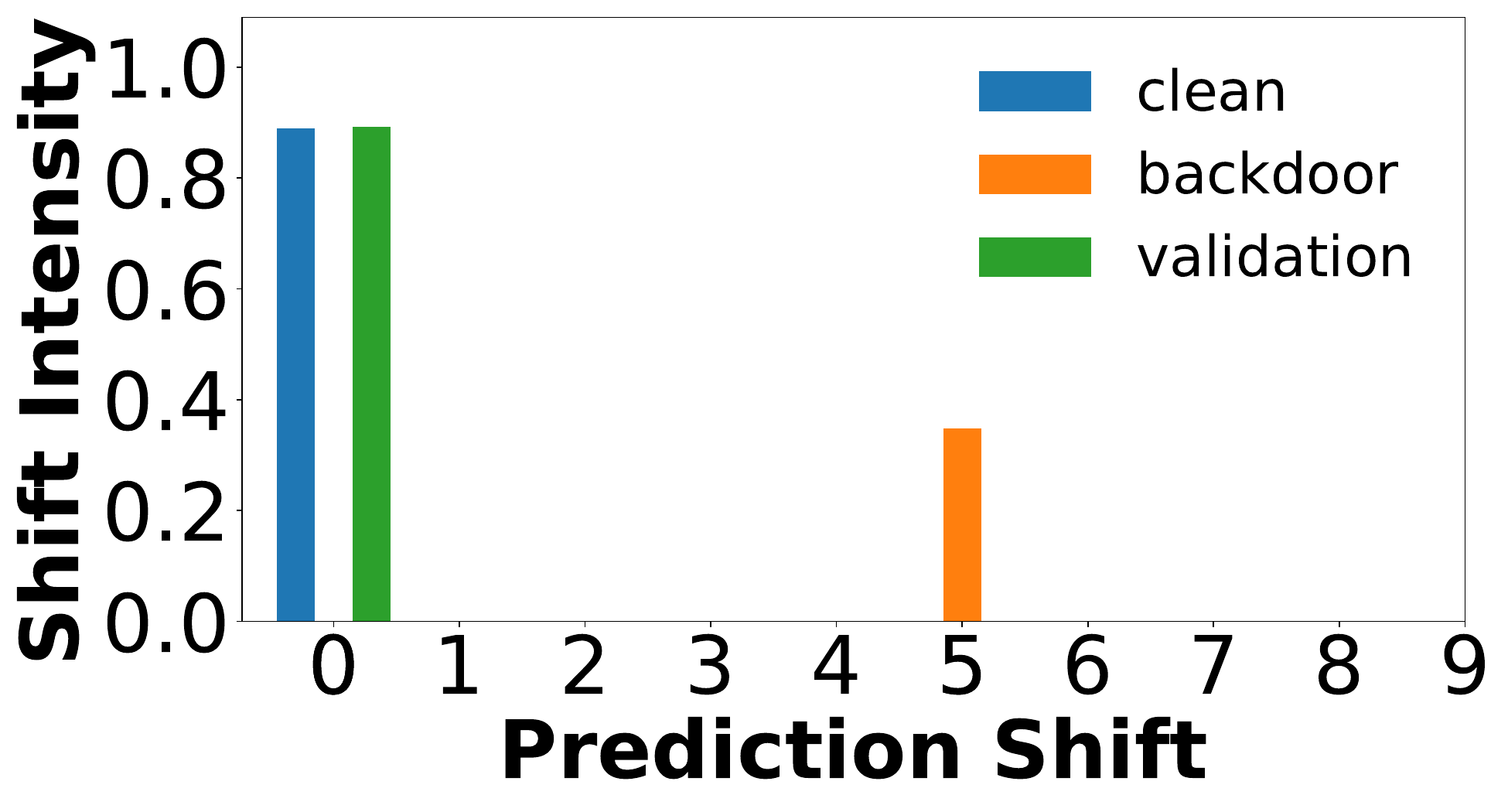}
         \caption{WaNet}
     \end{subfigure}
    \caption{The shift ratio curves and shift intensity used VGG16-bn architecture for the benign model, BadNets model, and WaNet model, respectively.}
    \label{fig:PS on vgg appendix}
\end{figure*}
As illustrated in Figure~\ref{fig:pilot study 2}, under most attack scenarios, we observe an increase in average uncertainty for both clean training data and clean validation data, with their uncertainties nearly overlapping. This will enable us to better utilize the uncertainty of validation data to approximate the uncertainty of clean training data. From Figure~\ref{fig:pilot study 2 benign}, we can observe that the average uncertainty of benign model on the three data types is significantly increasing after scaling pixel values. This indicates that the introduction of input-level uncertainty indeed enhances the model predictive uncertainty further, and its strength can be easily controlled by adjusting the multiplicative factor of SCP.

However, under WaNet scenario, one can see from Figure~\ref{fig:pilot study 2 wanet} that although the uncertainty of the clean training data becomes closer to that of the clean validation data after scaling, the uncertainty gap between clean training and backdoor training data further reduces. Furthermore, as shown in Figure~\ref{fig:pilot study 2 lc} and Figure~\ref{fig:pilot study 2 adaptive-blend},
under the Label-Consistence\cite{turner2019label} and the Adaptive-Blend attack scenarios, the backdoor training data exhibits 
only slightly smaller uncertainty than that of clean training data, which poses significant challenges for their differentiation. 

Backdoor training data exhibits the same uncertainty as both clean training data and clean validation data, indicating a failure in the combination of MC-Dropout uncertainty and input-level uncertainty. In addition, the scaling factor is a parameter that is difficult to ascertain when we have a lack of the knowledge about backdoor attack. Therefore, we cannot directly utilize this method.

\section{Prediction Shift Phenomenon on More Scenarios}
In this section, we will show the more results of PS phenomenon on the more poisoned models, more architectures, and more dataset. It demonstrates the broad applicability of our approach in diverse real-world scenarios. We maintain the same experimental setup with Section~\ref{PS}.
\subsection{PS Phenomenon on More Poisoned Models.}
\label{PS on more poison appendix}

As shown in Figure~\ref{fig:PS on more poison appendix}, the shift ratio curve of clean data maintains consistency across diverse attack scenarios, i.e., the shift ratio $\sigma$ increases with the growth of dropout rate $p$ and eventually stabilizes. Similarly, the shift ratio curve of backdoor data also exhibits certain consistency across various attack scenarios, always presenting a relatively lower $\sigma$ at a specific $p$. This suggests that the PS phenomenon and neuron bias effect do not depend on the specific type of backdoor attack, but rather are intrinsic properties of the model. Furthermore, different attack scenarios lead to distinct shift ratio curves of backdoor data.

Except for the ISSBA\cite{li2021invisible} attack, the shift intensity of clean data is not pronounced. In other attacks, the shift intensity of clean data is extremely strong. More importantly, in all attack scenarios, clean data exhibits a bias towards the target class (class 0 in our experiments). This further indicates that when the model has good generalizability, the neuron bias path established by the backdoor data in the model becomes more stable and specific. This property may be exploited in the future to detect the target class of backdoor attacks.

\subsection{PS Phenomenon on Tiny ImageNet Dataset.}
\label{PS on tiny appendix}

From the observation of Figure~\ref{fig:PS on tiny appendix}, we can find that the PS phenomenon and neuron bias effect persist even in the more complex Tiny ImageNet dataset\cite{russakovsky2015imagenet}.

The shift ratio curve trends for the three data types(clean training data, clean validation data, and backdoor training data) in the model trained on the Tiny ImageNet dataset remain consistent with the trends observed in the model trained on the CIFAR-10 dataset. Specifically, for the benign model, the shift ratio $\sigma$ still increases with the dropout rate $p$ and eventually stabilizes. For the poisoned models, BadNet and WaNet, the shift ratio curve for the backdoor data always presents a relatively lower $\sigma$ at a specific $p$.

The key difference is that the PS phenomenon is less pronounced in the models trained on the Tiny ImageNet dataset, as evidenced by a significant reduction in the shift intensity. Additionally, the shift class in these models tends to be biased towards certain specific classes, rather than the target class (class 0 in our experiments).

As the more complex features and larger number of classes in the Tiny ImageNet dataset, the model's generalization capacity may still be insufficient    , despite the use of data augmentation techniques.
We hypothesize that the inadequate generalization capability results in less stable and distinct neuron bias paths. This allows a relatively small $p$ to cause the neuron bias path to overweigh the normal feature path, resulting in the presence of PS phenomenon in clean data, but without a strong neuron bias towards the target class. Meanwhile, the backdoor data remains relatively stable and almost does not exhibit the PS phenomenon. Our method effectively leverages the key difference in the PS phenomenon between clean data and backdoor data to enable the effective detection of backdoor data.

\subsection{PS Phenomenon on VGG Architecture.}
\label{PS on vgg appendix}

Despite assuming that the defender can freely choose the model architecture, we also conducted experiments using the VGG16-bn\cite{simonyan2014very} model to show that our method is not dependent on any specific model architecture. The experiment was conducted as follows Section~\ref{PS}.

The results presented in Figure~\ref{fig:PS on vgg appendix} demonstrate that the PS phenomenon is also evident within the VGG architecture, similar with the observations made in the main paper for the ResNet-18 model. Notably, a certain degree of PS is also present even in the benign model, affecting both the clean training data and the clean validation data. Consistent with the findings under the BadNet and WaNet attacks, there still exist specific dropout $p$ that can cause the clean data to exhibit a strong PS phenomenon while that of backdoor data is extremely weak.

The findings align with our conclusion that the PS phenomenon and neuron bias effect are widely prevalent in DNNs, rather than being specific to particular model architectures. While the variations in model architecture may result in different shift ratio curves and shift intensity, they do not impact the existence of the PS phenomenon and neuron bias effect.

\section{Detailed Results of Neuron Bias Effect}
\label{features w/ and w/o dropout}
\begin{figure*}[h]
    \centering
     \begin{subfigure}[h]{0.4\textwidth}
         \centering
         \includegraphics[width=\textwidth]{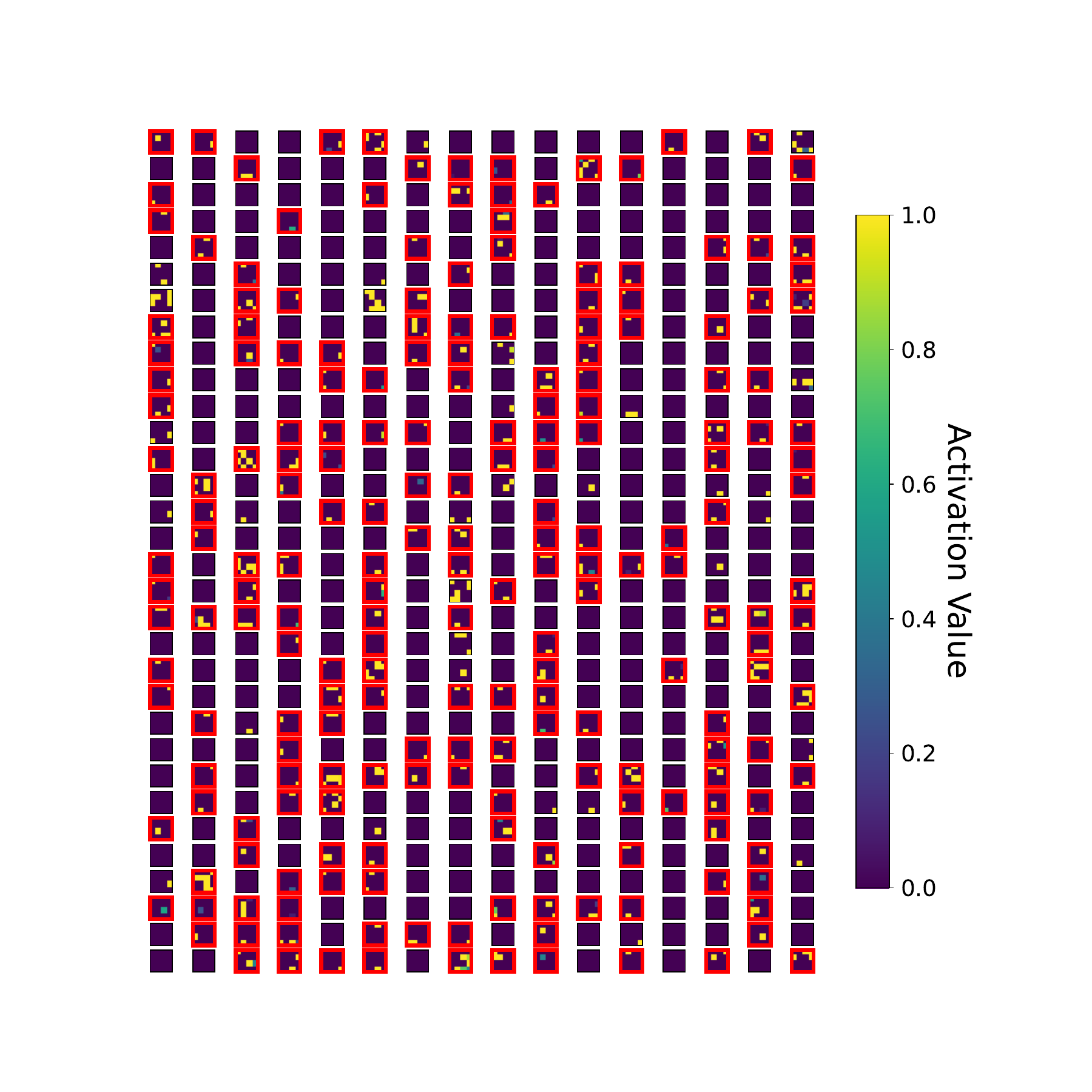}
         \vspace*{-10mm}
         \caption{Clean Image}
     \end{subfigure}
     \hspace*{-2mm}
     \begin{subfigure}[h]{0.4\textwidth}
         \centering
         \includegraphics[width=\textwidth]{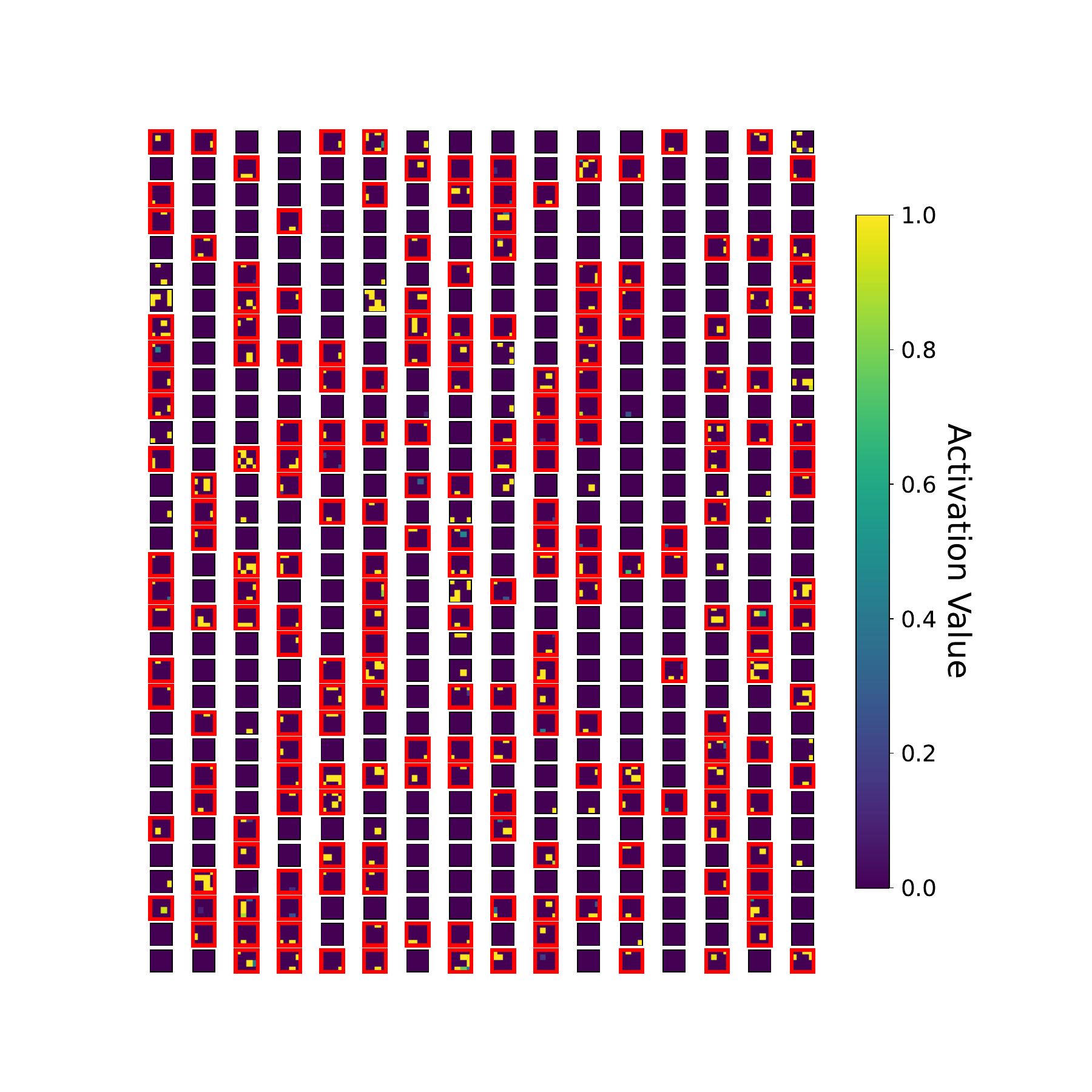}
         \vspace*{-10mm}
         \caption{Backdoor Image}
     \end{subfigure}
    \caption{\footnotesize{The all 512 activation maps extracted by the top layer of the BadNets model with dropout. The red boxes represent the feature map values are non-zero and the difference between each activation value in the clean and backdoor feature maps is no greater than 1. The features of clean and backdoor image become almost identical with dropout, verifying the existence of neuron bias effect.}}
    \label{fig: full activation map w/ dropout under badnet}
\end{figure*}

\begin{figure*}[h]
    \centering
     \begin{subfigure}[h]{0.4\textwidth}
         \centering
         \includegraphics[width=\textwidth]{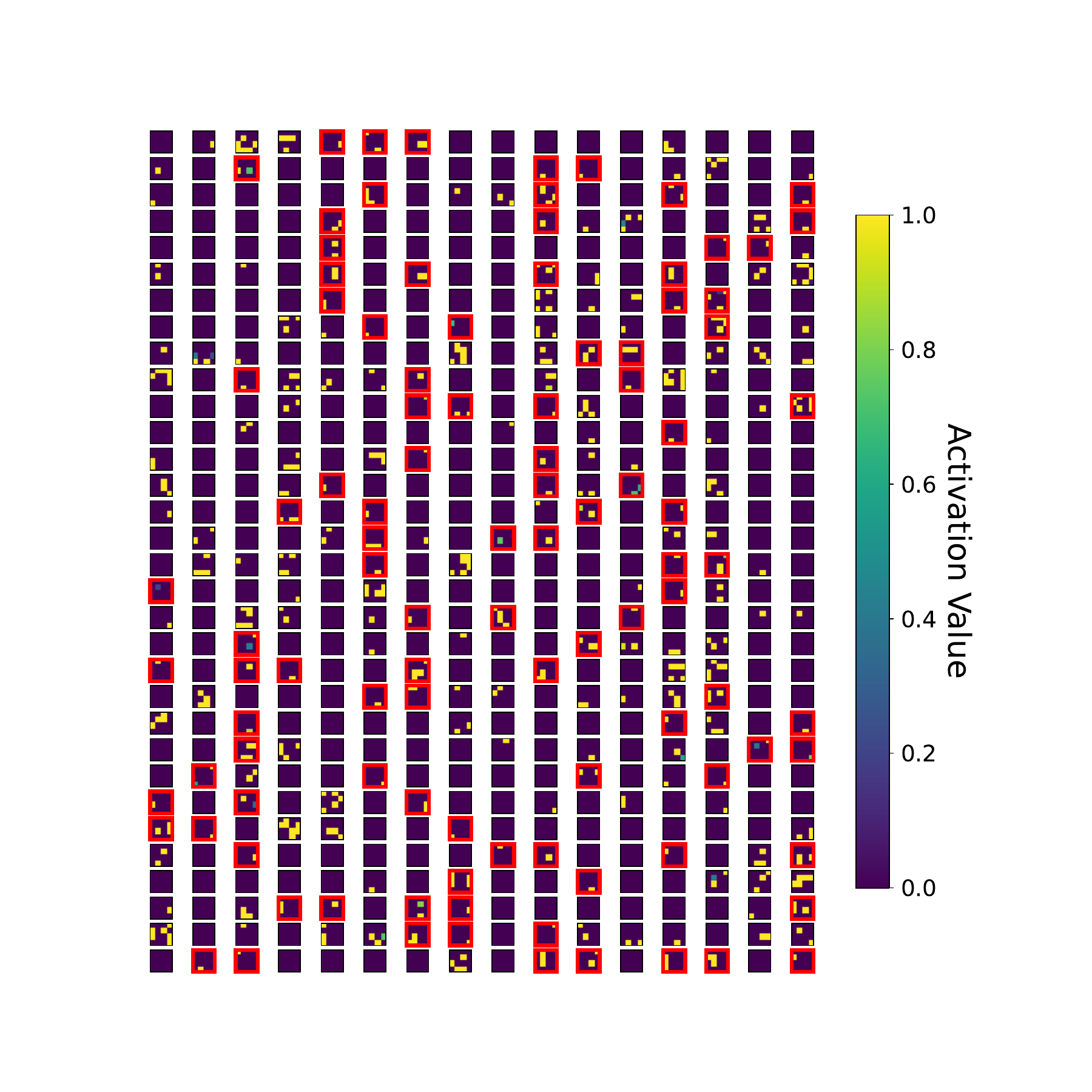}
         \vspace*{-10mm}
         \caption{Clean Image}
     \end{subfigure}
     \hspace*{-2mm}
     \begin{subfigure}[h]{0.4\textwidth}
         \centering
         \includegraphics[width=\textwidth]{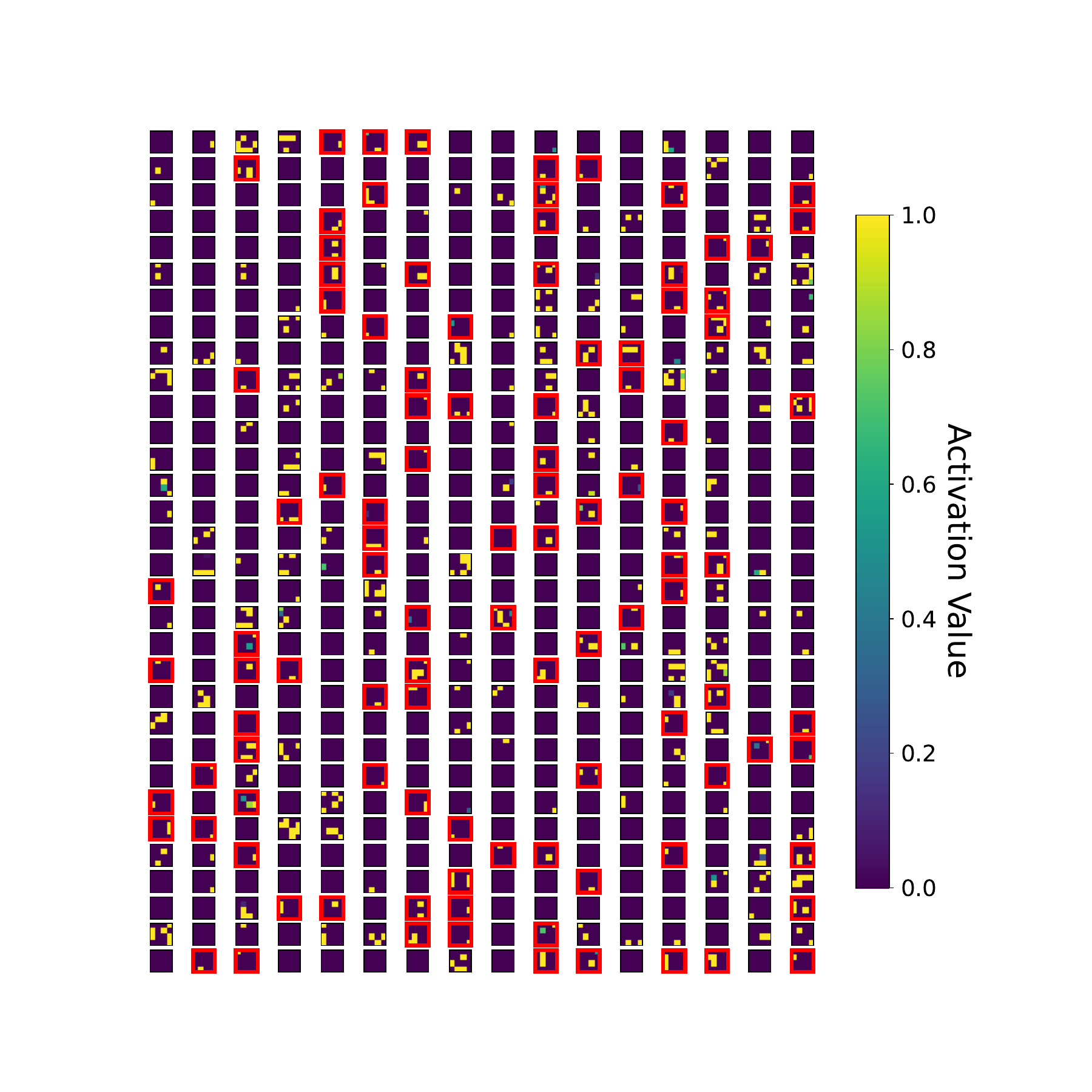}
         \vspace*{-10mm}
         \caption{Backdoor Image}
     \end{subfigure}
    \caption{\footnotesize{The all 512 activation maps extracted by the top layer of the WaNet model with dropout. The features of clean and backdoor image become more identical with dropout, verifying the neuron bias effect is not limited to specific attack.}}
    \label{fig: full activation map w/ dropout under wanet}
\end{figure*}

In this section, we present the more complete results of the ``neuron bias'' effect on the BadNets and WaNet models. As illustrated in Figures~\ref{fig: full activation map w/ dropout under badnet} and~\ref{fig: full activation map w/ dropout under wanet}, there is a pronounced and widespread presence of the neuron bias effect. This confirms our hypothesis that the PS Phenomenon is a result of the neuron bias effect.
\vspace{-9mm}

\section{Experiments Details}
\label{exp details appendix}
\subsection{Detailed Settings for Datasets and Training of Backdoored Models.}
\label{dataset detail appendix}
\begin{table*}[h]
\vspace{-5mm}
\centering
  \caption{\footnotesize{Details for all datasets used in our experiments.}}
\label{tab: dataset details}
  \resizebox{0.7\textwidth}{!}
  {
  \begin{tabular}{cccccccc}
    \toprule
    & Dataset & \# Input size & Classes & Training images & Testing images \\
    \midrule
    \midrule 
    &CIFAR-10 & 3$\times$32$\times$32 & 10 & 50,000 & 10,000\\
    &GTSRB & 3$\times$32$\times$32 & 43 & 39,209 & 12,630\\
    &Tiny ImageNet & 3$\times$64$\times$64 & 200 & 100,000 & 10,000\\
    \bottomrule
  \end{tabular}
  } 
\end{table*}
\begin{table*}[h]
\centering
  \caption{\footnotesize{Details for training models with different datasets used in our experiments.}}
\label{tab: model details}
  \resizebox{1\textwidth}{!}
  {
  \begin{tabular}{ccccccccc}
    \toprule
    \multirow{2}{*}{Dataset} & \multirow{2}{*}{Models} & \multirow{2}{*}{Optimizer} & \multirow{2}{*}{Epochs} & Initial  & Learning Rate & Learning Rate  & \multirow{2}{*}{Momentum} & Weight \\
      &  &  &  &  Learning Rate & Scheduler  & Decay Epoch &  &  Decay\\
    \midrule
    \midrule
    CIFAR-10 & ResNet-18 & SGD & 100 & 0.1 & MultiStep LR & 50,75 & 0.9 & 1e-4\\
    GTSRB & ResNet-18 & SGD & 100 & 0.1 & MultiStep LR & 50,75 & 0.9 & 1e-4\\
    Tiny ImageNet & ResNet-18 & SGD & 100 & 0.1 & MultiStep LR & 50,75 & 0.9 & 1e-4\\
    \bottomrule
  \end{tabular}
  }    
\end{table*}
\begin{figure*}[h]
\centering
\begin{adjustbox}{max width=1\textwidth,bgcolor=verylightgray}
  \begin{tabular}{ccccccccc}
 \begin{tabular}{@{\hskip 0.76in}c@{\hskip 0.0in} }
 \parbox{5em}{\centering \footnotesize{Benign}}
 \end{tabular}
 \begin{tabular}{@{\hskip 0.0in}c@{\hskip 0.0in} }
 \parbox{5em}{\centering \footnotesize{BadNet}}
 \end{tabular}
 \begin{tabular}{@{\hskip 0.0in}c@{\hskip 0.0in} }
 \parbox{5em}{\centering \footnotesize{Blend}}
 \end{tabular}
 \begin{tabular}{@{\hskip 0.0in}c@{\hskip 0.0in} }
 \parbox{5em}{\centering \footnotesize{TrojanNN}}
 \end{tabular}
 \begin{tabular}{@{\hskip 0.0in}c@{\hskip 0.0in} }
 \parbox{5em}{\centering \footnotesize{Label-Consistent}}
 \end{tabular}
 \begin{tabular}{@{\hskip 0.0in}c@{\hskip 0.0in} }
 \parbox{5em}{\centering \footnotesize{WaNet}}
 \end{tabular}
 \begin{tabular}{@{\hskip 0.0in}c@{\hskip 0.0in} }
 \parbox{5em}{\centering \footnotesize{ISSBA}}
 \end{tabular}
 \begin{tabular}{@{\hskip 0.0in}c@{\hskip 0.0in} }
 \parbox{5em}{\centering \footnotesize{Adaptive-Blend}}
 \end{tabular}
\vspace{1mm}
\\
 \begin{tabular}{@{\hskip 0.0in}c@{\hskip 0.0in} }
 {\parbox{5em}{\centering \footnotesize{Poisoned Samples}}}
 \end{tabular}
  \begin{tabular}{@{\hskip 0.0in}c@{\hskip 0.0in} }
 \parbox[c]{5em}{\includegraphics[width=5em]{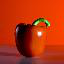}}
  \end{tabular}
  \begin{tabular}{@{\hskip 0.0in}c@{\hskip 0.0in} }
 \parbox[c]{5em}{\includegraphics[width=5em]{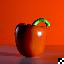}}
  \end{tabular}
  \begin{tabular}{@{\hskip 0.0in}c@{\hskip 0.0in} }
 \parbox[c]{5em}{\includegraphics[width=5em]{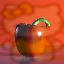}}
  \end{tabular}
  \begin{tabular}{@{\hskip 0.0in}c@{\hskip 0.0in} }
 \parbox[c]{5em}{\includegraphics[width=5em]{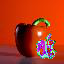}}
  \end{tabular}
  \begin{tabular}{@{\hskip 0.0in}c@{\hskip 0.0in} }
  \parbox[c]{5em}{\includegraphics[width=5em]{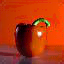}}
  \end{tabular}
  \begin{tabular}{@{\hskip 0.0in}c@{\hskip 0.0in} }
  \parbox[c]{5em}{\includegraphics[width=5em]{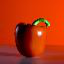}}
  \end{tabular}
  \begin{tabular}{@{\hskip 0.0in}c@{\hskip 0.0in} }
  \parbox[c]{5em}{\includegraphics[width=5em]{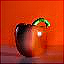}}
  \end{tabular}
  \begin{tabular}{@{\hskip 0.0in}c@{\hskip 0.0in} }
  \parbox[c]{5em}{\includegraphics[width=5em]{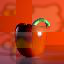}}
  \end{tabular}
 \\
 \\
  \begin{tabular}{@{\hskip 0.0in}c@{\hskip 0.0in} }
  {\parbox{5em}{\centering \footnotesize{Backdoor Trigger}}}
  \end{tabular}
  \begin{tabular}{@{\hskip 0.0in}c@{\hskip 0.0in} }
  \parbox[c]{5em}{\includegraphics[width=5em]{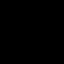}}
  \end{tabular}
  \begin{tabular}{@{\hskip 0.0in}c@{\hskip 0.0in} }
  \parbox[c]{5em}{\includegraphics[width=5em]{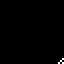}}
  \end{tabular}
  \begin{tabular}{@{\hskip 0.0in}c@{\hskip 0.0in} }
  \parbox[c]{5em}{\includegraphics[width=5em]{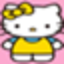}}
  \end{tabular}
  \begin{tabular}{@{\hskip 0.0in}c@{\hskip 0.0in} }
  \parbox[c]{5em}{\includegraphics[width=5em]{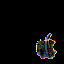}}
  \end{tabular}
  \begin{tabular}{@{\hskip 0.0in}c@{\hskip 0.0in} }
  \parbox[c]{5em}{\includegraphics[width=5em]{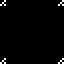}}
  \end{tabular}
  \begin{tabular}{@{\hskip 0.0in}c@{\hskip 0.0in} }
  \parbox[c]{5em}{\includegraphics[width=5em]{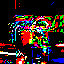}}
  \end{tabular}
  \begin{tabular}{@{\hskip 0.0in}c@{\hskip 0.0in} }
  \parbox[c]{5em}{\includegraphics[width=5em]{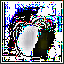}}
  \end{tabular}
  \begin{tabular}{@{\hskip 0.0in}c@{\hskip 0.0in} }
  \parbox[c]{5em}{\includegraphics[width=5em]{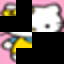}}
  \end{tabular}
\\
\\
\end{tabular}
\end{adjustbox}
\vspace*{-3mm}
\caption{\footnotesize{The various triggers of the attacks used in our study and corresponding poisoned samples.}}
\label{fig: trigger appendix}\vspace*{-4mm}
\end{figure*}
\begin{table*}[h]
\centering
\caption{\footnotesize{The performance of the benign and poisoned models with ResNet-18 architecture.}}
\label{tab: asr and ca appendix}\vspace{-3mm}
  \resizebox{1\textwidth}{!}
  {
  \begin{tabular}{ccccccccc}
    \toprule
    \multirow{2}{*}{Dataset} & Benign & BadNet & Blend & TrojanNN & Label-Consistence  & WaNet & ISSBA  & Adaptive-Blend \\
      &  CA & {ASR \hskip 2mm CA} & {ASR \hskip 2mm CA} &  {ASR \hskip 2mm CA} & {ASR \hskip 2mm CA}  & {ASR \hskip 2mm CA} & {ASR \hskip 2mm CA} & {ASR \hskip 2mm CA}\\
    \midrule
    \midrule
    CIFAR-10 & 0.853 & {1.000 \hskip 2mm 0.843} & {0.999 \hskip 2mm 0.848} & {1.000 \hskip 2mm 0.841} & {1.000 \hskip 2mm 0.850} & {0.955 \hskip 2mm 0.828} & {0.985 \hskip 2mm 0.825} & {0.871 \hskip 2mm 0.938}
    \\
    {GTSRB} & 0.981 & {1.000 \hskip 2mm 0.982} & {0.999 \hskip 2mm 0.975} & {1.000 \hskip 2mm 0.980} & {0.994 \hskip 2mm 0.975} & {0.988 \hskip 2mm 0.982} & {0.999 \hskip 2mm 0.983} & {0.913 \hskip 2mm 0.952}
    \\
    Tiny ImageNet & 0.615 & {0.998 \hskip 2mm 0.608} & {0.994 \hskip 2mm 0.607} & {0.999 \hskip 2mm 0.470} & {0.997 \hskip 2mm 0.595} & {0.996 \hskip 2mm 0.593} & {0.975 \hskip 2mm 0.455} & {0.905 \hskip 2mm 0.616}\\
    \bottomrule
  \end{tabular}
  }    
\end{table*}

The details of datasets and training procedures of DNN models in our experiments are summarized in Table~\ref{tab: dataset details} and Table~\ref{tab: model details}. Note that clean validation data refers to the clean data we used to filter backdoor training data, which was 5\% of the total quantity of whole training dataset randomly selected from the test set of CIFAR-10 and GTSRB\cite{stallkamp2011german}, and the validation set of Tiny ImageNet, respectively.

\subsection{Implements of Backdoor Attacks.}
\label{attack details appendix}
BadNets, TrojanNN\cite{liu2018trojaning}, and Blend\cite{chen2017targeted} correspond to typical all-to-one label-poisoned attacks with patch-like trigger, generated trigger, and blending trigger respectively. Label-Consistent is a representative clean label attack. WaNet is an image transformation-based invisible attack. ISSBA is an effective sample-specific invisible attack which generates sample-specific invisible additive noises as backdoor triggers. It generates sample-specific invisible additive noises as backdoor triggers by encoding an attacker-specified string into benign images through an encoder-decoder network. 
Adaptive-Blend is an adaptive poisoning strategy suggested that can suppress the latent separability characteristic.

In order to better reconstruct the different methods of obtaining backdoor data in practice, we have implemented a portion of the attacks using an open-source toolkit - ``backdoor-toolbox''. 
\footnote{\url{https://github.com/vtu81/backdoor-toolbox}} 
We are able to control the relevant settings for this subset of attacks. For the other attacks, we directly utilize the backdoor data provided by an open-source repository - ``BackdoorBench'', 
\footnote{\url{https://github.com/SCLBD/BackdoorBench}} 
which is more common and important in practice, as we lack the corresponding knowledge about backdoor attacks. We show the examples of both triggers and the corresponding poisoned samples in Figure~\ref{fig: trigger appendix}.

\paragraph{BadNet.}
We implemented this attack using the backdoor-toolbox. The trigger we used on CIFAR-10 and GTSRB is a 3$\times$3 checkerboard placed in the bottom right corner of an image, and a 6$\times$6 trigger placed in the same position on Tiny ImageNet.

\paragraph{Blend.}
We implemented this attack using the backdoor-toolbox. Following the original paper \cite{chen2017targeted}, we choose ``Hello Kitty'' trigger. The blend ratio is set to 0.2.

\paragraph{TrojanNN.}
We directly use the data provided by BackdoorBench.

\paragraph{Label-Consistent.}
On CIFAR-10, we directly use the adversarial images provided by the original paper;
\footnote{\url{https://github.com/MadryLab/label-consistent-backdoor-code}} 
on GTSRB and Tiny ImageNet we use the adversarial images provided by BackdoorBench. 

\paragraph{WaNet.}
We implemented this attack using the backdoor-toolbox. In line with the original paper \cite{nguyen2021wanet}, we maintained consistency by setting the cover ratio to twice the poisoning ratio. This means that for every poisoned data sample, there were two additional interference data samples. These interference data samples still carried the backdoor trigger but their labels were not modified to the target class.

\paragraph{ISSBA.}
We directly use the data provided by BackdoorBench.

\paragraph{Adaptive-Blend.}
We implemented this attack using the backdoor-toolbox. Following the original paper \cite{qi2022revisiting}, we choose ``Hello Kitty'' trigger and set the  the cover ratio equal to the poisoning ratio.
Compared to the original paper, we only added the cover ratio and poisoning ratio to ensure that the attack success rate exceeds 85\%. On CIFAR-10 and GTSRB, we selected the ``Hello Kitty'' trigger, setting both the cover ratio and poisoning ratio to 0.01, and the blend ratio to 0.2. On Tiny ImageNet, we chose a random noise trigger, setting both the cover ratio and poisoning ratio to 0.02, and the blend ratio to 0.15.

\subsection{Implements of Baseline Defences.}
\label{baseline details appendix}
We implement Spectral Signature \cite{tran2018spectral}, Strip\cite{gao2019strip}, Spectre\cite{hayase2021spectre} and SCAN \cite{tang2021demon} based on the original implementation provided by the backdoor-toolbox. We have implemented the SCP \cite{guo2023scale} based on the backdoor-toolbox. {We use the original implementation of CD-L \cite{huang2023distilling} and follow the hyperparameter settings specified in the original paper.}

\subsection{Performance of the Benign and Poisoned Models.}
Consistent with the methodology employed in previous backdoor attack studies, we utilize performance metrics to assess the effectiveness of the backdoor attacks: attack success rate (ASR) and clean accuracy (CA). ASR denotes the success rate in classifying the poisoned samples into the corresponding target classes. CA measures the accuracy of the backdoored model on the benign test dataset.
ASR and CA for different backdoor attacks are included in Table~\ref{tab: asr and ca appendix}. The poisoning ratio is 10\%.

\subsection{Computational Environment}
\label{computational env}
All experiments are conducted on a server with the Ubuntu 18.04.6 LTS operating system, a 2.10GHz CPU, 3 NVIDIA’s GeForce GTX3090 GPUs with 24G RAM.

\section{Detection Performance on Low Poisoning Ratio Scenario}
\label{low pr appendix}
We evaluate the performance of all detection methods under scenarios with low poisoning ratios. A small poisoning ratio prevents models from overfitting to triggers, thereby weakening the connection between triggers and target labels and presenting a significant challenge for backdoor data detection.

As shown in Table~\ref{tab: poison rate 0.05}, PSBD demonstrated superior performance against various backdoor attacks under low poisoning scenarios, outperforming all other baseline methods on average. However, its performance exhibited a slight decline in certain attack settings. This degradation may be attributed to the reduced robustness of neuron bias paths within the model due to the limited amount of backdoor data, making it more susceptible to random fluctuations. The performance of other baseline methods deteriorated significantly, particularly on the more challenging Tiny ImageNet dataset. 

\begin{table*}[t]
\centering
\caption{\footnotesize{
The performance (TPR/FPR) on CIFAR-10, GTSRB and Tiny ImageNet. We mark the \textbf{best result} in boldface while the value with underline denotes the \underline{second-best}. The \textcolor{lightgray}{failed cases} (i.e., TPR < 0.8) are marked in gray. We use the same results as in Table~\ref{tab: poison rate 0.1} for Adaptive-Blend attack as the 1\% and 2\% poisoning ratios are sufficiently low. Other attacks have a 5\% poisoning ratio. OOT indicates that the method did not finish within the allocated time limit.
}}
\label{tab: poison rate 0.05}
\resizebox{0.9\textwidth}{!}{
\begin{tabular}{c|c|c|c|c|c|c|c}
\toprule[1pt]
\midrule
\multirow{1}{*}{Defenses$\rightarrow$} & \multirow{2}{*}{PSBD (\textbf{Ours})} & \multirow{2}{*}{Spectral Signature} & \multirow{2}{*}{Strip} & \multirow{2}{*}{Spectre} & \multirow{2}{*}{SCAN} & \multirow{2}{*}{SCP} & \multirow{2}{*}{CD-L} \\
\multirow{1}{*}{Attacks$\downarrow$} & & & & & & \\
\rowcolor{LightCyan}
\midrule
\multicolumn{8}{c}{\hspace{20mm}\textbf{CIFAR-10}\hspace{8mm}} \\
\midrule
Badnet & {0.979/0.158} & {0.977/0.475} & \underline{1.000/0.115} & 1.000/0.473 & \textbf{1.000/0.090} & 1.000/0.199 &
0.999/0.164 \\ 
Blend  & {0.899/0.176} & {0.892/0.479} & \underline{0.984/0.114} & \textbf{1.000/0.473} & {0.973/0.015} & 
0.979/0.236 &
0.957/0.161 \\
TrojanNN & 0.951/0.175 & {0.925/0.478} & \textbf{1.000/0.117} & 0.969/0.475 & {0.998/0.018} & 0.967/0.201 &
\underline{1.000/0.162} \\
Label-Consistent & \textbf{1.000/0.107} & {0.895/0.479} & {0.977/0.117} & \underline{0.999/0.473} & 0.970/0.019 & 0.910/0.201
& 0.994/0.166 \\ 
WaNet & \textbf{1.000/0.113} & {0.820/0.483} & \textcolor{lightgray}{0.044/0.107} & \underline{0.985/0.475} & 0.856/0.036 & {0.861/0.220} &
\textcolor{lightgray}{0.430/0.149} \\  
ISSBA & \underline{0.998/0.153} & {0.877/0.480} & \textcolor{lightgray}{0.712/0.120} & \textbf{0.999/0.474} & {0.945/0.008} & 0.937/0.271 &
0.984/0.163\\
Adaptive-Blend & \textbf{0.982/0.184} & \textcolor{lightgray}{0.608/0.145} & \textcolor{lightgray}{0.014/0.069} & \textcolor{lightgray}{0.753/0.144} & \textcolor{lightgray}{0.000/0.023} & \textcolor{lightgray}{0.779/0.246} &
\textcolor{lightgray}{0.432/0.167} \\

\rowcolor{Gray}
    Average & \textbf{0.973/0.152} & 0.861/0.431 & 0.714/0.109 & \underline{0.963/0.427} & {0.839/0.018} & 0.918/0.225
    & 0.828/0.162\\ 
\rowcolor{LightCyan}
\midrule
\multicolumn{8}{c}{\hspace{20mm}\textbf{GTSRB}\hspace{8mm}} \\
\midrule
Badnet & {0.993/0.202} & \textcolor{lightgray}{0.448/0.502} & \underline{0.995/0.094} & \textcolor{lightgray}{0.552/0.497} & OOT & \textbf{1.000/0.328} &
0.839/0.188 \\ 
Blend  & \underline{0.859/0.223} & \textcolor{lightgray}{0.448/0.502} & \textbf{0.915/0.094} & \textcolor{lightgray}{0.552/0.497} & {OOT} & 
\textcolor{lightgray}{0.301/0.334} &
\textcolor{lightgray}{0.072/0.198} \\
TrojanNN & \textbf{0.978/0.206} & \textcolor{lightgray}{0.449/0.502} & \textcolor{lightgray}{0.408/0.093} & \textcolor{lightgray}{0.551/0.497} & {OOT} & \textcolor{lightgray}{0.150/0.329} &
\textcolor{lightgray}{0.515/0.191} \\
Label-Consistent & {0.844/0.202} & \textcolor{lightgray}{0.449/0.502} & \textbf{0.998/0.113} & \textcolor{lightgray}{0.551/0.497} & OOT & \underline{0.956/0.403} & \textcolor{lightgray}{0.198/0.172} \\ 
WaNet & \textbf{0.999/0.085} & \textcolor{lightgray}{0.448/0.502} & \textcolor{lightgray}{0.030/0.102} & \textcolor{lightgray}{0.552/0.497} & OOT & \textcolor{lightgray}{0.043/0.320} &
\textcolor{lightgray}{0.022/0.185} \\  
ISSBA & \textbf{0.986/0.214} & \textcolor{lightgray}{0.449/0.502} & \textcolor{lightgray}{0.469/0.102} & \textcolor{lightgray}{0.551/0.497} & {OOT} & \textcolor{lightgray}{0.590/0.334} &
\textcolor{lightgray}{0.446/0.195}\\
Adaptive-Blend & \textbf{0.899/0.194} & \textcolor{lightgray}{0.299/0.392} & \textcolor{lightgray}{0.004/0.094} & \textcolor{lightgray}{0.750/0.388} & {OOT} & \textcolor{lightgray}{0.071/0.332} &
\textcolor{lightgray}{0.028/0.158} \\

\rowcolor{Gray}
    Average & \textbf{0.937/0.189} & 0.427/0.486 & 0.546/0.099 & \underline{0.580/0.481 }& {OOT} & 0.444/0.340
    & 0.303/0.184 \\ 
\rowcolor{LightCyan}
\midrule
\multicolumn{8}{c}{\hspace{20mm}\textbf{Tiny ImageNet}\hspace{8mm}} \\
\midrule
Badnet & \underline{0.996/0.093} & \textcolor{lightgray}{0.452/0.502} & 0.878/0.109 & \textcolor{lightgray}{0.548/0.497} & OOT & \textbf{0.998/0.279} &
\textcolor{lightgray}{0.390/0.178}\\
Blend  & \textbf{0.871/0.065} & \textcolor{lightgray}{0.453/0.502} & \textcolor{lightgray}{0.558/0.097} & \textcolor{lightgray}{0.548/0.497} & OOT & \textcolor{lightgray}{0.624/0.269} &
\textcolor{lightgray}{0.376/0.185} \\
TrojanNN & 0.939/0.203 & \textcolor{lightgray}{0.453/0.502} & {0.980/0.107} & \textcolor{lightgray}{0.547/0.497} & OOT & \textbf{0.991/0.279} &
\underline{0.990/0.166}\\
Label-Consistent & \textbf{0.983/0.100} & \textcolor{lightgray}{0.452/0.502} & \textcolor{lightgray}{0.518/0.090} & \textcolor{lightgray}{0.548/0.497} & OOT & \underline{0.978/0.092} &
0.967/0.154\\
WaNet & \textbf{0.944/0.109} & \textcolor{lightgray}{0.452/0.502} & \textcolor{lightgray}{0.107/0.093} & \textcolor{lightgray}{0.548/0.497} & OOT & \textcolor{lightgray}{0.314/0.267} &
\textcolor{lightgray}{0.403/0.150}\\
ISSBA & \underline{0.890/0.216} & \textcolor{lightgray}{0.452/0.502} & \textbf{0.994/0.104} & \textcolor{lightgray}{0.547/0.497} & OOT & \textcolor{lightgray}{0.663/0.320} & \textcolor{lightgray}{0.644/0.140}\\
Adaptive-Blend & \textbf{0.949/0.095} & \textcolor{lightgray}{0.392/0.502} & \textcolor{lightgray}{0.210/0.099} & \textcolor{lightgray}{0.621/0.497} & OOT & \textcolor{lightgray}{0.505/0.218} &
\textcolor{lightgray}{0.331/0.176}\\

\rowcolor{Gray}
    Average & \textbf{0.939/0.126} & 0.445/0.502 & 0.595/0.100 & 0.557/0.497& OOT & \underline{0.684/0.265} &
    0.586/0.164\\
\midrule
\bottomrule[1pt]
\end{tabular}}
\vspace*{-2mm}
\end{table*}

\section{Resistance to Potential Adaptive Attacks}
\label{adaptive_attack appendix}
The most common adaptive attack scenario is one with a low poisoning ratio. As shown in Section~\ref{low pr appendix}, our PSBD method demonstrates effective performance.

We further evaluate the robustness of PSBD against potential adaptive attacks in the worst-case scenario, where adversaries have complete knowledge of our defense. Typically, a vanilla backdoored model performs normally with benign samples but produces adversary-specific predictions when exposed to poisoned samples. The objective function for training such a model with a poisoned training dataset can be represented as follows:
\begin{equation}
    \min_{\boldsymbol{\theta}} \mathcal{L}_{bd}(\trainc \cup \trainb;\boldsymbol{\theta})
\end{equation}
where $\boldsymbol{\theta}$ denotes the model parameters and $\mathcal{L}$ is the cross entropy loss function.
We develop an adaptive attack by introducing a loss term specifically designed to ensure that benign samples have a low PSU value. This adaptive loss item $\mathcal{L}_{ada}$ is defined as:
\begin{equation}
    \mathcal{L}_{ada} = \ \phi_{PSU}(\mathbf{x}), \mathbf{x} \in \trainc \cup \trainb
\end{equation}

Subsequently, we integrate this adaptive loss $\mathcal{L}_{ada}$ with the vanilla loss $\mathcal{L}_{bd}$ to formulate the overall loss function as $\mathcal{L} = (1-\alpha) \mathcal{L}_{bd} + \alpha \mathcal{L}_{ada}$ , where $\alpha$ is a weighting factor. We then optimize the original model’s parameters $\boldsymbol \theta$ by minimizing $\mathcal{L}$ during the training phase. Please note that, to maintain the effectiveness of the training process, we compute $\mathcal{L}_{ada}$ every 50 iterations. As in previous experiments, we also use two representative backdoor attacks, BadNets and WaNet, to develop adaptive attacks on the CIFAR-10 dataset.

The adversary aims to find a value of $\alpha$ that best balances the ASR and CA. Table~\ref{tab: asr and ca adaptive attack appendix} presents the performance of
\begin{table}[!h]
\centering
\caption{\footnotesize{
The attack performance of adaptive attacks.
}}
\label{tab: asr and ca adaptive attack appendix}
\resizebox{0.4\textwidth}{!}{
\begin{tabular}{cccc}
\toprule[1pt]
\midrule
\multirow{1}{*}{$\alpha$ $\rightarrow$} & \multirow{1}{*}{0.2} & \multirow{1}{*}{0.5} & \multirow{1}{*}{0.9} \\ 
\multirow{1}{*}{Attacks$\downarrow$} & {ASR \hskip 2mm CA} & {ASR \hskip 2mm CA} & {ASR \hskip 2mm CA} \\
\midrule
Badnet & 1.000 0.828 & 1.000 0.838 & 1.000 0.836\\
WaNet & 0.899 0.803 & 0.931 0.820 & 0.925 0.823 \\
\midrule
\bottomrule[1pt]
\end{tabular}}
\end{table}
\begin{table}[!h]
\centering
\caption{\footnotesize{
Performance (TPR/FPR) of PSBD under adaptive attacks.
}}
\label{tab: against adaptive attack appendix}
\resizebox{0.4\textwidth}{!}{
\begin{tabular}{cccc}
\toprule[1pt]
\midrule
\multirow{1}{*}{$\alpha$ $\rightarrow$} & \multirow{2}{*}{0.2} & \multirow{2}{*}{0.5} & \multirow{2}{*}{0.9} \\ 
\multirow{1}{*}{Attacks$\downarrow$} & & \\
\midrule
Badnet & 0.989/0.157 & 0.997/0.131 & 0.967/0.127 \\
WaNet & 1.000/0.119 & 0.998/0.138 & 1.000/0.114 \\
\midrule
\bottomrule[1pt]
\end{tabular}}
\end{table}
the adaptive attacks under various $\alpha$ settings. As shown in the results, both attacks (BadNets and WaNet) on the CIFAR-10 dataset consistently exhibit high ASR and CA across different values of $\alpha$, highlighting the effectiveness of the adaptive attacks.

On the other hand, Table~\ref{tab: against adaptive attack appendix} shows that adaptive attacks can still be effectively defended by our method. We conducted further investigation and observed that the adaptive loss indeed caused the model to behave differently from the non-adaptive version. 
However, our defense can adapt to modified backdoor models. Specifically, we observed that a dropout rate of 0.7 was used for the non-adaptive backdoor model, as detailed in Section~\ref{PS} of the main paper. In contrast, the dropout rate for the adaptive model was 0.8. In other words, our algorithm learns to select the appropriate dropout rate for different models. This ability to counter adaptive attacks is a key advantage of our method compared to previous approaches.

\begin{table}[!h]
\centering
\caption{\footnotesize{
The AUROC values (AUROC) on Tiny ImageNet.
}}
\label{tab: auroc}
\resizebox{0.4\textwidth}{!}{
\begin{tabular}{c|c|c}
\toprule[1pt]
\midrule
\multirow{1}{*}{Defenses$\rightarrow$} & \multirow{2}{*}{PSBD (\textbf{Ours})} & \multirow{2}{*}{CD-L} \\
\multirow{1}{*}{Attacks$\downarrow$} & & \\
\midrule
Badnet & 0.993 & \textcolor{lightgray}{0.474}\\
Blend  & 0.958 & 0.922\\
TrojanNN & 0.963 & 0.877\\
Label-Consistent & 0.996 & \textcolor{lightgray}{0.548}\\
WaNet & 0.985 & 0.831 \\
ISSBA & 0.907 & 0.801\\
Adaptive-Blend & 0.969 & \textcolor{lightgray}{0.705} \\

\rowcolor{Gray}
    Average & 0.967 & 0.737 \\
\midrule
\bottomrule[1pt]
\end{tabular}}
\vspace*{-2mm}
\end{table}
\section{AUROC Metric}
\label{AUROC appendix}
In addition to the TPR and FPR metrics, we also compare the AUROC metric with the CD-L method. We evaluate the AUROC on the more challenging Tiny ImageNet dataset, and the results are presented in Table~\ref{tab: auroc}. We observe that our method achieves a high AUROC score across various attacks, which verifies the robustness of our method in selecting the threshold parameter $T$.

\section{Limitations}
\label{limitations}
While this study introduces a promising approach to enhancing the security of DNNs through the PSBD method, it also has several limitations. 

The majority of existing approaches, such as SCP, CatchBackdoor~\cite{jin2025catchbackdoor} and our PSBD primarily rely on empirical findings with limited theoretical foundations. Developing solid theoretical justifications for these methods remains important future work.

Our experiments use model architectures and datasets consistent with prior studies~\cite{guo2023scale, jin2025catchbackdoor} to ensure fair comparability, which represents the most common experimental setup in the field. As discussed in Sections~\ref{defender's goal} and~\ref{exp settings}, while defenders can employ any model or training strategy to effectively detect backdoor data, the generalizability of the Prediction Shift phenomenon to more complex architectures, such as ViT\cite{dosovitskiy2021an}, and larger-scale datasets remains a valuable avenue for future exploration.